\documentclass[11pt,a4paper]{article}
\usepackage[left=25mm,right=25mm,top=18mm,bottom=20mm,headheight=14pt,headsep=10pt]{geometry}
\usepackage[T1]{fontenc}
\usepackage{mathptmx}
\usepackage[scaled=0.94]{helvet}
\usepackage{inconsolata}
\usepackage[authoryear,round]{natbib}
\PassOptionsToPackage{x11names,table}{xcolor}

\usepackage{amsmath,amsfonts,bm}

\def\eqref#1{equation~\ref{#1}}

\def\1{\bm{1}}

\DeclareMathAlphabet{\mathsfit}{\encodingdefault}{\sfdefault}{m}{sl}
\SetMathAlphabet{\mathsfit}{bold}{\encodingdefault}{\sfdefault}{bx}{n}

\DeclareMathOperator*{\argmin}{arg\,min}

\usepackage{graphicx}
\usepackage{float}
\graphicspath{{figures/}}
\DeclareGraphicsExtensions{.pdf,.jpeg,.png}
\usepackage{amsmath,amssymb,amsfonts,bm}
\usepackage{algorithm}
\usepackage{algorithmic}
\usepackage{array}
\usepackage{tabulary,xfrac,enumitem}
\usepackage{booktabs}
\usepackage{nicefrac}
\usepackage{microtype}
\usepackage{tabularx}
\usepackage{multicol}
\usepackage{multirow}
\usepackage{textcomp}
\usepackage{makecell}
\usepackage{placeins}
\usepackage{pifont}
\usepackage{xspace}
\usepackage[x11names,table]{xcolor}
\usepackage{verbatim}
\usepackage[breakable,skins,listings]{tcolorbox}

\definecolor{promptbackground}{RGB}{247,249,251}
\definecolor{promptheader}{RGB}{44,57,74}
\definecolor{promptborder}{RGB}{202,211,221}
\definecolor{prompttext}{RGB}{35,43,53}
\definecolor{promptoutput}{RGB}{39,49,63}
\definecolor{promptlight}{RGB}{245,247,250}
\tcbset{unimetric prompt style/.style={
  enhanced standard,
  use color stack,
  colback=promptbackground,
  colframe=promptborder,
  colbacktitle=promptheader,
  coltitle=promptlight,
  coltext=prompttext,
  fonttitle=\small\bfseries,
  fontupper=\fontsize{8.5}{10.5}\selectfont\rmfamily,
  before upper={\setlength{\parindent}{0pt}\setlength{\parskip}{2pt}},
  boxrule=0.4pt,
  sharp corners,
  boxsep=0pt,
  left=8pt,right=8pt,top=5pt,bottom=6pt,
  toptitle=3pt,bottomtitle=3pt,
  before skip=7pt,after skip=7pt
}}
\newtcolorbox[auto counter,number within=section]{unimetricprompt}[2]{
  unimetric prompt style,
  breakable,
  lines before break=10,
  title={Box~\thetcbcounter.\ #1},
  title after break={Box~\thetcbcounter.\ #1 (continued)},
  label={#2}
}
\newtcolorbox{unimetricpromptfigure}{
  unimetric prompt style,
  title={Height-only user prompt},
  before skip=0pt,after skip=0pt
}
\newtcolorbox{unimetricpromptbox}[1]{
  unimetric prompt style,
  title={#1},
  before skip=0pt,after skip=0pt
}
\newenvironment{promptsection}[1]{%
  \par\penalty0\addvspace{4pt}\begin{samepage}%
  \noindent\textbf{#1}\par\nobreak
}{\par\end{samepage}\penalty0}
\newlist{promptitems}{itemize}{1}
\setlist[promptitems]{label=\textendash,leftmargin=10pt,labelsep=4pt,
  itemsep=1pt,parsep=0pt,topsep=1pt,partopsep=0pt}
\newtcolorbox{unimetricpromptoutput}{
  enhanced standard,
  colback=promptoutput,colframe=promptoutput,
  coltext=promptlight,colbacktitle=promptoutput,coltitle=promptlight,
  title={Output format},fonttitle=\fontsize{8.5}{10.5}\selectfont\bfseries,
  fontupper=\fontsize{8.5}{10.5}\selectfont\rmfamily,
  before upper={\setlength{\parindent}{0pt}\setlength{\parskip}{0pt}},
  sharp corners,boxrule=0pt,boxsep=0pt,
  left=7pt,right=7pt,top=3pt,bottom=5pt,toptitle=4pt,bottomtitle=0pt,
  before skip=4pt,after skip=4pt
}
\newcommand{\promptfield}[2]{\par\noindent\texttt{#1}: #2}
\lstdefinestyle{promptjson}{
  basicstyle=\fontsize{8}{10}\selectfont\ttfamily\color{promptlight},
  columns=fullflexible,keepspaces=true,breaklines=true,
  breakatwhitespace=true,showstringspaces=false,
  aboveskip=0pt,belowskip=0pt
}
\usepackage{marvosym}
\usepackage{ragged2e}
\usepackage{url}
\usepackage{hyperref}

\usepackage[capitalize]{cleveref}
\crefname{section}{Sec.}{Secs.}
\Crefname{section}{Sec.}{Secs.}
\crefname{appendix}{Appendix}{Appendices}
\Crefname{appendix}{Appendix}{Appendices}
\crefname{table}{Tab.}{Tabs.}
\Crefname{table}{Table}{Tables}
\crefname{equation}{Eq.}{Eqs.}
\Crefname{equation}{Eq.}{Eqs.}
\crefname{figure}{Fig.}{Figs.}
\Crefname{figure}{Fig.}{Figs.}

\newcommand{\rev}[1]{#1}                 
\newcommand{\chg}[1]{#1}                 

\definecolor{citecolor}{RGB}{34,200,34}
\definecolor{metricimprove}{RGB}{22,128,65}
\definecolor{metriczero}{RGB}{100,100,100}

\newcommand{\unimetrictablefont}{\fontsize{8}{9.5}\selectfont}

\usepackage{titlesec}
\usepackage{fancyhdr}
\usepackage{caption}
\renewcommand{\bibfont}{\small}

\definecolor{reportbackground}{HTML}{F3F0F8}
\definecolor{reportink}{HTML}{000000}
\definecolor{reportblue}{HTML}{62438D}
\definecolor{reportmuted}{HTML}{776A87}
\hypersetup{colorlinks=true,urlcolor=reportblue,citecolor=reportblue,linkcolor=reportblue,
  pdftitle={UniMetric: Universal Metric Scale Recovery for Visual Geometry Foundation Models},
  pdfauthor={Lei Yang, Yuner Zhang, Ziyang Liu, Ziying Song, Kai WU, Yiyang Gao, Jiangtao Su, Yiming Li},
  pdfsubject={Technical Report}}
\titleformat{\section}{\large\bfseries}{\thesection}{0.8em}{}
\titleformat{\subsection}{\normalsize\bfseries}{\thesubsection}{0.8em}{}
\titleformat{\subsubsection}{\normalsize\bfseries}{\thesubsubsection}{0.8em}{}
\titleformat{\paragraph}[runin]{\normalsize\bfseries}{}{0pt}{}
\titlespacing*{\section}{0pt}{16pt}{7pt}
\titlespacing*{\subsection}{0pt}{12pt}{5pt}
\titlespacing*{\subsubsection}{0pt}{10pt}{4pt}
\titlespacing*{\paragraph}{0pt}{7pt}{0.6em}
\renewcommand{\headrulewidth}{0pt}
\renewcommand{\footrulewidth}{0pt}
\fancypagestyle{reportfirst}{
  \fancyhf{}
  \fancyfoot[C]{\fontsize{9}{11}\selectfont\color{reportmuted}\thepage}
  \renewcommand{\headrulewidth}{0pt}
  \renewcommand{\footrulewidth}{0pt}
}
\AddToHook{file/exps_depth_main.tex/before}{\begingroup\setlength{\extrarowheight}{3pt}}
\AddToHook{file/exps_depth_main.tex/after}{\endgroup}
\AddToHook{file/exps_summary_suite.tex/before}{\begingroup\setlength{\extrarowheight}{3pt}}
\AddToHook{file/exps_summary_suite.tex/after}{\endgroup}

\newtcolorbox{reportfront}{
  enhanced,frame hidden,colback=reportbackground,coltext=reportink,
  arc=9pt,boxrule=0pt,boxsep=0pt,
  left=14pt,right=14pt,top=14pt,bottom=13pt,
  before skip=0pt,after skip=0pt
}

\begin{document}
\thispagestyle{reportfirst}
\begin{reportfront}
{\sffamily\fontsize{8}{10}\selectfont\color{reportmuted}
\MakeUppercase{Technical Report}\hfill October 2026\par}
\vspace{7pt}
{\sffamily\bfseries\fontsize{21.5}{24}\selectfont\raggedright\color{reportblue}
UniMetric: Universal Metric Scale Recovery for Visual Geometry Foundation Models\par}
\vspace{6pt}
{\sffamily\bfseries\fontsize{9.5}{11.5}\selectfont\raggedright
\mbox{Lei Yang\textsuperscript{1}},
\mbox{Yuner Zhang\textsuperscript{2}},
\mbox{Ziyang Liu\textsuperscript{3}},
\mbox{Ziying Song\textsuperscript{1}},
\mbox{Kai WU\textsuperscript{4}},
\mbox{Yiyang Gao\textsuperscript{5}},
\mbox{Jiangtao Su\textsuperscript{1}},
\mbox{Yiming Li\textsuperscript{6}}\par}
\vspace{3pt}
{\fontsize{8.5}{10.5}\selectfont\raggedright\setlength{\parskip}{0pt}
\textsuperscript{1}Nanyang Technological University\quad
\textsuperscript{2}Carnegie Mellon University\quad
\textsuperscript{3}The Hong Kong Polytechnic University\par
\textsuperscript{4}Tongji University\quad
\textsuperscript{5}University of Toronto\quad
\textsuperscript{6}Tsinghua University\par}
\vspace{9pt}
{\fontsize{10.5}{12.7}\selectfont\justifying\setlength{\parindent}{0pt}
\rev{\chg{
Recent advances in multimodal foundation models, exemplified by GPT-6 Astra, raise the prospect of general-purpose embodied reasoning and planning. Using Internet videos for embodied pretraining calls for reliable physical-scale annotations. Yet visual geometry foundation models often recover scene structure and camera motion without reliable metric scale.
We show that uncertain size priors of familiar objects can jointly
constrain a reconstruction's shared scale. Building on this insight, we
present UniMetric, a universal and training-free metric scale recovery
framework for geometry foundation models, without additional sensors or
task-specific metric supervision. Specifically, Metric Prior Acquisition elicits object height intervals
from a vision--language model. Semantic--Geometric Alignment measures
reconstructed object heights along the estimated support-surface normal in
each frame and converts the priors into scale intervals. Global Scale
Recovery reconciles these intervals through a convex objective, yielding a
jointly feasible scale when they overlap and a unique compromise otherwise.
The estimated sequence-level scale is applied uniformly to reconstructed
points and camera translations. Controlled experiments with real VLM priors
show that combining observations improves scale accuracy with exact relative
object heights. To evaluate metric accuracy on GFM reconstructions without
ground-truth scale alignment, we introduce MetricBench, covering twelve embodied, driving, and roadside subsets with sensor-derived ground truth. Across nine geometry foundation models, UniMetric reduces mean depth Abs Rel, reconstruction Chamfer distance, and trajectory error by 54.2\%, 83.8\%, and 65.6\%, respectively, improving every model on all three tasks at the suite level. We release the UniMetric code and MetricBench data.}}
\par}
\vspace{8pt}
{\sffamily\fontsize{9}{11}\selectfont\raggedright\setlength{\parskip}{0pt}
\textbf{Code:} \url{https://github.com/yanglei18/UniMetric}\par
\textbf{Data:} \url{https://huggingface.co/datasets/yanglei18/MetricBench}\par}
\end{reportfront}

\suppressfloats[t]
\begin{figure}[t]
\centering
\setlength{\abovecaptionskip}{1pt}
\includegraphics[width=\textwidth,trim=0bp 43bp 0bp 0bp,clip]{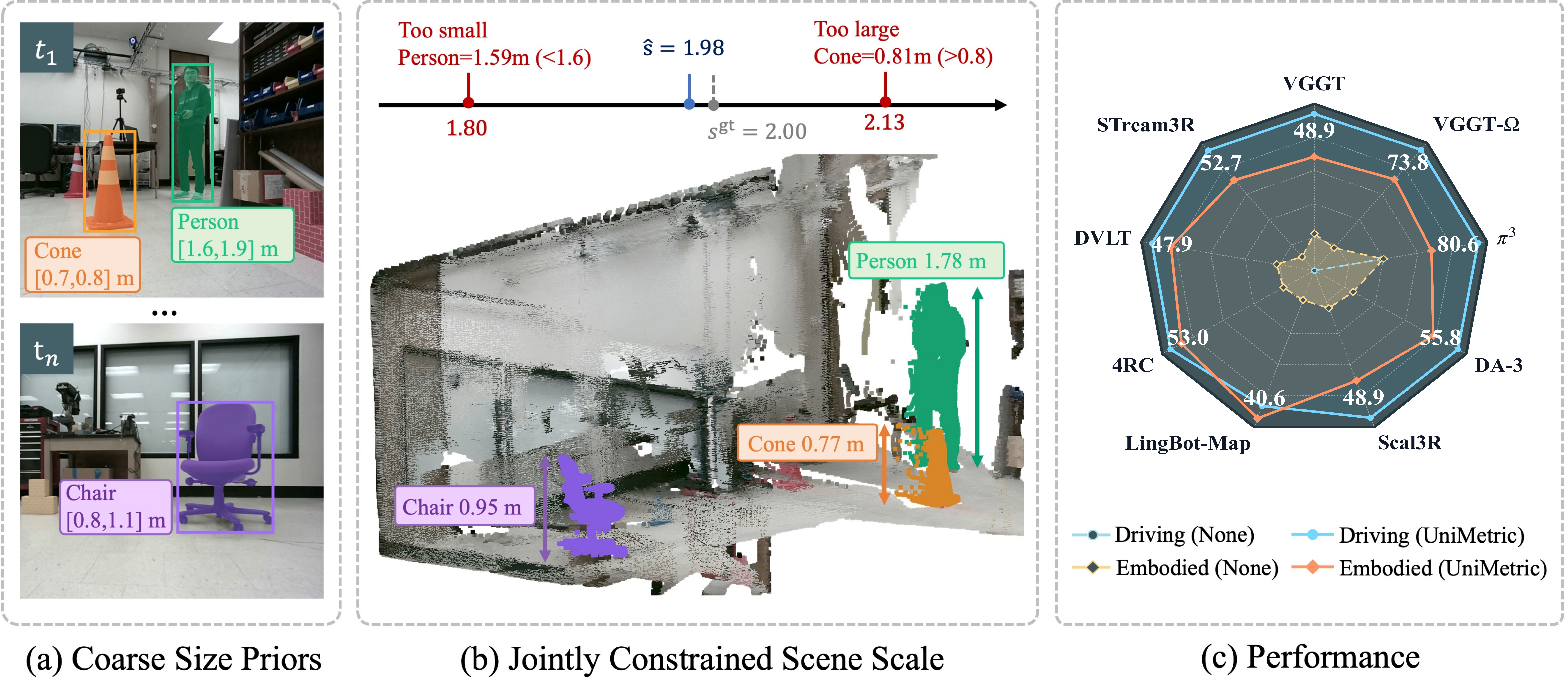}
\caption{\rev{\chg{\textbf{Uncertain object-size priors jointly constrain metric
scale.} (a) A vision--language model provides plausible height intervals for
familiar objects across frames. (b) Combined with reconstructed object
heights, these intervals constrain a shared scale. A candidate consistent
with one object may violate another's prior; their joint constraints narrow
the admissible range. Here, $\hat{s}$ and $s^{\mathrm{gt}}$ denote the
estimated and reference scales. (c) UniMetric improves metric depth accuracy
across nine geometry foundation models on MetricBench's
driving-and-infrastructure and embodied-AI suites. Curves compare original
(None) and scale-corrected (UniMetric) predictions using $\delta_1$ (\%,
higher is better), averaged equally over six subsets per suite. Evaluation
uses no ground-truth scale alignment.}}}
\label{fig:teaser}
\end{figure}

\section{Introduction}
\label{sec:introduction}
\rev{\chg{Recent advances in general-purpose multimodal foundation models,
exemplified by GPT-6 Astra~\citep{openai2026gpt6astra}, raise the prospect of a
general-purpose embodied ``brain'' for reasoning and planning across diverse
physical platforms~\citep{cheng2026incontext}. Grounding these capabilities in the physical world
requires understanding real dimensions, distances, and motion. This
motivates enriching Internet videos with reliable physical-scale annotations
to support embodied training at scale. Visual geometry foundation models
(GFMs)~\citep{wang2025vggt,wang2025pi} can recover dense structure and camera
motion from such videos, yet their reconstructions often lack reliable metric
scale. Bridging this gap is therefore key to turning large-scale visual
experience into metric data for embodied foundation-model pretraining.}}

\rev{\chg{The missing scale is commonly supplied by additional sensor
measurements~\citep{campos2021orbslam3,qin2018vinsmono,tian2021scale} or
learned from metric
supervision~\citep{bhat2023zoedepth,hu2024metric3dv2,piccinelli2024unidepth,keetha2025mapanything}.
These routes provide metric information through the acquisition setup or
metric-labelled training data. For a pretrained geometry model, however, a
practical goal is to recover the physical unit without introducing new
sensing requirements or training an additional metric predictor. This
raises the question we study: \textbf{can the content of a scene provide enough
information to recover the metric scale of its reconstruction?}
}}

\rev{\chg{Our motivation is that uncertainty about individual object sizes need not
imply equally large uncertainty about a scene's scale. Familiar objects such
as cars, doors, and pedestrians have plausible physical size ranges, even
when their exact dimensions are unknown. Each object alone provides only a
coarse reference. Within a reconstruction, however, these references
constrain the same scale: a factor that makes the car plausible must also
make the door and the pedestrian plausible. A scale acceptable for one
object may be ruled out by another (\Cref{fig:teaser}(b)). Their joint
compatibility can therefore narrow the range of plausible scene scales
without requiring any individual object to have an exactly known size. The
opportunity lies in using uncertain references to constrain shared
scale.}}

We examine this premise through complementary theoretical and empirical
evidence. With exact relative heights, averaging point-prior scale
observations yields a decreasing mean-square error bound when their errors
are zero-mean, pairwise uncorrelated, and uniformly bounded in variance.
A controlled study then evaluates real VLM height intervals while keeping
relative heights exact, separating semantic uncertainty from errors in
reconstructed geometry (\Cref{fig:empirical_validation}). Across four
dataset subsets, increasing the observation budget reduces scale error
(\Cref{tab:empirical_joint_scale_results}). These results motivate combining
multiple uncertain priors in a reconstruction, whose measured object
heights introduce an additional source of error.

\rev{\chg{Building on this motivation, we present UniMetric
(\Cref{subsec:unimetric}), a universal and training-free metric scale
recovery framework for geometry foundation models. \textbf{Metric Prior
Acquisition} queries a pretrained vision--language model (VLM) to identify
familiar objects and provide their plausible height ranges.
\textbf{Semantic--Geometric Alignment} measures each object in the
reconstruction along the support direction estimated for its frame and
converts its height range into an interval of admissible scales.
\textbf{Global Scale Recovery} combines these intervals across objects and
frames by minimizing a convex objective whose zero set is exactly their
common intersection. The recovered scalar uniformly rescales the pointmaps
and camera translations, preserving relative geometry without modifying the
underlying GFM.}}

\rev{\chg{Evaluating scale recovery requires retaining the scale error rather than
removing it during alignment. A protocol that fits scale to ground
truth~\citep{umeyama1991least,eigen2014depth} cannot reveal whether a
prediction already has the correct physical unit. We therefore introduce
MetricBench (\Cref{subsec:metricbench}), comprising twelve dataset subsets
across driving, roadside, and embodied scenarios with sensor-derived metric
references. Its protocol evaluates depth, 3D reconstruction, and camera
trajectories where applicable, allowing rigid registration when needed but
never ground-truth scale fitting. Across nine GFMs, UniMetric reduces mean
depth absolute relative error (Abs Rel), reconstruction Chamfer distance
(CD), and mean absolute trajectory error (ATE Mean) by 54.2\%, 83.8\%, and
65.6\%, respectively, relative to uncorrected predictions. Every evaluated
model improves its suite-averaged performance on all three tasks
(\Cref{tab:exps_summary_suite}), demonstrating the practical benefit of a
shared scale correction across scene geometry and camera motion.
The per-subset depth, CD, and ATE comparisons
(\Cref{tab:exps_depth_main,tab:exps_3d_reconstruction,tab:exps_ego_pose})
test this transfer directly: each sequence uses the same estimated scale
for all three outputs, without fitting a separate factor for each task.}}

Our contributions are threefold:
\begin{enumerate}[leftmargin=*,nosep,topsep=3pt]
  \item \rev{\chg{We recast metric scale recovery as joint estimation from uncertain
  semantic size priors. We establish conditions for mean-square consistency
  with point priors and validate the benefit of joint estimation with real
  VLM intervals with exact relative object heights.}}
  \item \rev{\chg{We propose UniMetric, a universal and training-free framework that
  reconciles semantic size intervals with a GFM reconstruction and applies
  one recovered scale to its geometry and camera translations.}}
  \item \rev{\chg{We introduce MetricBench, a cross-platform benchmark that measures
  metric accuracy across twelve sensor-referenced dataset subsets. Paired
  comparisons across nine GFMs demonstrate improvements in depth,
  reconstruction, and camera motion, while identifying cases where a
  shared scale alone is insufficient.}}
\end{enumerate}

\section{Related Work}
\label{sec:related_work}
\paragraph{Visual Geometry Foundation Models.}
\rev{Building on the pairwise pointmaps of DUSt3R~\citep{wang2024dust3r},
VGGT~\citep{wang2025vggt}, $\pi^3$~\citep{wang2025pi}, and Depth Anything
3~\citep{lin2025depth} infer structure and cameras for many views in one
forward pass, and later work extends the formulation to longer, streaming,
and dynamic
inputs~\citep{wang2026vggtomega,xie2026scal3r,burzio2026dejaview,lan2025stream3r,chen2026geometric,luo20264rc}.
The quality of relative geometry and the accuracy of its physical scale
are distinct properties. Evaluation with per-image median
scaling~\citep{eigen2014depth} or similarity alignment~\citep{umeyama1991least}
removes scale discrepancies before measuring error. MetricBench instead
retains them across depth, reconstruction, and camera motion
(\Cref{subsec:metricbench}), while UniMetric supplies a shared scale without
altering the relative geometry.}

\paragraph{Metric Scale from Sensors and Supervision.}
\rev{Metric scale is conventionally observed by calibrated stereo, depth, or
inertial sensors~\citep{campos2021orbslam3,qin2018vinsmono,salih2026vidar}
or by a known camera height~\citep{tian2021scale}.
Height-based lifting~\citep{yang2023bevheight,yang2025bevheightpp}
and ground-aware embeddings~\citep{yang2024monogae} also incorporate
scene geometry into monocular 3D detection.
Metric scale can also be learned from metric labels in single-image
depth~\citep{bhat2023zoedepth,yin2023metric3d,hu2024metric3dv2,piccinelli2024unidepth,bochkovskii2025depthpro}
and its video and multi-view
extensions~\citep{chen2025videodepthanything,keetha2025mapanything};
language-derived scale predictors~\citep{zeng2024rsa,wu2025vgld} are
trained in the same way. The first route requires hardware; the second
inherits the scale statistics of its training data. UniMetric uses neither:
it combines the reconstruction of an existing model with the size knowledge
of an existing vision--language model, without training. \chg{Object size
priors have previously scaled filtering-based monocular SLAM, using height
priors for a small set of predefined detector
classes~\citep{sucar2017probabilistic}; UniMetric instead obtains
open-vocabulary height intervals from a VLM and combines them with a GFM's
reconstructed geometry. The metric-depth calibration comparison in
\Cref{tab:ablation} evaluates these alternative sources of scale using the
same backbone predictions.}}

\section{Method}
\label{sec:method}
\rev{Given an RGB sequence $\mathcal{I}=\{I_t\}_{t=1}^{T}$ and a pretrained
geometry foundation model \chg{(GFM)} $\mathcal{G}$, we estimate a positive scale
$\hat{s}$ that converts reconstructed points and camera translations from a
shared unknown unit into metres. UniMetric obtains size priors from RGB
content and measures their geometric counterparts in the GFM reconstruction;
the resulting correction preserves relative geometry.}

\subsection{Metric Scale Recovery from Uncertain Semantic Size Priors}
\label{subsec:semantic_scale_principle}

\rev{\chg{We study how aggregating uncertain object-size priors affects scale
accuracy when relative object heights are exact. We first analyze point
estimates, then evaluate real VLM height intervals in a controlled study.}}

\paragraph{Mathematical analysis of joint scale estimation.}
\rev{\chg{For observation $i$, let $H_i>0$ be the physical height in metres and
$h_i>0$ its relative height, i.e., the height in the arbitrary unit of the
reconstruction. They satisfy $H_i=s_\star h_i$, where $s_\star>0$ is the
shared scale to recover. A semantic height estimate $\widetilde H_i$
therefore provides a scale observation}}
\begin{equation}
y_i=\frac{\widetilde H_i}{h_i}=s_\star+\epsilon_i,
\qquad
\epsilon_i=\frac{\widetilde H_i-H_i}{h_i}.
\label{eq:semantic_scale_observation}
\end{equation}
\rev{\chg{Although objects differ in height, these ratios estimate the same quantity.
We combine $N$ observations using their mean,
$\bar{s}_N=N^{-1}\sum_{i=1}^{N}y_i$, with expectations taken over semantic
errors while the relative heights are held fixed.}}

\noindent\textbf{Proposition 1 (Mean-square consistency of joint scale estimation).}
\emph{Suppose the errors $\{\epsilon_i\}_{i\geq1}$ have zero means, are
pairwise uncorrelated, and satisfy
$\operatorname{Var}(\epsilon_i)\leq\sigma^2<\infty$ for all $i$. Then}
\begin{equation}
\mathbb{E}\!\left[(\bar{s}_N-s_\star)^2\right]
=\frac{1}{N^2}\sum_{i=1}^{N}\operatorname{Var}(\epsilon_i)
\leq\frac{\sigma^2}{N},
\label{eq:joint_scale_mse}
\end{equation}
\emph{and $\bar{s}_N$ converges in mean square to $s_\star$} \rev{\chg{(proof in
\Cref{app:mean_square_consistency}).}}

\rev{\chg{Under these assumptions, the error bound decreases as observations
accumulate, even when individual priors remain noisy. This analysis
concerns point priors, the special case of UniMetric's interval objective
in which every interval collapses to a point (\Cref{app:interval_objective}); we next
evaluate the interval estimator with real VLM priors.}}

\begin{figure}[!t]
\centering
\setlength{\abovecaptionskip}{1pt}
\includegraphics[width=\textwidth,trim=0bp 44bp 0bp 34bp,clip]{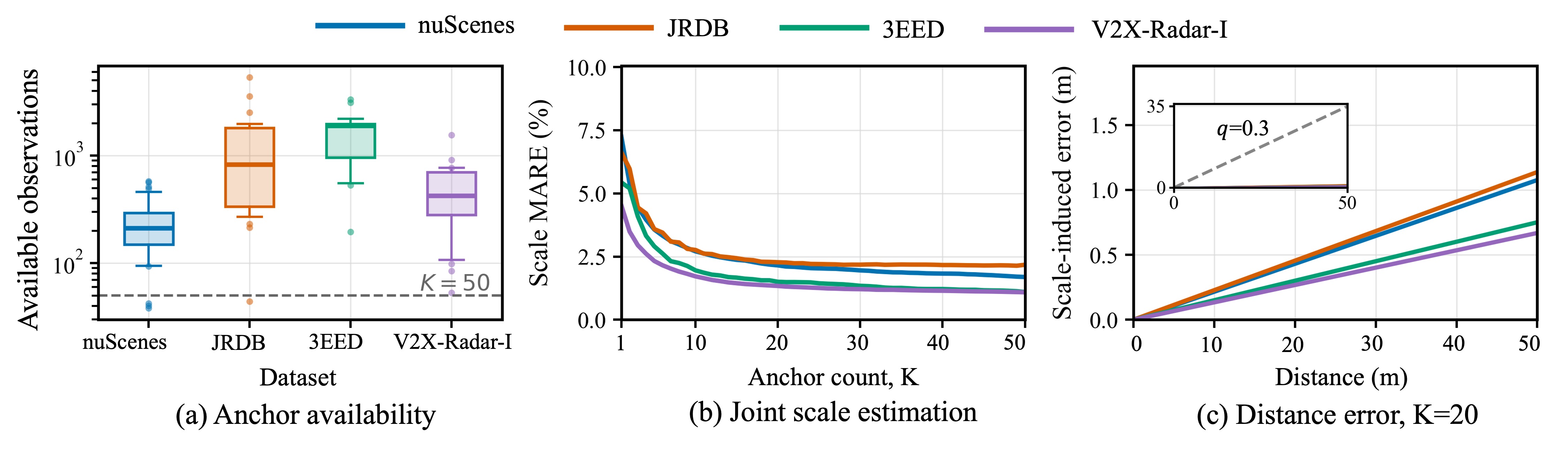}
\caption{\chg{\textbf{Joint scale estimation with real VLM height priors and
exact relative heights.} (a) Available frame--object observations per
sequence; the dashed line marks the budget of 50 observations. (b) Mean
absolute relative scale error as the observation budget increases.
(c) Scale-induced distance error computed from the results in (b) at
$K=20$; the inset shows the uncorrected reference.}}
\label{fig:empirical_validation}
\end{figure}

\paragraph{Controlled evaluation with VLM height intervals.}
\rev{\chg{We construct relative heights from annotated object heights as $h_i=qH_i$,
with $q=0.3$, so that the true recovery factor is $s_\star=1/q$. The VLM
receives an image with the target object marked and its annotated category,
and provides a height interval and confidence without access to $H_i$ or
$q$. These priors are combined with the relative
heights using UniMetric's weighted interval estimator
(\Cref{subsec:unimetric}).}}

\rev{\chg{The study covers 193 sequences from the MetricBench subsets of
nuScenes~\citep{caesar2020nuscenes}, JRDB~\citep{martin2021jrdb},
3EED~\citep{li20253eed}, and V2X-Radar-I~\citep{yangv2x}.
We vary the budget $K$ from 1 to 50
frame--object observations per sequence, keeping the same sequence set at
every budget and using all available observations when fewer than $K$
exist. We report mean absolute relative scale error
(MARE), obtained by averaging $100\,|\hat{s}/s_\star-1|$ over sampling
trials and sequences within each subset. \Cref{tab:empirical_joint_scale_results} reports representative
budgets and sequence-bootstrap $95\%$ confidence intervals; the complete
query and sampling protocol is given in \Cref{app:empirical_validation}.}}

\begin{table}[!htbp]
\centering
\caption{\textbf{Scale recovery with different observation budgets.}
$K$ is the maximum number of frame--object observations per sequence.
Mean absolute relative scale error (MARE) is reported in percent
(lower is better). Brackets give sequence-bootstrap $95\%$ confidence
intervals. Reduction is relative to $K=1$.}
\label{tab:empirical_joint_scale_results}
\unimetrictablefont
\setlength{\tabcolsep}{3pt}
\begin{tabular*}{\linewidth}{@{\extracolsep{\fill}}lrrrrr@{}}
\toprule
Dataset & $K=1$ & $K=10$ & $K=20$ & $K=50$ [95\% CI]
& \makecell{Reduction\\(\%)} \\
\midrule
nuScenes    & 7.245 & 2.707 & 2.152 & 1.691 [1.178, 2.279] & 76.7 \\
JRDB        & 6.627 & 2.761 & 2.275 & 2.175 [1.762, 2.576] & 67.2 \\
3EED        & 5.435 & 1.954 & 1.501 & 1.097 [0.768, 1.441] & 79.8 \\
V2X-Radar-I & 4.526 & 1.717 & 1.337 & 1.086 [0.886, 1.310] & 76.0 \\
\bottomrule
\end{tabular*}
\end{table}

\paragraph{Effect of the observation budget.}
Increasing the budget from $K=1$ to $50$ lowers MARE from
4.526--7.245\% to 1.086--2.175\% across the four subsets
(\Cref{fig:empirical_validation,tab:empirical_joint_scale_results}),
corresponding to relative reductions of 67.2--79.8\%.
Much of the reduction is already obtained at $K=10$, with further gains
at $K=20$ and $50$. The improvement from $20$ to $50$ observations varies
across subsets: MARE decreases from 2.275\% to 2.175\% on JRDB and from
1.501\% to 1.097\% on 3EED. Thus, additional observations improve the
aggregate estimate, but the benefit depends on the available priors and
their errors.

This controlled study isolates uncertainty in semantic priors by using
exact relative heights. The observation budget counts frame--object
measurements, including repeated views of the same object.
The distance-error curves in \Cref{fig:empirical_validation}(c) translate
the measured relative scale errors into their consequences at different
distances; they are not additional GFM reconstruction measurements.
We next replace the annotation-derived
relative heights with measurements from GFM reconstructions, and evaluate
the resulting metric geometry in \Cref{sec:exps}.

\subsection{UniMetric}
\label{subsec:unimetric}

\begin{figure}[!t]
\centering
\includegraphics[width=\textwidth]{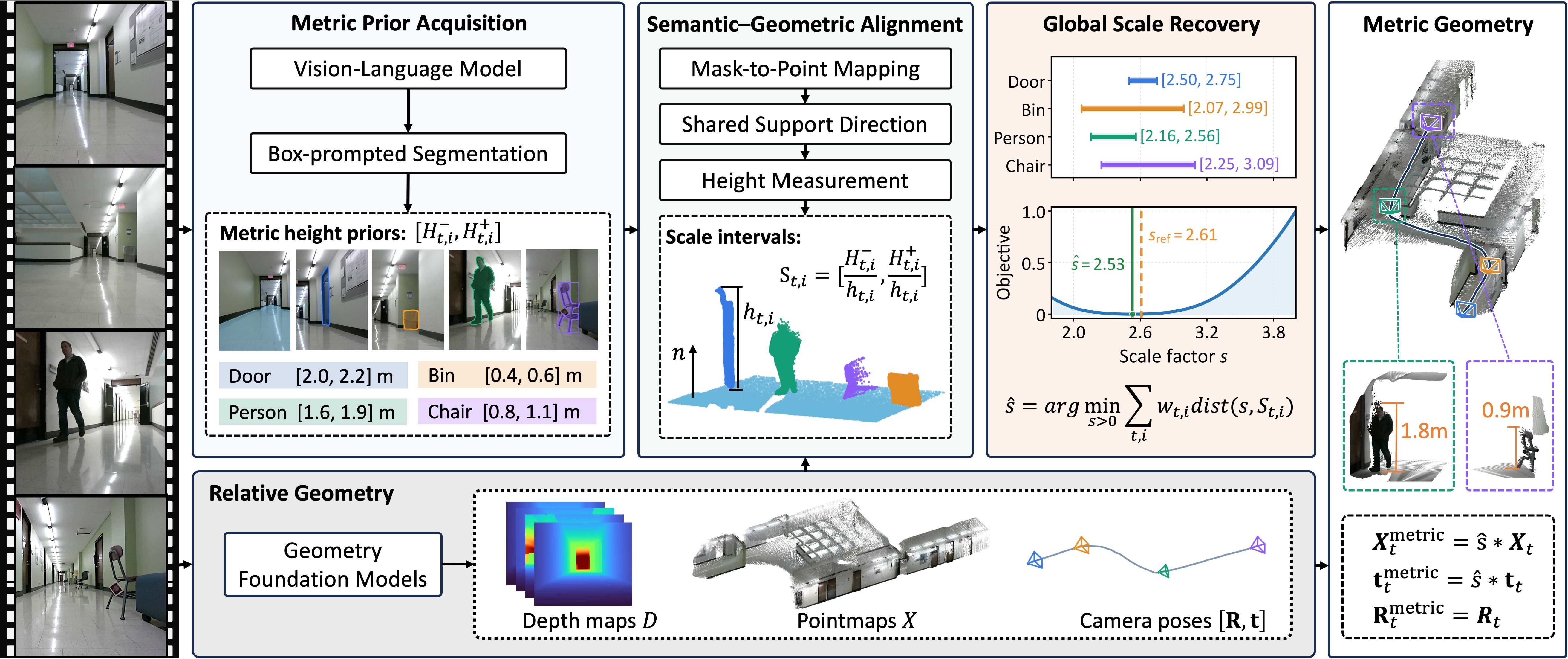}
\caption{\chg{\textbf{Overview of UniMetric.} Metric Prior Acquisition obtains
object height intervals and masks from RGB images. Semantic--Geometric
Alignment measures reconstructed heights along frame-specific support
directions and converts the priors into scale intervals. Global Scale
Recovery combines these intervals through a weighted convex objective,
yielding a shared scale for reconstructed geometry and camera translations.}}
\label{fig:framework}
\end{figure}

\paragraph{Overall Architecture.}
\rev{\chg{UniMetric combines semantic size priors with the relative geometry that
a GFM reconstructs from the same RGB sequence $\mathcal{I}=\{I_t\}_{t=1}^{T}$
(\Cref{fig:framework}). The GFM provides camera-frame pointmaps
$X_t\in\mathbb{R}^{H\times W\times 3}$, depth maps $D_t$, and world-to-camera
poses $[R_t\mid\mathbf{t}_t]$, whose points and translations we assume to
share one unknown unit across frames. The three modules below turn object
height priors into scale intervals and estimate the scalar that converts the
reconstruction to metric units.}}

\paragraph{Metric Prior Acquisition.}
\rev{\chg{A pretrained vision--language model $\mathcal{V}$, Seed1.8~\citep{seed2026seed18}
in our experiments, identifies familiar
objects and provides their plausible physical heights. For each frame
$I_t$, it returns candidate anchors and the dominant support surface:}}
\begin{equation}
\Big(\big\{(c_{t,i},\,b_{t,i},\,[H^-_{t,i},H^+_{t,i}],\,q_{t,i})\big\}_{i=1}^{N_t},\;
b^{s}_t\Big)=\mathcal{V}(I_t).
\label{eq:vlm}
\end{equation}
\rev{\chg{Here, $c_{t,i}$ and $b_{t,i}$ denote the object category and bounding box,
$[H^-_{t,i},H^+_{t,i}]$ is its height interval in metres,
$q_{t,i}\in[0,1]$ is the reported confidence, and $b^{s}_t$ localizes the
support surface. Retaining both height bounds preserves uncertainty about
the instance's dimensions rather than assigning a single nominal size
(prompt templates in \Cref{app:prompt_templates}).}}

\rev{\chg{We retain non-truncated anchors standing upright on the selected support
surface and remove duplicate detections within each frame. Box-prompted
Segmentation then converts the retained object and support boxes into pixel
masks:}}
\begin{equation}
M_{t,i}=\mathcal{S}(I_t,\,b_{t,i}),
\qquad
M^{s}_t=\mathcal{S}(I_t,\,b^{s}_t),
\label{eq:sam}
\end{equation}
\rev{\chg{where $\mathcal{S}$ is a pretrained segmentation
model~\citep{ravi2024sam2}. This module therefore supplies both metric
height intervals and the image regions needed to associate them with
reconstructed objects.}}

\paragraph{Semantic--Geometric Alignment.}
\rev{\chg{This module connects the semantic and geometric branches through the three
operations shown in \Cref{fig:framework}. Its goal is to measure the
reconstructed extent corresponding to each height prior and express their
relationship as a constraint on scale.}}

\rev{\chg{\textit{Mask-to-Point Mapping.} The object and support masks select the
corresponding samples from each pointmap. Using the predicted camera pose,
we transform these samples into the common reconstruction frame:}}
\begin{equation}
\begin{aligned}
\mathcal{P}_{t,i}&=\big\{R_t^{\top}\big(X_t(\mathbf{u})-\mathbf{t}_t\big)\;\big|\;M_{t,i}(\mathbf{u})=1\big\},\\
\mathcal{P}^{s}_t&=\big\{R_t^{\top}\big(X_t(\mathbf{u})-\mathbf{t}_t\big)\;\big|\;M^{s}_t(\mathbf{u})=1\big\},
\end{aligned}
\label{eq:lift}
\end{equation}
\rev{\chg{where $\mathbf{u}$ indexes pixels. The resulting point sets associate each
semantic anchor and its support surface with their reconstructed geometry.}}

\rev{\chg{\textit{Shared Support Direction.} To measure height without assuming that
a camera axis is vertical, we fit a plane to the support points and use its
unit normal:}}
\begin{equation}
\mathbf{n}_t=\operatorname{PlaneFit}\big(\mathcal{P}^{s}_t\big).
\label{eq:normal}
\end{equation}
\rev{\chg{This direction is shared by the anchors within frame $t$, but estimated
independently across frames. The method therefore does not impose a single
support plane on the entire sequence.}}

\rev{\chg{\textit{Height Measurement.} For anchors whose height axes align with
$\mathbf{n}_t$, the reconstructed height is the extent of their points
along this direction:}}
\begin{equation}
h_{t,i}=\max_{\mathbf{p}\in\mathcal{P}_{t,i}}\mathbf{n}_t^{\top}\mathbf{p}
-\min_{\mathbf{p}\in\mathcal{P}_{t,i}}\mathbf{n}_t^{\top}\mathbf{p}.
\label{eq:height}
\end{equation}
\rev{\chg{The support plane supplies only the direction; both endpoints come from
the object's own points, which must cover its vertical extent. The
resulting $h_{t,i}$ and the semantic prior describe the same physical
height in reconstruction and metric units, respectively. For $h_{t,i}>0$,
a scale $s$ is admissible when the rescaled height lies within the prior:}}
\begin{equation}
s\,h_{t,i}\in\big[H^-_{t,i},\,H^+_{t,i}\big]
\;\Longleftrightarrow\;
s\in S_{t,i}:=\Big[\frac{H^-_{t,i}}{h_{t,i}},\;\frac{H^+_{t,i}}{h_{t,i}}\Big].
\label{eq:interval}
\end{equation}
\rev{\chg{Each retained frame--object observation thus constrains the same
sequence-level conversion to metres.}}

\paragraph{Global Scale Recovery.}
\rev{\chg{We collect the scale intervals across objects and frames and jointly
estimate a single scalar (\Cref{fig:framework}). Each observation receives a weight}}
\begin{equation}
w_{t,i}=q_{t,i}\,\alpha_{c_{t,i}},
\label{eq:weight}
\end{equation}
\rev{\chg{where $\alpha_c\in[0,1]$ is a category-specific reliability weight, and
$\mathcal{A}$ denotes the retained observations with positive weights. A
candidate scale is penalized only outside an interval, by its distance to the
nearest bound: for $S=[a,b]$,}}
\begin{equation}
\operatorname{dist}(s,S)=\max\{a-s,\;0,\;s-b\},
\label{eq:dist}
\end{equation}
\rev{\chg{and minimize the weighted sum of squared interval distances:}}
\begin{equation}
\hat{s}\in\argmin_{s>0}\;J(s),
\qquad
J(s)=\sum_{(t,i)\in\mathcal{A}} w_{t,i}\,\operatorname{dist}(s,S_{t,i})^{2}.
\label{eq:objective}
\end{equation}
\rev{\chg{The weighted interval objective is convex and piecewise quadratic.
Its minimizers coincide with the common intersection when one exists;
otherwise, it admits a unique minimizer. We solve this one-dimensional
problem by golden-section search~\citep{kiefer1953sequential} over the interval hull
(\Cref{app:interval_objective}).}}

\rev{\chg{When all intervals collapse to points and the weights are equal,
\Cref{eq:objective} reduces to the mean estimator analyzed in
\Cref{subsec:semantic_scale_principle}.}}

\rev{\chg{Finally, we apply the recovered scale to the reconstruction:}}
\begin{equation}
X^{m}_t=\hat{s}\,X_t,
\qquad
\mathbf{t}^{m}_t=\hat{s}\,\mathbf{t}_t,
\qquad
R^{m}_t=R_t,
\qquad t=1,\dots,T.
\label{eq:apply}
\end{equation}
\rev{\chg{The same correction applies to scene structure and camera motion, leaving
rotations, intrinsics, and relative geometry unchanged. The associated
depth maps inherit the scaling, $D^{m}_t=\hat{s}\,D_t$, through their
camera-frame pointmaps.}}

\section{Experiments}
\label{sec:exps}
The controlled study in \Cref{subsec:semantic_scale_principle} examines
semantic priors with exact relative heights. We now evaluate UniMetric on
GFM reconstructions, where object measurements and camera motion also
contain geometric errors. MetricBench provides sensor-derived references
for testing whether one recovered scale improves these outputs together.

\subsection{MetricBench}
\label{subsec:metricbench}

\rev{We introduce \textbf{MetricBench}, a cross-platform benchmark for
evaluating metric geometry across diverse Physical AI settings.
It brings together \textbf{twelve dataset subsets comprising 694 sequences
and 51{,}339 frames}, organized into \textbf{driving-and-infrastructure} and
\textbf{embodied-AI} suites (\Cref{tab:bench_summary}). \chg{The subsets are
drawn from KITTI~\citep{geiger2012we,geiger2013vision}, nuScenes~\citep{caesar2020nuscenes},
Waymo~\citep{sun2020scalability}, Argoverse~\citep{chang2019argoverse},
V2X-Radar-I~\citep{yangv2x}, DAIR-V2X-I~\citep{yu2022dair},
CEAR~\citep{zhu2024cear}, JRDB~\citep{martin2021jrdb},
FusionPortable~\citep{jiao2022fusionportable}, DROID~\citep{khazatsky2024droid},
Humanoid Everyday~\citep{zhao2025humanoid}, and 3EED~\citep{li20253eed}.}
MetricBench combines
broad platform and scene coverage with a unified evaluation of depth, 3D
reconstruction, and ego pose. All reported results refer to these fixed
evaluation subsets; the sequence selection and sampling details are given
in \Cref{app:bench_subsets}.}

\begin{table}[!htbp]
\centering
\caption{\textbf{MetricBench: platforms, sensor references, and physical scale.}
Counts refer to the evaluated subsets. Reference identifies the sensor
source of geometry; trajectory references use dataset metric poses.
Median depth is computed from valid projected sensor samples in metres.
Seq.: sequences; UGV: unmanned
ground vehicle. Subset construction and complete scale statistics are
given in \Cref{app:bench}.}
\label{tab:bench_summary}
\label{tab:bench_detail}
\unimetrictablefont
\setlength{\tabcolsep}{3pt}
\renewcommand{\arraystretch}{1.05}
\begin{tabularx}{\textwidth}{@{}l>{\raggedright\arraybackslash}Xlrrr@{}}
\toprule
\textbf{Dataset} & \textbf{Hardware platform(s)} & \textbf{Reference}
  & \textbf{Seq.} & \textbf{Frames} & \shortstack{\textbf{Median}\\\textbf{depth (m)}} \\
\midrule
\multicolumn{6}{@{}l}{\textit{Driving and infrastructure}} \\
KITTI & passenger vehicle & LiDAR & 96 & 5{,}211 & 11.755 \\
nuScenes & passenger vehicle & LiDAR & 116 & 4{,}627 & 11.585 \\
Waymo & passenger vehicle & LiDAR & 149 & 5{,}951 & 19.095 \\
Argoverse & passenger vehicle & LiDAR & 70 & 6{,}613 & 21.215 \\
V2X-Radar-I & roadside unit & LiDAR & 30 & 1{,}092 & 36.865 \\
DAIR-V2X-I & roadside unit & LiDAR & 86 & 3{,}704 & 33.655 \\
\midrule
\multicolumn{6}{@{}l}{\textit{Embodied AI}} \\
CEAR & quadruped & LiDAR & 20 & 5{,}670 & 2.895 \\
JRDB & wheeled robot & LiDAR & 27 & 3{,}729 & 4.655 \\
FusionPortable & handheld/legged/UGV & LiDAR & 5 & 3{,}847 & 6.505 \\
DROID & Franka arm/external camera & depth camera & 45 & 3{,}102 & 0.885 \\
Humanoid Everyday & G1 humanoid & depth camera & 30 & 3{,}949 & 0.695 \\
3EED & drone/quadruped & LiDAR & 20 & 3{,}844 & 10.035 \\
\midrule
\textbf{Total} & & & \textbf{694} & \textbf{51{,}339} & \\
\bottomrule
\end{tabularx}
\end{table}

\paragraph{Platform and scene coverage.}
\rev{The benchmark spans vehicles, roadside units, wheeled and quadruped
robots, robotic arms, humanoids, drones, and handheld rigs, covering diverse
viewpoints and acquisition setups. Its two suites encompass driving and
roadside perception alongside indoor and outdoor embodied scenes, including
close-range manipulation and aerial observations. Beyond platform diversity,
the subsets vary substantially in physical scale: median reference depth
ranges from 0.695\,m on Humanoid Everyday to 36.865\,m on V2X-Radar-I,
representing more than a $50\times$ variation in characteristic scene depth.
The 90th-percentile reference depth reaches 81.245\,m on DAIR-V2X-I,
while mean trajectory length per sequence for moving cameras ranges from
1.305\,m on Humanoid Everyday to 251.145\,m on FusionPortable
(\Cref{tab:bench_scale}). These differences allow us to examine metric
accuracy across both scene extent and camera travel distance.}

\paragraph{Reference data and evaluation.}
\rev{Each evaluated frame contains a matched RGB image, a calibrated metric
point cloud, and camera metadata. Reference geometry comes from LiDAR or
depth cameras, with dataset-provided metric poses supplying trajectory
references where applicable. These measurements provide the physical units
against which reconstructed depth, geometry, and camera motion are evaluated.
The main comparisons permit rigid $SE(3)$ registration where coordinate
frames differ but do not fit scale to ground truth, retaining errors in
both physical scale and geometry. The two static roadside subsets are
excluded from ego-pose evaluation. DROID retains its stationary external
camera as a case with near-zero reference motion.}

\subsection{Experimental Settings}
\label{subsec:exp_setup}

\paragraph{Models and comparison settings.}
\rev{We evaluate UniMetric with nine geometry foundation models:
\chg{VGGT~\citep{wang2025vggt}, VGGT-$\Omega$~\citep{wang2026vggtomega},
$\pi^3$~\citep{wang2025pi}, Depth Anything~3 (DA-3;~\citealp{lin2025depth}),
Scal3R~\citep{xie2026scal3r}, LingBot-Map~\citep{chen2026geometric},
4RC~\citep{luo20264rc}, D\'ej\`a View (DVLT;~\citealp{burzio2026dejaview}), and
STream3R~\citep{lan2025stream3r}.}
Each \chg{GFM} uses its released weights without additional training. For
each sequence, \emph{None} and \emph{UniMetric} share the same \chg{GFM}
predictions and differ only in the scale applied to depth, reconstructed
points, and camera translations. UniMetric estimates one factor per
sequence and uses it for all three tasks. The VLM \chg{(Seed1.8)} and segmentation
model \chg{(SAM~2.1)} are shared across \chg{GFMs}, while geometric filtering follows the dataset--\chg{GFM}
configuration. The CEAR and FusionPortable prompts additionally include
category-size tables; VGGT-$\Omega$ uses dataset camera intrinsics on 3EED.
Complete prompts, inference settings, and optimization parameters are
provided in \Cref{app:impl,app:bench_datasets}.}

\paragraph{Metrics and aggregation.}
We report absolute relative depth error (Abs Rel) and threshold accuracy
at 1.25 ($\delta_1$, \%;~\citealp{eigen2014depth}). For reconstruction,
Chamfer distance (CD) averages the two directed nearest-neighbour distances
between accumulated predicted and reference point clouds, in metres.
For ego pose, ATE Mean is the mean Euclidean camera-position error after
rigid alignment of each sequence, with scale fixed to one
\citep{sturm2012benchmark}. Depth errors are averaged over valid frames;
CD is averaged over sequences; ATE Mean pools position errors over valid
frames after sequence-wise alignment. Suite summaries weight subsets equally.
Error reductions compare the means before and after correction over the
stated model--subset pairs. Full definitions are given in
\Cref{app:bench_metrics}. The Oracle comparison in \Cref{subsec:ablations}
is explicitly distinguished by its use of ground-truth scale fitting.

\subsection{Quantitative Results}
\label{subsec:main_results}

We first examine metric depth across all twelve subsets, then test whether
the same scale transfers to accumulated 3D geometry and camera trajectories.
The paired comparisons isolate scale correction from changes to the
underlying GFM predictions.

\begin{table}[t]
\centering
\caption{\textbf{Metric depth estimation on MetricBench.}
Absolute relative depth error (Abs Rel, lower is better) before (None) and
after UniMetric correction, without ground-truth scale alignment. Bold
marks the better result within each None/UniMetric pair. DA-3 denotes
Depth Anything 3.}
\label{tab:exps_depth_main}
\unimetrictablefont
\setlength{\tabcolsep}{2pt}
\renewcommand{\arraystretch}{0.7}
\begin{tabular*}{\linewidth}{@{\extracolsep{\fill}}ll*{12}{c}@{}}
\toprule
\multirow{2}{*}{\textbf{Model}}
& \multirow{2}{*}{\shortstack{\textbf{Scale}\\\textbf{Recovery}}}
& \multicolumn{6}{c}{\textbf{Autonomous Driving and Infrastructure}}
& \multicolumn{6}{c}{\textbf{Embodied AI}} \\
\cmidrule(lr){3-8}\cmidrule(lr){9-14}
&
& \textbf{KITTI}
& \textbf{nuScenes}
& \textbf{Waymo}
& \textbf{Argoverse}
& \shortstack[c]{\textbf{V2X-}\\\textbf{Radar-I}}
& \shortstack[c]{\textbf{DAIR-}\\\textbf{V2X-I}}
& \textbf{CEAR}
& \textbf{JRDB}
& \shortstack[c]{\textbf{Fusion}\\\textbf{Portable}}
& \textbf{DROID}
& \shortstack[c]{\textbf{Humanoid}\\\textbf{Everyday}}
& \textbf{3EED} \\
\midrule
\multirow{2}{*}{VGGT}
& None & 0.974 & 0.983 & 0.983 & 0.974 & 0.943 & 0.965 & 0.850 & 0.805 & 0.949 & 0.439 & 0.359 & 0.936 \\
& \textbf{UniMetric} & \textbf{0.722} & \textbf{0.316} & \textbf{0.311} & \textbf{0.311} & \textbf{0.392} & \textbf{0.866} & \textbf{0.551} & \textbf{0.520} & \textbf{0.676} & \textbf{0.332} & \textbf{0.335} & \textbf{0.534} \\
\midrule
\multirow{2}{*}{VGGT-$\Omega$}
& None & 0.971 & 0.979 & 0.983 & 0.972 & 0.949 & 0.966 & 0.786 & 0.790 & 0.885 & 0.294 & 0.311 & 0.937 \\
& \textbf{UniMetric} & \textbf{0.347} & \textbf{0.152} & \textbf{0.154} & \textbf{0.244} & \textbf{0.123} & \textbf{0.542} & \textbf{0.538} & \textbf{0.312} & \textbf{0.331} & \textbf{0.284} & \textbf{0.221} & \textbf{0.343} \\
\midrule
\multirow{2}{*}{$\pi^3$}
& None & 0.190 & 0.297 & 0.165 & 0.282 & 0.341 & 0.704 & 0.328 & 0.310 & 0.330 & 0.315 & 1.009 & 0.354 \\
& \textbf{UniMetric} & \textbf{0.188} & \textbf{0.098} & \textbf{0.095} & \textbf{0.140} & \textbf{0.147} & \textbf{0.471} & \textbf{0.322} & \textbf{0.253} & \textbf{0.219} & \textbf{0.221} & \textbf{0.480} & \textbf{0.295} \\
\midrule
\multirow{2}{*}{DA-3}
& None & 0.967 & 0.948 & 0.962 & 0.937 & 0.922 & 0.956 & 0.754 & 0.788 & 0.894 & 0.335 & 0.482 & 0.915 \\
& \textbf{UniMetric} & \textbf{0.372} & \textbf{0.331} & \textbf{0.162} & \textbf{0.425} & \textbf{0.168} & \textbf{0.531} & \textbf{0.406} & \textbf{0.345} & \textbf{0.358} & \textbf{0.327} & \textbf{0.364} & \textbf{0.436} \\
\midrule
\multirow{2}{*}{Scal3R}
& None & 0.966 & 0.971 & 0.974 & 0.959 & 0.938 & 0.967 & 0.763 & 0.770 & 0.904 & 0.338 & 0.320 & 0.898 \\
& \textbf{UniMetric} & \textbf{0.231} & \textbf{0.234} & \textbf{0.241} & \textbf{0.287} & \textbf{0.411} & \textbf{0.856} & \textbf{0.361} & \textbf{0.528} & \textbf{0.478} & \textbf{0.336} & \textbf{0.263} & \textbf{0.489} \\
\midrule
\multirow{2}{*}{LingBot-Map}
& None & 0.958 & 0.971 & 0.975 & 0.964 & 0.952 & 0.973 & 0.749 & 0.799 & 0.887 & 0.356 & 0.381 & 0.929 \\
& \textbf{UniMetric} & \textbf{0.244} & \textbf{0.272} & \textbf{0.325} & \textbf{0.343} & \textbf{0.351} & \textbf{0.658} & \textbf{0.283} & \textbf{0.338} & \textbf{0.403} & \textbf{0.297} & \textbf{0.282} & \textbf{0.347} \\
\midrule
\multirow{2}{*}{4RC}
& None & 0.973 & 0.979 & 0.985 & 0.976 & 0.938 & 0.960 & 0.777 & 0.789 & 0.906 & 0.324 & 0.436 & 0.932 \\
& \textbf{UniMetric} & \textbf{0.326} & \textbf{0.310} & \textbf{0.192} & \textbf{0.480} & \textbf{0.151} & \textbf{0.524} & \textbf{0.579} & \textbf{0.287} & \textbf{0.370} & \textbf{0.302} & \textbf{0.270} & \textbf{0.340} \\
\midrule
\multirow{2}{*}{DVLT}
& None & 0.957 & 0.968 & 0.974 & 0.962 & 0.881 & 0.926 & 0.672 & 0.569 & 0.867 & 1.001 & 0.722 & 0.898 \\
& \textbf{UniMetric} & \textbf{0.435} & \textbf{0.220} & \textbf{0.156} & \textbf{0.245} & \textbf{0.171} & \textbf{0.576} & \textbf{0.448} & \textbf{0.392} & \textbf{0.447} & \textbf{0.264} & \textbf{0.348} & \textbf{0.538} \\
\midrule
\multirow{2}{*}{STream3R}
& None & 0.960 & 0.971 & 0.970 & 0.961 & 0.956 & 0.970 & 0.719 & 0.796 & 0.858 & 0.421 & 0.379 & 0.932 \\
& \textbf{UniMetric} & \textbf{0.515} & \textbf{0.205} & \textbf{0.242} & \textbf{0.250} & \textbf{0.497} & \textbf{0.685} & \textbf{0.380} & \textbf{0.330} & \textbf{0.734} & \textbf{0.241} & \textbf{0.317} & \textbf{0.449} \\
\bottomrule
\end{tabular*}
\end{table}

\begin{table}[!htbp]
\centering
\caption{\textbf{Suite-averaged metric geometry on MetricBench.}
Metrics are averaged equally over applicable subsets; Average further
averages over the nine models. $\delta_1$ denotes depth accuracy within a
factor of 1.25 (\%), CD denotes Chamfer distance (m), and ATE denotes mean
absolute trajectory error (m). Roadside subsets are excluded from ATE.
Bold marks the better result within each None/UniMetric pair.}
\label{tab:exps_summary_suite}
\unimetrictablefont
\setlength{\tabcolsep}{4pt}
\renewcommand{\arraystretch}{0.7}
\begin{tabular*}{\linewidth}{@{\extracolsep{\fill}}ll*{8}{c}@{}}
\toprule
\multirow{2}{*}{\textbf{Model}}
& \multirow{2}{*}{\shortstack{\textbf{Scale}\\\textbf{Recovery}}}
& \multicolumn{4}{c}{\textbf{Autonomous Driving and Infrastructure}}
& \multicolumn{4}{c}{\textbf{Embodied AI}} \\
\cmidrule(lr){3-6}\cmidrule(lr){7-10}
&
& Abs Rel$\mathord{\downarrow}$ & $\delta_1(\%)\mathord{\uparrow}$ & CD (m)$\mathord{\downarrow}$ & ATE (m)$\mathord{\downarrow}$
& Abs Rel$\mathord{\downarrow}$ & $\delta_1(\%)\mathord{\uparrow}$ & CD (m)$\mathord{\downarrow}$ & ATE (m)$\mathord{\downarrow}$ \\
\midrule
\multirow{2}{*}{VGGT}
& None & 0.970 & 0.0 & 20.186 & 42.574 & 0.723 & 11.5 & 5.354 & 10.302 \\
& \textbf{UniMetric} & \textbf{0.486} & \textbf{48.9} & \textbf{4.038} & \textbf{14.878} & \textbf{0.491} & \textbf{35.5} & \textbf{1.189} & \textbf{6.406} \\
\midrule
\multirow{2}{*}{VGGT-$\Omega$}
& None & 0.970 & 0.0 & 19.167 & 42.332 & 0.667 & 13.8 & 4.763 & 9.885 \\
& \textbf{UniMetric} & \textbf{0.260} & \textbf{73.8} & \textbf{1.292} & \textbf{6.746} & \textbf{0.338} & \textbf{55.7} & \textbf{0.628} & \textbf{2.037} \\
\midrule
\multirow{2}{*}{$\pi^3$}
& None & 0.330 & 33.0 & 2.506 & 8.869 & 0.441 & 34.1 & 0.880 & 3.298 \\
& \textbf{UniMetric} & \textbf{0.190} & \textbf{80.6} & \textbf{1.018} & \textbf{4.417} & \textbf{0.298} & \textbf{57.7} & \textbf{0.691} & \textbf{1.969} \\
\midrule
\multirow{2}{*}{DA-3}
& None & 0.949 & 0.0 & 18.716 & 41.834 & 0.695 & 15.1 & 4.891 & 9.864 \\
& \textbf{UniMetric} & \textbf{0.331} & \textbf{55.8} & \textbf{2.239} & \textbf{14.285} & \textbf{0.373} & \textbf{46.2} & \textbf{0.918} & \textbf{3.708} \\
\midrule
\multirow{2}{*}{Scal3R}
& None & 0.962 & 0.0 & 19.892 & 42.162 & 0.665 & 12.5 & 5.189 & 9.947 \\
& \textbf{UniMetric} & \textbf{0.377} & \textbf{48.9} & \textbf{3.426} & \textbf{10.219} & \textbf{0.409} & \textbf{36.6} & \textbf{1.587} & \textbf{4.118} \\
\midrule
\multirow{2}{*}{LingBot-Map}
& None & 0.966 & 0.0 & 18.815 & 41.769 & 0.683 & 8.1 & 4.654 & 9.899 \\
& \textbf{UniMetric} & \textbf{0.365} & \textbf{37.2} & \textbf{2.724} & \textbf{11.071} & \textbf{0.325} & \textbf{40.6} & \textbf{0.679} & \textbf{3.311} \\
\midrule
\multirow{2}{*}{4RC}
& None & 0.969 & 0.0 & 20.052 & 42.941 & 0.694 & 11.4 & 4.944 & 9.721 \\
& \textbf{UniMetric} & \textbf{0.331} & \textbf{53.0} & \textbf{2.282} & \textbf{13.709} & \textbf{0.358} & \textbf{49.0} & \textbf{0.828} & \textbf{2.711} \\
\midrule
\multirow{2}{*}{DVLT}
& None & 0.945 & 0.1 & 18.399 & 42.667 & 0.788 & 11.1 & 4.865 & 9.920 \\
& \textbf{UniMetric} & \textbf{0.300} & \textbf{47.9} & \textbf{2.366} & \textbf{14.187} & \textbf{0.406} & \textbf{42.3} & \textbf{0.850} & \textbf{3.674} \\
\midrule
\multirow{2}{*}{STream3R}
& None & 0.965 & 0.0 & 19.508 & 42.558 & 0.684 & 6.0 & 4.020 & 9.506 \\
& \textbf{UniMetric} & \textbf{0.399} & \textbf{52.7} & \textbf{3.370} & \textbf{19.556} & \textbf{0.409} & \textbf{39.7} & \textbf{1.707} & \textbf{7.507} \\
\midrule
\multirow{2}{*}{\textbf{Average}}
& None & 0.892 & 3.7 & 17.471 & 38.634 & 0.671 & 13.7 & 4.396 & 9.149 \\
& \textbf{UniMetric} & \textbf{0.338} & \textbf{55.4} & \textbf{2.528} & \textbf{12.119} & \textbf{0.379} & \textbf{44.8} & \textbf{1.008} & \textbf{3.938} \\
\bottomrule
\end{tabular*}
\end{table}

\paragraph{Metric depth across models and scenes.}
\rev{UniMetric reduces Abs Rel in all 108 model--subset pairs in
\Cref{tab:exps_depth_main}, with a 54.2\% reduction in the mean error.
Improvements span both driving and infrastructure scenes and embodied-AI
environments. As summarized in \Cref{tab:exps_summary_suite}, mean Abs Rel
decreases from 0.892 to 0.338 in the driving suite and from 0.671 to 0.379
in the embodied suite. The larger reduction in driving scenes accompanies
their larger uncorrected metric errors.}

The effect also depends on the model's original scale. For $\pi^3$ on
KITTI, Abs Rel changes only from 0.190 to 0.188, whereas VGGT-$\Omega$
on nuScenes improves from 0.979 to 0.152. These paired results distinguish
cases with substantial scale discrepancies from those where the original
prediction already has a relatively small metric error. The complete
per-subset $\delta_1$ results and depth errors in metres are provided in
\Cref{app:results_depth}.

\paragraph{Shared correction across geometric tasks.}
\rev{The same estimated scale also improves reconstructed geometry and
camera trajectories. Every \chg{GFM} improves its suite-average performance
on all three tasks in both suites (\Cref{tab:exps_summary_suite}).
On the driving suite, mean CD decreases from 17.471 to
2.528\,m, and mean ATE decreases from 38.634 to 12.119\,m, corresponding to
reductions of 14.943 and 26.515\,m, respectively. These improvements show
that a scale estimated from semantic object priors can benefit multiple
outputs of the same reconstruction without task-specific scale fitting.
The per-subset results below examine how these gains vary and where
scale correction is insufficient.}

\paragraph{Metric 3D reconstruction.}
\label{subsec:reconstruction_results}
\Cref{tab:exps_3d_reconstruction} reports CD for all twelve subsets,
with reductions in 107 of 108 model--subset pairs.
The correction rescales both pointmaps and camera translations, preserving
the relative placement of frames when their points are accumulated.
For VGGT-$\Omega$ on FusionPortable, CD decreases from 15.848 to
0.653\,m; the same correction reduces it from 14.594 to 0.726\,m on
nuScenes. Improvements therefore extend from image-space depth to fused
3D geometry. The sole increase is for $\pi^3$ on DROID, from 0.090 to
0.097\,m, where the original reconstruction already has a small metric
error. This case also illustrates that improving depth does not guarantee
a reduction in every geometric metric.

\begin{table}[!htb]
\centering
\caption{\textbf{3D reconstruction across autonomous-driving, infrastructure, and embodied-AI benchmarks.}
We report CD (Chamfer distance) in metres; lower values are better. Bold marks the
better result within each None/UniMetric pair. DA-3 denotes \chg{Depth Anything 3}.
Results use the evaluation subsets described in \Cref{app:bench_subsets}.}

\unimetrictablefont
\setlength{\tabcolsep}{2pt}
\renewcommand{\arraystretch}{0.95}

\begin{tabular*}{\linewidth}{@{\extracolsep{\fill}}ll*{12}{c}@{}}
\toprule
\multirow{2}{*}{\textbf{Model}}
& \multirow{2}{*}{\shortstack{\textbf{Scale}\\\textbf{Recovery}}}
& \multicolumn{6}{c}{\textbf{Autonomous Driving and Infrastructure}}
& \multicolumn{6}{c}{\textbf{Embodied AI}} \\
\cmidrule(lr){3-8}\cmidrule(lr){9-14}
&
& \textbf{KITTI}
& \textbf{nuScenes}
& \textbf{Waymo}
& \textbf{Argoverse}
& \shortstack[c]{\textbf{V2X-}\\\textbf{Radar-I}}
& \shortstack[c]{\textbf{DAIR-}\\\textbf{V2X-I}}
& \textbf{CEAR}
& \textbf{JRDB}
& \shortstack[c]{\textbf{Fusion}\\\textbf{Portable}}
& \textbf{DROID}
& \shortstack[c]{\textbf{Humanoid}\\\textbf{Everyday}}
& \textbf{3EED} \\
\midrule
\multirow{2}{*}{VGGT}
& None & 23.094 & 15.127 & 20.034 & 16.584 & 22.986 & 23.293 & 3.216 & 3.466 & 18.317 & 0.224 & 0.119 & 6.783 \\
& \textbf{UniMetric} & \textbf{6.797} & \textbf{1.850} & \textbf{3.073} & \textbf{1.737} & \textbf{4.089} & \textbf{6.685} & \textbf{1.222} & \textbf{1.418} & \textbf{2.859} & \textbf{0.187} & \textbf{0.117} & \textbf{1.328} \\
\midrule
\multirow{2}{*}{VGGT-$\Omega$}
& None & 22.773 & 14.594 & 19.060 & 15.462 & 21.698 & 21.415 & 2.555 & 3.090 & 15.848 & 0.158 & 0.151 & 6.778 \\
& \textbf{UniMetric} & \textbf{1.906} & \textbf{0.726} & \textbf{0.630} & \textbf{1.215} & \textbf{1.244} & \textbf{2.033} & \textbf{1.037} & \textbf{0.772} & \textbf{0.653} & \textbf{0.118} & \textbf{0.135} & \textbf{1.053} \\
\midrule
\multirow{2}{*}{$\pi^3$}
& None & 1.668 & 1.710 & 0.848 & 1.596 & 3.454 & 5.761 & 0.893 & 0.952 & 0.971 & \textbf{0.090} & 0.259 & 2.116 \\
& \textbf{UniMetric} & \textbf{1.017} & \textbf{0.598} & \textbf{0.463} & \textbf{0.645} & \textbf{1.593} & \textbf{1.795} & \textbf{0.849} & \textbf{0.668} & \textbf{0.482} & 0.097 & \textbf{0.170} & \textbf{1.882} \\
\midrule
\multirow{2}{*}{DA-3}
& None & 22.813 & 13.748 & 18.846 & 15.017 & 20.615 & 21.256 & 2.794 & 3.157 & 16.496 & 0.155 & 0.126 & 6.619 \\
& \textbf{UniMetric} & \textbf{4.817} & \textbf{1.888} & \textbf{1.077} & \textbf{1.680} & \textbf{1.589} & \textbf{2.381} & \textbf{0.893} & \textbf{0.969} & \textbf{1.741} & \textbf{0.124} & \textbf{0.119} & \textbf{1.664} \\
\midrule
\multirow{2}{*}{Scal3R}
& None & 22.714 & 14.776 & 19.621 & 16.151 & 22.958 & 23.133 & 2.931 & 3.323 & 17.504 & 0.219 & 0.156 & 6.999 \\
& \textbf{UniMetric} & \textbf{2.129} & \textbf{1.320} & \textbf{1.453} & \textbf{2.123} & \textbf{6.196} & \textbf{7.336} & \textbf{1.398} & \textbf{1.599} & \textbf{3.861} & \textbf{0.126} & \textbf{0.150} & \textbf{2.387} \\
\midrule
\multirow{2}{*}{LingBot-Map}
& None & 21.881 & 14.021 & 18.956 & 13.616 & 22.141 & 22.274 & 2.541 & 2.981 & 15.754 & 0.219 & 0.156 & 6.273 \\
& \textbf{UniMetric} & \textbf{2.984} & \textbf{1.455} & \textbf{2.396} & \textbf{1.923} & \textbf{3.658} & \textbf{3.929} & \textbf{0.868} & \textbf{0.990} & \textbf{0.730} & \textbf{0.151} & \textbf{0.137} & \textbf{1.198} \\
\midrule
\multirow{2}{*}{4RC}
& None & 23.125 & 15.028 & 19.578 & 17.196 & 22.500 & 22.883 & 2.806 & 3.293 & 16.274 & 0.203 & 0.175 & 6.910 \\
& \textbf{UniMetric} & \textbf{3.739} & \textbf{1.277} & \textbf{1.892} & \textbf{1.945} & \textbf{2.583} & \textbf{2.256} & \textbf{1.084} & \textbf{0.964} & \textbf{0.848} & \textbf{0.134} & \textbf{0.147} & \textbf{1.788} \\
\midrule
\multirow{2}{*}{DVLT}
& None & 22.258 & 14.498 & 19.119 & 16.163 & 18.536 & 19.821 & 2.462 & 2.891 & 16.776 & 0.284 & 0.176 & 6.604 \\
& \textbf{UniMetric} & \textbf{5.068} & \textbf{1.292} & \textbf{0.736} & \textbf{1.094} & \textbf{1.929} & \textbf{4.076} & \textbf{0.586} & \textbf{1.173} & \textbf{1.524} & \textbf{0.110} & \textbf{0.129} & \textbf{1.576} \\
\midrule
\multirow{2}{*}{STream3R}
& None & 22.569 & 14.472 & 18.672 & 15.921 & 22.736 & 22.680 & 2.549 & 3.110 & 11.753 & 0.213 & 0.142 & 6.355 \\
& \textbf{UniMetric} & \textbf{6.695} & \textbf{1.408} & \textbf{2.450} & \textbf{1.176} & \textbf{4.428} & \textbf{4.063} & \textbf{0.883} & \textbf{0.924} & \textbf{6.362} & \textbf{0.103} & \textbf{0.126} & \textbf{1.843} \\
\bottomrule
\end{tabular*}
\label{tab:exps_3d_reconstruction}
\end{table}

\paragraph{Metric camera trajectories.}
\label{subsec:pose_results}
\Cref{tab:exps_ego_pose} evaluates whether a scale estimated from object
heights transfers to camera motion. No separate trajectory scale is fitted.
ATE Mean decreases in 81 of 90 model--subset pairs; for VGGT-$\Omega$
on FusionPortable, it falls from 39.863 to 4.238\,m under the same scale
that improves its depth and CD. Seven of the nine pairs without a decrease
occur on DROID, whose external camera is stationary and whose ATE values
are at the millimetre-to-centimetre scale. The other two are VGGT on CEAR
(8.296 to 8.838\,m) and Humanoid Everyday (0.150 to 0.158\,m), despite
improved depth on both subsets. These results delimit the role of a shared
scale: it calibrates translation magnitudes but cannot correct errors in
relative rotations or trajectory shape.

\begin{table}[!htbp]
\centering
\caption{\textbf{Ego pose estimation across autonomous-driving and embodied-AI benchmarks.}
We report ATE Mean (mean absolute trajectory error): the average camera-position
error in \chg{metres} after rigid alignment of each \chg{sequence} with scale fixed to 1;
lower is better. Bold marks the better result within each None/UniMetric pair;
ties at the displayed precision are not bolded. DA-3 denotes \chg{Depth Anything 3}.}

\unimetrictablefont
\setlength{\tabcolsep}{2pt}
\renewcommand{\arraystretch}{0.95}

\begin{tabular*}{\linewidth}{@{\extracolsep{\fill}}llcccccccccc@{}}
\toprule
\multirow{2}{*}{\textbf{Model}}
& \multirow{2}{*}{\shortstack{\textbf{Scale}\\\textbf{Recovery}}}
& \multicolumn{4}{c}{\textbf{Autonomous Driving}}
& \multicolumn{6}{c}{\textbf{Embodied AI}} \\
\cmidrule(lr){3-6}
\cmidrule(lr){7-12}
&
& \textbf{KITTI}
& \textbf{nuScenes}
& \textbf{Waymo}
& \textbf{Argoverse}
& \textbf{CEAR}
& \textbf{JRDB}
& \shortstack[c]{\textbf{Fusion}\\\textbf{Portable}}
& \textbf{DROID}
& \shortstack[c]{\textbf{Humanoid}\\\textbf{Everyday}}
& \textbf{3EED} \\
\midrule

\multirow{2}{*}{VGGT}
& None & 83.455 & 24.125 & 34.825 & 27.892 & \textbf{8.296} & 2.418 & 41.462 & 0.015 & \textbf{0.150} & 9.471 \\
& \textbf{UniMetric} & \textbf{36.194} & \textbf{5.146} & \textbf{12.331} & \textbf{5.840} & 8.838 & \textbf{2.397} & \textbf{21.360} & 0.015 & 0.158 & \textbf{5.666} \\
\midrule

\multirow{2}{*}{VGGT-$\Omega$}
& None & 83.272 & 23.792 & 34.364 & 27.901 & 7.374 & 2.442 & 39.863 & \textbf{0.007} & 0.136 & 9.485 \\
& \textbf{UniMetric} & \textbf{15.511} & \textbf{2.476} & \textbf{5.090} & \textbf{3.906} & \textbf{2.715} & \textbf{0.921} & \textbf{4.238} & 0.009 & \textbf{0.101} & \textbf{4.237} \\
\midrule

\multirow{2}{*}{$\pi^3$}
& None & 15.176 & 6.738 & 5.780 & 7.781 & 3.054 & 0.964 & 10.318 & 0.018 & 0.346 & 5.090 \\
& \textbf{UniMetric} & \textbf{10.838} & \textbf{1.397} & \textbf{3.019} & \textbf{2.415} & \textbf{2.381} & \textbf{0.840} & \textbf{4.323} & \textbf{0.014} & \textbf{0.244} & \textbf{4.012} \\
\midrule

\multirow{2}{*}{DA-3}
& None & 83.171 & 23.253 & 34.059 & 26.853 & 7.272 & 2.359 & 39.946 & 0.020 & 0.176 & 9.413 \\
& \textbf{UniMetric} & \textbf{33.504} & \textbf{7.274} & \textbf{5.574} & \textbf{10.790} & \textbf{2.990} & \textbf{0.953} & \textbf{12.833} & 0.020 & \textbf{0.164} & \textbf{5.288} \\
\midrule

\multirow{2}{*}{Scal3R}
& None & 82.702 & 23.896 & 34.544 & 27.506 & 7.517 & 2.402 & 40.236 & \textbf{0.015} & 0.133 & 9.381 \\
& \textbf{UniMetric} & \textbf{19.057} & \textbf{5.041} & \textbf{9.668} & \textbf{7.109} & \textbf{4.185} & \textbf{1.099} & \textbf{12.909} & 0.021 & \textbf{0.114} & \textbf{6.380} \\
\midrule

\multirow{2}{*}{LingBot-Map}
& None & 81.219 & 23.804 & 34.446 & 27.607 & 7.119 & 2.352 & 40.385 & \textbf{0.003} & 0.113 & 9.422 \\
& \textbf{UniMetric} & \textbf{17.922} & \textbf{5.141} & \textbf{13.188} & \textbf{8.033} & \textbf{2.866} & \textbf{0.985} & \textbf{11.181} & 0.004 & \textbf{0.076} & \textbf{4.755} \\
\midrule

\multirow{2}{*}{4RC}
& None & 83.342 & 23.807 & 34.414 & 30.199 & 7.149 & 2.440 & 38.982 & \textbf{0.009} & 0.160 & 9.589 \\
& \textbf{UniMetric} & \textbf{27.065} & \textbf{7.001} & \textbf{7.609} & \textbf{13.159} & \textbf{2.770} & \textbf{0.686} & \textbf{6.670} & 0.016 & \textbf{0.099} & \textbf{6.026} \\
\midrule

\multirow{2}{*}{DVLT}
& None & 82.929 & 23.633 & 34.189 & 29.918 & 7.109 & 2.296 & 40.506 & 0.040 & 0.256 & 9.310 \\
& \textbf{UniMetric} & \textbf{38.951} & \textbf{5.192} & \textbf{6.462} & \textbf{6.142} & \textbf{3.951} & \textbf{1.012} & \textbf{10.880} & \textbf{0.023} & \textbf{0.169} & \textbf{6.010} \\
\midrule

\multirow{2}{*}{STream3R}
& None & 83.091 & 23.561 & 33.937 & 29.642 & 7.913 & 2.297 & 37.322 & \textbf{0.010} & 0.125 & 9.371 \\
& \textbf{UniMetric} & \textbf{53.184} & \textbf{5.915} & \textbf{12.517} & \textbf{6.609} & \textbf{6.569} & \textbf{1.170} & \textbf{31.107} & 0.016 & \textbf{0.118} & \textbf{6.065} \\

\bottomrule
\end{tabular*}
\label{tab:exps_ego_pose}
\end{table}

Across these comparisons, metric accuracy improves at the image, point-cloud,
and trajectory levels using one scale per sequence. The absolute errors
remain dependent on scene depth, camera travel, and the original geometry;
the per-subset paired comparisons therefore complement the suite means.
Additional depth and corresponding-point error metrics are reported in
\Cref{sec:supplementary_material}.

\subsection{Ablation Studies}
\label{subsec:ablations}

\rev{The preceding results establish the effect of shared scale correction.
We next compare sources of metric scale and examine how UniMetric combines
its semantic priors, using VGGT-$\Omega$ on nuScenes and FusionPortable.}

\begin{table}[!htbp]
\centering
\caption{\chg{\textbf{Scale calibration of VGGT-$\Omega$ predictions.} All rows
rescale the same \chg{GFM} outputs by one factor per sequence, estimated
from metric depth (\Cref{app:metric_depth_calibration}) or by UniMetric.
Oracle (gray) fits the factor to ground truth separately for each metric.
\textbf{Bold} and \underline{underline} mark the best and second-best
non-oracle results.}}
\label{tab:ablation}
\unimetrictablefont
\setlength{\tabcolsep}{2pt}
\renewcommand{\arraystretch}{0.7}
\begin{tabular*}{\linewidth}{@{\extracolsep{\fill}}lcccccccc@{}}
\toprule
\multirow{2}{*}{\textbf{Scale calibration}}
& \multicolumn{4}{c}{\textbf{nuScenes}} & \multicolumn{4}{c}{\textbf{FusionPortable}} \\
\cmidrule(lr){2-5}\cmidrule(lr){6-9}
& Abs Rel$\downarrow$ & $\delta_1\uparrow$ & CD$\downarrow$ & ATE$\downarrow$
& Abs Rel$\downarrow$ & $\delta_1\uparrow$ & CD$\downarrow$ & ATE$\downarrow$ \\
\midrule
None & 0.979 & 0.0 & 14.594 & 23.792 & 0.885 & 0.9 & 15.848 & 39.863 \\
Depth Pro & \underline{0.212} & 55.6 & 1.107 & \underline{5.134} & 0.543 & 22.6 & 9.378 & 66.352 \\
UniDepth v2 & 0.278 & 53.1 & 1.638 & 13.515 & 0.485 & 14.8 & 1.450 & 23.485 \\
Depth Anything V2 & 0.347 & 29.0 & 1.705 & 11.423 & \underline{0.396} & 47.0 & \underline{0.702} & \underline{10.855} \\
ZoeDepth-NK & 0.218 & \underline{65.9} & \underline{1.044} & 6.965 & 0.432 & \underline{49.4} & 3.940 & 56.668 \\
UniMetric (Ours) & \textbf{0.152} & \textbf{82.0} & \textbf{0.726} & \textbf{2.476} & \textbf{0.331} & \textbf{52.2} & \textbf{0.653} & \textbf{4.238} \\
\midrule
Oracle & \textcolor{gray}{0.087} & \textcolor{gray}{92.9} & \textcolor{gray}{0.546} & \textcolor{gray}{1.407} & \textcolor{gray}{0.219} & \textcolor{gray}{80.1} & \textcolor{gray}{0.384} & \textcolor{gray}{2.972} \\
\bottomrule
\end{tabular*}
\end{table}

\begin{table}[!htbp]
\centering
\caption{\textbf{Design ablations of UniMetric with VGGT-$\Omega$.}
The variants replace scale intervals with their midpoints (Point priors),
assign equal weights to valid observations (Uniform weights), or retain
only the lowest-weight observation (Single anchor).
\chg{\textbf{Bold} and \underline{underline} mark the best and second-best
results.}}
\label{tab:unimetric_design_ablations}
\unimetrictablefont
\setlength{\tabcolsep}{4pt}
\renewcommand{\arraystretch}{0.7}
\begin{tabular*}{\linewidth}{@{\extracolsep{\fill}}l*{8}{c}@{}}
\toprule
\multirow{2}{*}{\textbf{Variant}}
& \multicolumn{4}{c}{\textbf{nuScenes}}
& \multicolumn{4}{c}{\textbf{FusionPortable}} \\
\cmidrule(lr){2-5}\cmidrule(lr){6-9}
& Abs Rel$\downarrow$ & $\delta_1\uparrow$ & CD$\downarrow$ & ATE$\downarrow$
& Abs Rel$\downarrow$ & $\delta_1\uparrow$ & CD$\downarrow$ & ATE$\downarrow$ \\
\midrule
Point priors
& \underline{0.172} & \underline{79.1} & \underline{0.989} & \underline{5.231}
& \underline{0.452} & \underline{47.4} & \underline{1.806} & \underline{19.430} \\
Uniform weights
& 0.180 & 67.5 & 1.235 & 5.902
& 0.469 & 46.9 & 1.828 & 22.636 \\
Single anchor
& 0.334 & 45.4 & 1.983 & 21.643
& 0.683 & 18.0 & 6.086 & 48.649 \\
\midrule
\textbf{UniMetric (full)}
& \textbf{0.152} & \textbf{82.0} & \textbf{0.726} & \textbf{2.476}
& \textbf{0.331} & \textbf{52.2} & \textbf{0.653} & \textbf{4.238} \\
\bottomrule
\end{tabular*}
\end{table}

\paragraph{Comparison with metric-depth methods.}
\rev{\chg{We compare UniMetric with calibrating the same VGGT-$\Omega$
reconstruction using four pretrained metric depth estimators: Depth
Pro~\citep{bochkovskii2025depthpro}, UniDepth~v2\chg{~\citep{piccinelli2025unidepthv2}},
Depth Anything~V2~\citep{yang2024depthanythingv2}, and
ZoeDepth~\citep{bhat2023zoedepth}. Each baseline estimates one scale per
sequence from its metric depth maps, matching the granularity of UniMetric
(\Cref{app:metric_depth_calibration}).} UniMetric achieves the best
non-oracle results across all four metrics on both nuScenes and
FusionPortable \chg{(\Cref{tab:ablation})}. Oracle fits a ground-truth scale
separately for each metric, indicating further room for improvement
through global rescaling. This comparison keeps the GFM reconstruction fixed,
so the differences measure how effectively each source of metric information
calibrates that reconstruction.}

\paragraph{Ablation of key designs.}
\rev{The full method achieves the highest $\delta_1$ accuracy on both
nuScenes (\textbf{82.0\%}) and FusionPortable (\textbf{52.2\%})
(\Cref{tab:unimetric_design_ablations}). Replacing intervals with their
midpoints lowers accuracy to 79.1\% and 47.4\%; assigning uniform weights
lowers it to 67.5\% and 46.9\%. The full method also has lower Abs Rel,
CD, and ATE Mean than either variant. Together, these comparisons support
preserving the range of plausible sizes and accounting for observation
reliability when estimating a shared scale. Retaining only the lowest-weight
observation yields $\delta_1$ scores of 45.4\% and 18.0\%, illustrating
the sensitivity to a poor individual reference and the benefit of combining
the available observations.}

\subsection{Qualitative Results}
\label{subsec:qualitative_results}

\begin{figure}[t]
\centering
\setlength{\abovecaptionskip}{1pt}
\includegraphics[width=\linewidth]{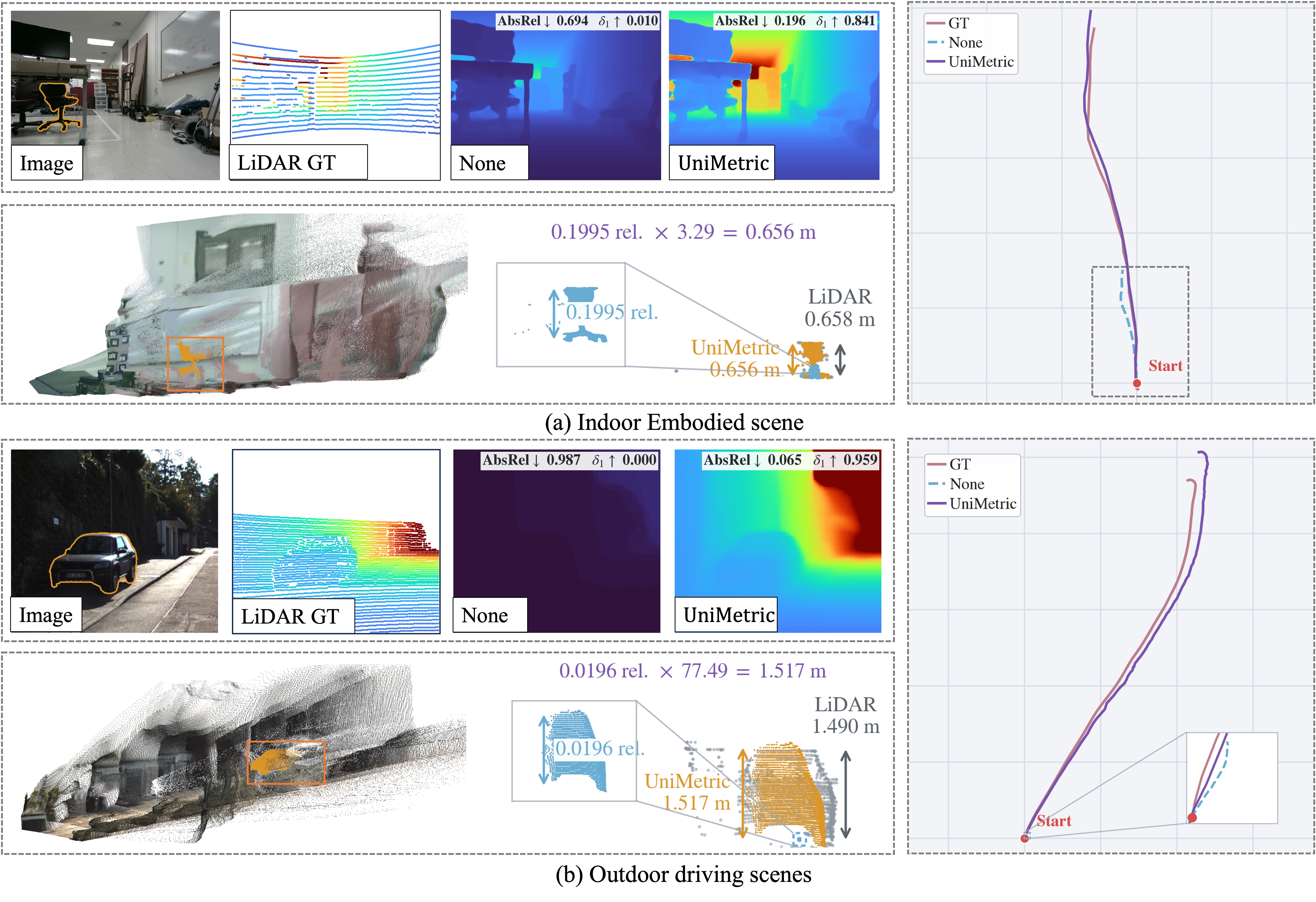}
\caption{\textbf{Qualitative scale recovery with VGGT-$\Omega$ on CEAR
(top) and KITTI (bottom).} Depth maps and camera trajectories compare
predictions before (None) and after (UniMetric) scale correction.
Point-cloud views show the highlighted regions and their measured heights
alongside LiDAR references. These measurements cover visible regions
rather than complete objects. GT denotes ground truth; rel. denotes the
original reconstruction unit.}
\label{fig:qualitative_main}
\end{figure}

\rev{\chg{We examine how the quantitative improvements appear in embodied and
autonomous-driving scenes. For VGGT-$\Omega$, UniMetric recovers
scale factors of 3.29 and 77.49 in the CEAR and KITTI examples,
respectively (\Cref{fig:qualitative_main}).
Both bring the measured spans of the highlighted regions to within 3\,cm
of their LiDAR references:
the chair measures 0.656\,m against 0.658\,m, and the car 1.517\,m against
1.490\,m. On the displayed frames, Abs Rel falls from 0.694 to 0.196 on CEAR
and from 0.987 to 0.065 on KITTI, and $\delta_1$ rises from 1.0\% to 84.1\%
and from 0.0\% to 95.9\%. The uncorrected trajectories are compressed to a
small fraction of the travelled distance; the same factors bring their
extent closer to that of the reference paths, although local deviations
remain. The examples connect scale recovery to visible object dimensions,
dense depth, and camera motion within the same reconstruction.}}

\begin{figure}[!tp]
\centering
\includegraphics[width=\linewidth,height=0.86\textheight,keepaspectratio]{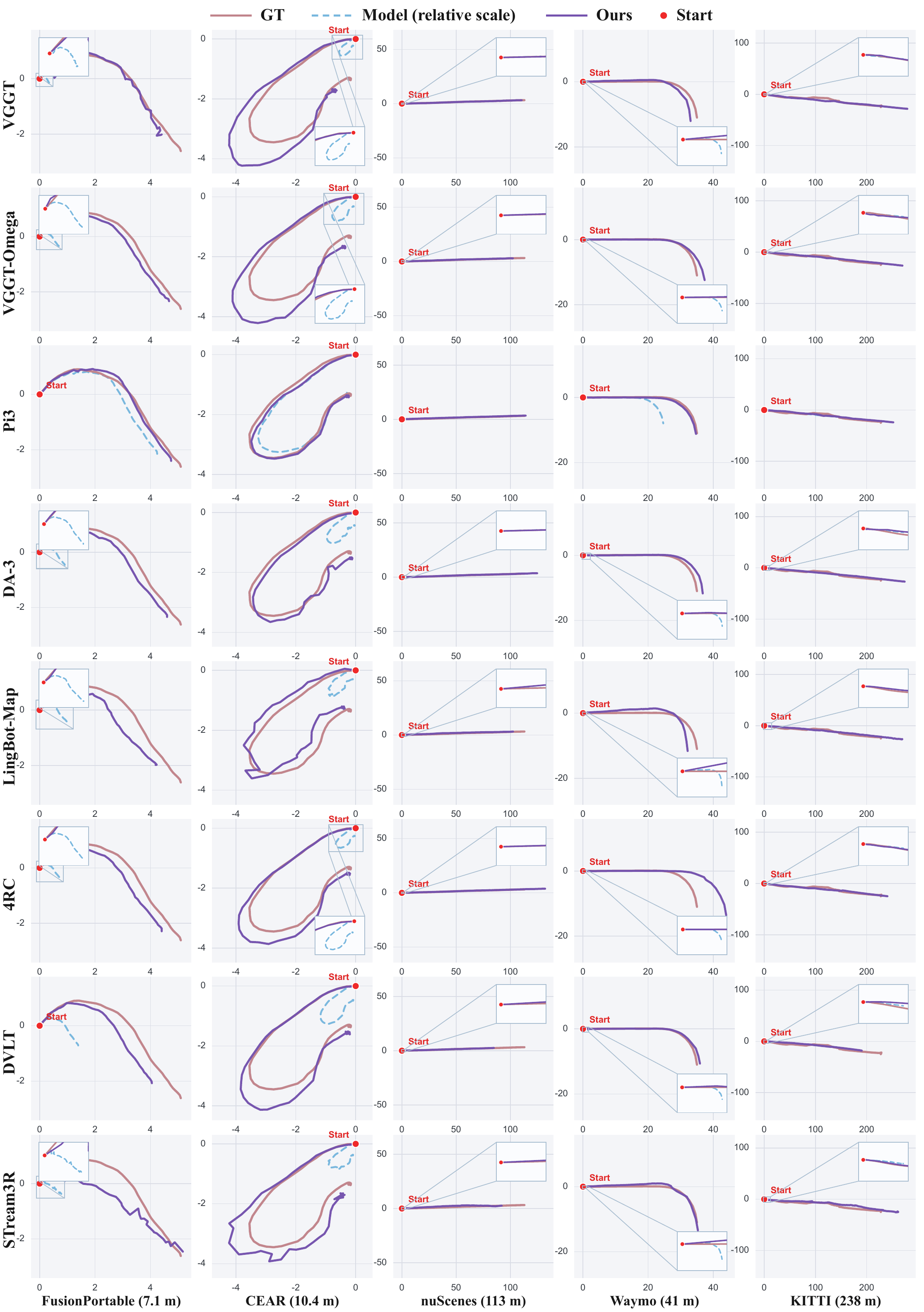}
\caption{\textbf{Camera trajectory scale recovery across eight GFMs.}
Rows show models; columns show selected scenes from FusionPortable, CEAR,
nuScenes, Waymo, and KITTI. Trajectories are projected onto
the $Z$--$X$ plane (horizontal--vertical). GT and UniMetric (Ours) are in
metres; Model retains its original relative scale. Scene labels give the
GT path length. Red dots mark the common starting point; insets magnify
the original predictions. GT denotes ground truth.}
\label{fig:camera_pose_eight_models}
\end{figure}

\paragraph{Camera trajectories across models.}
\Cref{fig:camera_pose_eight_models} compares eight GFMs on selected scenes
from FusionPortable, CEAR, nuScenes, Waymo, and KITTI. The examples show
how scale correction changes trajectory extent across embodied and driving
scenes, while preserving the path shape predicted by each model. On the
displayed FusionPortable and Waymo sequences, $\pi^3$, VGGT-$\Omega$, and
4RC recover extents closer to the references; residual deviations remain,
particularly for 4RC on Waymo. For display, predictions are rotated
by a global rigid fit and translated to share the reference starting point,
without fitting scale. These examples illustrate the trajectory effects
measured by the per-subset ATE results in \Cref{tab:exps_ego_pose}.
Complete depth visualizations on all twelve subsets are provided in
\Cref{sec:supp_qualitative}.

\FloatBarrier
\section{Conclusion}
\label{sec:conclusion}
\rev{We presented UniMetric, a training-free framework for recovering metric
scale in visual geometry foundation models. Its central insight is that
uncertain size priors of familiar objects can jointly constrain a shared
scale for scene geometry and camera motion. Theoretical analysis identifies
conditions under which aggregation reduces estimation error, and controlled
experiments with real VLM intervals demonstrate gains as more observations
are combined. MetricBench tests the resulting method across nine models
and twelve sensor-referenced subsets, spanning near-field manipulation,
mobile robotics, and driving scenes. The same sequence-level correction
improves depth, reconstruction, and trajectory accuracy at the suite level,
with per-subset comparisons showing both its broad benefit and cases where
rescaling is insufficient. \chg{Semantic size knowledge is thus a practical
source of metric information, which could
enrich large-scale visual data with metric supervision for embodied
foundation-model pretraining.}}

\rev{\textbf{Limitations.} UniMetric depends on the quality of semantic priors
and reconstructed object measurements, including the availability of anchors
with measurable heights. A shared scale corrects global scale discrepancies
but cannot resolve local geometric distortions or trajectory drift.}

\bibliography{egbib}
\bibliographystyle{plainnat}

\clearpage
\appendix
\raggedbottom
\setcounter{table}{0}
\setcounter{figure}{0}
\setcounter{equation}{0}
\renewcommand{\thetable}{S\arabic{table}}
\renewcommand{\thefigure}{S\arabic{figure}}
\renewcommand{\theequation}{S\arabic{equation}}
\renewcommand{\theHtable}{supp.\arabic{table}}
\renewcommand{\theHfigure}{supp.\arabic{figure}}
\renewcommand{\theHequation}{supp.\arabic{equation}}
\crefalias{section}{appendix}
\crefalias{subsection}{appendix}
\crefalias{subsubsection}{appendix}
\section{Proofs}
\label{app:proofs}

\subsection{Mean-Square Consistency}
\label{app:mean_square_consistency}

\noindent\textbf{Proof of Proposition 1.}
Zero means and pairwise uncorrelatedness imply
$\mathbb{E}[\epsilon_i\epsilon_j]=0$ for $i\ne j$ and
$\mathbb{E}[\epsilon_i^2]=\operatorname{Var}(\epsilon_i)$.
Expanding the squared estimation error and taking expectations gives
\begin{equation}
\begin{aligned}
\mathbb{E}\!\left[(\bar{s}_N-s_\star)^2\right]
&=\frac{1}{N^2}\mathbb{E}\!\left[
\left(\sum_{i=1}^{N}\epsilon_i\right)^2\right]
=\frac{1}{N^2}\left(
\sum_{i=1}^{N}\mathbb{E}[\epsilon_i^2]
+2\sum_{i<j}\mathbb{E}[\epsilon_i\epsilon_j]
\right)\\
&=\frac{1}{N^2}\sum_{i=1}^{N}\operatorname{Var}(\epsilon_i)
\leq\frac{N\sigma^2}{N^2}
=\frac{\sigma^2}{N}.
\end{aligned}
\label{eq:joint_scale_mse_proof}
\end{equation}
which is \Cref{eq:joint_scale_mse}. As $\sigma^2$ is independent of $N$, the
bound tends to zero, proving mean-square convergence.\nobreak\hfill$\square$

\subsection{Properties of the Interval Objective}
\label{app:interval_objective}

\noindent\textbf{Proposition 2 (Structure of the interval objective).}
\emph{Let $\mathcal{A}\neq\emptyset$, with positive weights and finite
intervals $S_{t,i}\subset(0,\infty)$. Then $J$ is convex and piecewise
quadratic, and its minimizers lie in the convex hull of the intervals. If
the intervals have a nonempty common intersection, the set of minimizers is
exactly $\bigcap_{(t,i)\in\mathcal{A}}S_{t,i}$; otherwise, the minimizer is
unique.}

\noindent\textbf{Proof.}
Write $S_{t,i}=[a_{t,i},b_{t,i}]$ with $0<a_{t,i}\le b_{t,i}$ and let
$\phi_{t,i}(s)=w_{t,i}\,\mathrm{dist}(s,S_{t,i})^{2}$ with $w_{t,i}>0$. Then
$\phi_{t,i}(s)=w_{t,i}(a_{t,i}-s)^{2}$ for $s<a_{t,i}$, $\phi_{t,i}(s)=0$ for
$s\in S_{t,i}$, and $\phi_{t,i}(s)=w_{t,i}(s-b_{t,i})^{2}$ for $s>b_{t,i}$;
each $\phi_{t,i}$ is therefore continuously differentiable, convex, and
piecewise quadratic, and so is $J=\sum_{t,i}\phi_{t,i}$.

\emph{Location of the minimizers.} Let $a_{\min}=\min_{t,i}a_{t,i}$ and
$b_{\max}=\max_{t,i}b_{t,i}$. For $s<a_{\min}$ every term is strictly
decreasing in $s$, and for $s>b_{\max}$ every term is non-decreasing and at
least one is strictly increasing; hence every minimizer of $J$ on
$(0,\infty)$ lies in $[a_{\min},b_{\max}]$, and a minimizer exists because $J$
is continuous on this compact interval.

\emph{Non-empty intersection.} Let
$\mathcal{S}_{\cap}=\bigcap_{t,i}S_{t,i}\neq\emptyset$. For
$s\in\mathcal{S}_{\cap}$ every term vanishes, so $J(s)=0$. For
$s\notin\mathcal{S}_{\cap}$ at least one interval excludes $s$, and the
corresponding term is strictly positive, so $J(s)>0$. Since $J\ge0$, the set
of minimizers is exactly $\mathcal{S}_{\cap}$.

\emph{Empty intersection.} If $\mathcal{S}_{\cap}=\emptyset$, every $s>0$ lies
outside at least one interval $S_{t,i}$; on a neighbourhood of such an $s$ the
term $\phi_{t,i}$ coincides with a quadratic of positive leading coefficient
and is strictly convex, so $J$ is strictly convex on a neighbourhood of every
point. A convex function that is strictly convex on a neighbourhood of every
point is strictly convex on its domain: if $J$ were affine on some segment
$[x,y]$, it could not be strictly convex around the midpoint of that segment.
A strictly convex function has at most one minimizer, and one exists by the
previous paragraph. \hfill$\square$

\noindent\textbf{Remark (point priors as a special case).}
\rev{In the special case that every interval collapses to a point and the
weights are equal, $S_{t,i}=\{y_{t,i}\}$ and $w_{t,i}\equiv1$,
\Cref{eq:objective} reduces to
\begin{equation}
J(s)=\sum_{t,i}\big(s-y_{t,i}\big)^{2},
\qquad
\hat{s}=\bar{s}_N,
\label{eq:degenerate}
\end{equation}
i.e., the averaged estimator analysed in
\Cref{subsec:semantic_scale_principle}. \chg{The interval objective therefore
generalizes the mean to uncertain priors, and Proposition~1 covers this
special case.}}

\section{Empirical Validation Protocol}
\label{app:empirical_validation}

We specify the height-prior acquisition and scale-estimation protocol for
\Cref{fig:empirical_validation}. \rev{The study is run on the MetricBench
subsets of nuScenes, JRDB, 3EED, and V2X-Radar-I in their entirety, so the
sequences, frames, and annotations are exactly those of
\Cref{tab:bench_detail}: 116, 27, 20, and 30 sequences, 193 in total, with
13{,}292 frames.}

\paragraph{VLM configuration and queries.}
We use \chg{Seed1.8~\citep{seed2026seed18}} (\texttt{seed-1.8-max}) with temperature $0.1$ and a maximum output
length of $512$ tokens. Each query contains a full RGB image with one
annotated target marked by a thick red rectangle and the label
\texttt{TARGET}, together with its semantic category. The surrounding
scene remains visible. The VLM receives neither the annotated physical
height nor the injected scale factor.
The scale references are physical height intervals, represented by
\texttt{height\_interval\_m}, with \texttt{confidence} used for weighting.
\Cref{fig:empirical_vlm_prompt} gives a height-only adaptation of the
query template, using the category and visible subtype to estimate the
target's bottom-to-top physical extent in metres. \chg{The reported results
use the height outputs of the original queries from which this template is
adapted.}

\begin{figure}[tbp]
\centering
\begin{unimetricpromptfigure}
\begin{promptsection}{Task}
The image contains exactly one evaluation target highlighted by a thick
red rectangle and the label TARGET. The provided semantic class is
``\texttt{\{semantic\_class\}}''. Estimate a plausible real-world physical height
interval for this specific target, using semantic category and visible
subtype cues. The rectangle is only for localization.

Height means the bottom-to-top physical extent in meters.

Do not output a scale factor. Do not infer height from rectangle
pixel size or camera geometry. Return one JSON object only, with this
exact schema:
\end{promptsection}
\begin{unimetricpromptoutput}
\begin{lstlisting}[style=promptjson]
{
  "object_type": "{semantic_class}",
  "height_interval_m": [minimum, maximum],
  "confidence": 0.0,
  "rationale": "brief semantic reason"
}
\end{lstlisting}
\end{unimetricpromptoutput}
Both height bounds must be finite positive numbers in meters, with
minimum not greater than maximum. Use a conservative but informative
height interval.
\end{unimetricpromptfigure}
\caption{Height-only user-prompt template, adapted for height-prior acquisition.
\texttt{\{semantic\_class\}} is replaced by the annotated category.
The output specifies a height interval in metres and its confidence.}
\label{fig:empirical_vlm_prompt}
\end{figure}

For example, \chg{the response for a \texttt{Car} in 3EED gives}
\texttt{height\_interval\_m: [1.5, 1.7]} and
\texttt{confidence: 0.8}; its rationale identifies a compact crossover
SUV. We parse the height bounds in metres and retain the model-reported
confidence as an observation weight.

\paragraph{Data scale.}
\rev{Candidate availability in \Cref{fig:empirical_validation}(a) and the
recovery curves in panels~(b,c) are computed on all $193$ sequences.}

\paragraph{Height references and scale constraints.}
Each observation corresponds to one annotated object in one image; the same
object may provide observations in multiple frames, and, as in UniMetric,
repeated sightings are treated as separate observations.
\Cref{fig:empirical_validation}(a) summarizes the candidate observation
counts. For observation $i$, let $H_i^{\mathrm{GT}}$ be the annotated object
height and construct the relative height $h_i=qH_i^{\mathrm{GT}}$ as in the
main text, with $q=0.3$ and true recovery factor $s_\star=1/q$. The height
bounds $H_i^-$ and $H_i^+$ are the raw interval returned by the
vision--language model, used without any calibration against annotated
heights, and the scale constraint is
\begin{equation}
S_i=\left[\frac{H_i^-}{h_i},\frac{H_i^+}{h_i}\right],
\label{eq:empirical_scale_interval}
\end{equation}
which is \Cref{eq:interval} with $h_i$ taken from ground truth instead of
from a predicted pointmap. Different objects thereby constrain the same
recovery factor rather than requiring a common physical height.

\paragraph{Joint estimation and error aggregation.}
\rev{The estimator is the Global Scale Recovery stage of UniMetric applied
unchanged. Each observation receives the weight $w_i=q_i\,\alpha_{c_i}$ of
\Cref{eq:weight}, where $q_i$ is the confidence returned with the interval
and $\alpha_c$ is the category reliability of \Cref{app:impl}; annotated
categories without supplied reliability metadata receive $\alpha_c=1$. The shared scale
$\hat s_K$ is the minimizer of the interval objective \Cref{eq:objective}
over the $K$ observations, computed by the golden-section search of
\Cref{app:impl}; when the intervals share a common intersection, the search
returns a point of that intersection (\Cref{app:interval_objective}).}

Let $K$ denote the number of frame--object observations used for fusion.
\rev{Recovery is evaluated on all $193$ sequences, the same set at every
$K$; for a sequence with $n_c<K$ candidates the observation set at $K$ is
its full set of $n_c$ observations.} We use $200$ sampling repetitions with
random seed $42$; \rev{each repetition draws a uniformly random ordering of
the sequence's observations}, and within a repetition the first $K$ entries
of that ordering form the observation set, so increasing $K$ retains earlier
observations and adds new ones. \chg{All budgets share the same VLM
predictions.}
The dimensionless absolute relative scale error for one sampling trial is
\begin{equation}
e_K=\left|\frac{\hat s_K}{s_\star}-1\right|
=\left|q\hat s_K-1\right|.
\label{eq:empirical_relative_scale_error}
\end{equation}
We average $e_K$ over sampling repetitions within each sequence, then
equally across sequences in each dataset. Multiplying by $100$ yields
the mean absolute relative scale error (MARE), reported as a percentage in
\Cref{fig:empirical_validation}(b).

\paragraph{Accuracy of joint scale estimation.}
\Cref{tab:empirical_joint_scale_results} reports MARE at
$K\in\{1,10,20,50\}$ in the main text. The $95\%$ confidence intervals at $K=50$ are
computed using $2{,}000$ bootstrap resamples of sequences, after
averaging the sampling repetitions within each sequence.

\paragraph{Scale-induced distance error.}
For a true distance $d$ in metres, \Cref{fig:empirical_validation}(c)
converts the relative scale error at $K=20$ into mean scale-induced
distance error:
\begin{equation}
E(d)=d\,\frac{\operatorname{MARE}(20)}{100},
\qquad E_0(d)=|q-1|d=0.7d.
\label{eq:empirical_distance_error}
\end{equation}
Here $E(d)$ is the corrected error (coloured curves), and $E_0(d)$ is the
uncorrected reference induced by the prescribed scaling (grey dashed
line); both are measured in metres. For example, a $2\%$ scale error
corresponds to a $1\,\mathrm{m}$ distance error at $50\,\mathrm{m}$.
These curves are an analytical conversion of the scale statistics in
panel~(b).

\section{Implementation Details}
\label{app:impl}

This section supplements the main method with the design of semantic-prior prompts and the settings for semantic inference, segmentation, and scale estimation. \Cref{fig:object_anchor_prompt,fig:support_surface_prompt} give a generalized formulation of the semantic queries; dataset-specific prior sources used for the reported evaluation are described in \Cref{app:bench_datasets}.

\subsection{Prompt Design and Templates}
\label{app:prompt_templates}

\paragraph{Object selection and size priors.}
The object-anchor template combines semantic recognition with explicit requirements on the visibility of the measured extent. A single task parameter, \texttt{\{ANCHOR\_GROUP\}}, distinguishes movable objects from static scene elements. Both groups follow the same instructions for selecting recognizable instances, providing physical-height intervals, and reporting their image locations. Movable objects refer to semantic categories, irrespective of whether an instance is moving in the observed frame.

The image provides evidence about category, recognizable subtype, pose, and visibility. Physical-height intervals are obtained from semantic knowledge about the identified object. This distinction allows appearance to inform the choice of a relevant prior without treating apparent image size as a metric measurement. The template requests a plausible range that reflects physical variation and instructs the VLM to omit objects for which such a range cannot be supported.

The geometric quantity of interest is the complete extent along the vertical direction. The template therefore emphasizes identifiable upper and lower boundaries and a pose compatible with the proposed height prior. Lateral truncation may be acceptable when the vertical extent remains clear; missing height boundaries or an unsuitable pose make an object unreliable as a height anchor. \Cref{fig:object_anchor_prompt} collects these requirements in a general template.

Replace \texttt{\{ANCHOR\_GROUP\}} with either ``movable objects'' or ``static scene elements.'' The template specifies a common selection criterion and response format without prescribing a dataset-specific category inventory or numerical height intervals.

\begin{figure}[tbp]
\begin{unimetricpromptbox}{General object-anchor prompt}
\begin{promptsection}{Task}
Identify familiar objects in the input RGB image that can provide reliable physical-height priors for metric-scale recovery. Requested anchor group: \texttt{\{ANCHOR\_GROUP\}}. For movable objects, actual motion is not required.
\end{promptsection}

\begin{promptsection}{Object selection}
\begin{promptitems}
\item Select individual, recognizable instances whose semantic category supports a meaningful physical-height range.
\item Use visible semantic attributes to identify a more specific subtype only when the image supports that distinction. Do not invent an exact product identity or unobservable object properties.
\item Prefer clear instances with identifiable upper and lower boundaries. Both boundaries must refer to the same object.
\item The requested dimension is the complete vertical extent of the object in its observed pose, not an arbitrary visible segment or its longest side.
\item Exclude instances whose pose or deformation makes the category-based height prior inappropriate. For people, use upright standing instances with reliable full-body height cues.
\end{promptitems}
\end{promptsection}

\begin{promptsection}{Visibility}
\begin{promptitems}
\item Distinguish image-boundary truncation from occlusion by another object.
\item Lateral truncation is acceptable only if category recognition and the complete top-to-bottom extent remain reliable.
\item Reject objects with missing upper or lower boundaries, height-direction occlusion, severe blur, ambiguous instance boundaries, or insufficient visible detail.
\item Do not treat reflections, shadows, or groups of objects as individual physical anchors.
\end{promptitems}
\end{promptsection}

\begin{promptsection}{Physical-height reasoning}
\begin{promptitems}
\item Provide a plausible lower and upper height bound in metres using semantic category knowledge and common knowledge of physical objects.
\item Account for plausible variation within the recognized category or subtype. Use a wider interval when the semantic evidence is less specific.
\item Do not derive metric height from pixel size, perspective, apparent distance, assumed camera height, or reconstructed depth.
\item Do not substitute a prescribed category value for semantic reasoning. If no defensible height interval is available, omit the object.
\item Require \(0 < \text{lower bound} \leq \text{upper bound}\). Do not imply exact measurement.
\end{promptitems}
\end{promptsection}

\begin{promptsection}{Localization and response}
\begin{promptitems}
\item Draw a tight bounding box around the visible body of the selected instance. Exclude neighboring objects and unnecessary background.
\item Use integer coordinates \texttt{[x1, y1, x2, y2]} normalized to [0, 1000], with (0, 0) at the top-left and (1000, 1000) at the bottom-right.
\item Return one JSON object per selected instance with these fields:
\end{promptitems}
\end{promptsection}
\begin{unimetricpromptoutput}
\promptfield{object\_class}{a concise lowercase category name;}
\promptfield{object\_bbox\_2d}{the normalized bounding box;}
\promptfield{height\_interval}{[lower bound, upper bound] in metres;}
\promptfield{confidence}{a value in [0, 1] indicating confidence in anchor suitability;}
\promptfield{reason}{a brief semantic justification for the height interval;}
\promptfield{is\_occluded}{whether another physical object blocks part of the instance;}
\promptfield{is\_truncated}{whether the instance is cut by the image boundary.}
\end{unimetricpromptoutput}
\begin{promptitems}
\item Return a JSON array only. If no suitable anchor is visible, return \texttt{[]}.
\end{promptitems}
\end{unimetricpromptbox}
\caption{General object-anchor prompt template. The task parameter
\texttt{\{ANCHOR\_GROUP\}} selects movable objects or static scene
elements; the template fixes the selection criteria, the physical-height
interval, and the response format without a dataset-specific category
inventory.}
\label{fig:object_anchor_prompt}
\end{figure}

\paragraph{Support-surface selection.}
The support-surface template identifies a region whose reconstructed geometry can provide a vertical reference. Suitable regions may occur on roads, floors, ground, or tabletops, depending on the scene. Their suitability depends on visible surface structure and consistency with the surrounding objects, rather than membership in a fixed category. The prompt favors a clear, approximately horizontal planar region and excludes surfaces whose inclination, curvature, or occlusion makes that reference ambiguous. The VLM supplies a semantic label and image region; the surface normal is subsequently estimated from the reconstructed geometry.

\paragraph{Common response and scene context.}
The templates use consistent output conventions across scene types: bounding boxes localize the semantic evidence, object intervals express plausible physical heights in metres, and confidence and visibility attributes support subsequent selection. Optional scene descriptions can help disambiguate object categories and support surfaces without prescribing their metric dimensions. Additional category reliability or priority metadata, when used for weighting, is specified separately from the common prompt and its physical-height estimates.

The same instructions accommodate outdoor, indoor, and manipulation scenes. Surface categories are illustrative rather than an exhaustive list, and no image location is assumed to contain a valid support surface.

\begin{figure}[tbp]
\begin{unimetricpromptbox}{General support-surface prompt}
\begin{promptsection}{Task}
Identify the single most reliable visible support-surface region in the input RGB image for establishing a vertical reference in the scene.
\end{promptsection}

\begin{promptsection}{Surface selection}
\begin{promptitems}
\item Select a recognizable surface that supports objects or forms the local ground reference. Depending on the scene, candidates may include a road, floor, ground surface, or tabletop.
\item Require a clear, approximately planar and approximately horizontal region that is consistent with the surrounding scene.
\item Prefer a sufficiently visible, unobstructed surface patch over a larger region containing mixed surfaces or foreground objects.
\item Do not select a surface solely because of its category or position in the image. Do not assume that the lower image region is always valid ground.
\item Reject walls, facades, ceilings, steep ramps, curved surfaces, and regions that combine different surface orientations or height levels.
\item If a reliable horizontal support region cannot be identified, return \texttt{[]}.
\end{promptitems}
\end{promptsection}

\begin{promptsection}{Region localization}
\begin{promptitems}
\item Draw a tight bounding box around one coherent visible surface patch.
\item Exclude foreground objects, people, vegetation, reflective regions, and parts of the observing camera platform or robot.
\item Avoid boxes spanning surface boundaries, steps, or adjacent vertical faces.
\item Use integer coordinates \texttt{[x1, y1, x2, y2]} normalized to [0, 1000], with (0, 0) at the top-left and (1000, 1000) at the bottom-right.
\end{promptitems}
\end{promptsection}

\begin{promptsection}{Reasoning and response}
\begin{promptitems}
\item Use semantic and visible scene evidence to select and localize the region.
\item Do not predict a numerical surface normal, camera orientation, metric depth, camera height, or scale factor.
\item Return at most one JSON object with these fields:
\end{promptitems}
\end{promptsection}
\begin{unimetricpromptoutput}
\promptfield{plane\_type}{a concise lowercase surface category;}
\promptfield{plane\_bbox\_2d}{the normalized bounding box;}
\promptfield{confidence}{a value in [0, 1] indicating confidence in surface suitability;}
\promptfield{reason}{a brief explanation of why the region is a reliable support surface.}
\end{unimetricpromptoutput}
\begin{promptitems}
\item Return a JSON array only. If no suitable surface is visible, return \texttt{[]}.
\end{promptitems}
\end{unimetricpromptbox}
\caption{General support-surface prompt template. The VLM returns a
semantic label and an image region for the most reliable support surface;
its normal is then estimated from the reconstructed geometry
(\Cref{subsec:unimetric}).}
\label{fig:support_surface_prompt}
\end{figure}

\subsection{Implementation Settings}
\label{app:implementation_settings}

\paragraph{Semantic inference and segmentation.}
We use the \chg{Seed1.8~\citep{seed2026seed18}} VLM (\texttt{seed-1.8-max}) with temperature 0.2 and extended reasoning disabled. Each query receives the full RGB frame. Object and support-region boxes prompt SAM2.1 Hiera Large~\citep{ravi2024sam2} to produce one mask per region. Segmentation is performed independently on each image, so constructing the anchor observations does not require temporal mask propagation or persistent object identities.

\paragraph{Anchor reliability and scale estimation.}
Candidate observations are screened using their semantic confidence, visibility, and available reconstructed geometry. Geometric confidence filtering follows the selected \chg{GFM} configuration because the confidence outputs differ across geometry models. Where supplied by the prompt, category reliability and priority complement the observation confidence in determining its contribution to scale estimation. Category reliability uses four levels with weights 0.95, 0.85, 0.70, and 0.55; the weighting attributes are neutral when absent. The resulting weights express the relative reliability of individual size constraints. When a category priority rank $r_c$ is supplied, its weight is multiplied by $1-0.3\,\operatorname{clip}((r_c-1)/11,0,1)$. Missing category and priority factors are set to one.

The shared scale is obtained by minimizing the interval-based objective defined in the main text. We use one-dimensional golden-section search over the range bounded by the smallest lower endpoint and the largest upper endpoint of the retained scale intervals, with an absolute bracket-width tolerance of $10^{-5}$. \Cref{tab:implementation_settings} summarizes the shared inference and optimization settings.

\begin{table}[htbp]
\centering
\small
\caption{Shared inference and optimization settings. Optional category metadata is separate from the common prompt; geometric confidence filtering follows the \chg{GFM} configuration.}
\label{tab:implementation_settings}
\unimetrictablefont
\begin{tabular*}{\linewidth}{@{\extracolsep{\fill}}ll@{}}
\toprule
Component & Setting \\
\midrule

VLM & \texttt{seed-1.8-max} \\
Sampling temperature & 0.2 \\
Extended reasoning & Disabled \\
Segmentation model & SAM2.1 Hiera Large \\
Segmentation prompt and output & Bounding box; one mask per region \\
Category reliability weights & 0.95 / 0.85 / 0.70 / 0.55 \\
Scale optimizer & Golden-section search \\
Search tolerance & $10^{-5}$ \\
\bottomrule
\end{tabular*}
\end{table}

\paragraph{Metric-depth calibration baselines.}
\label{app:metric_depth_calibration}
\rev{\chg{Each baseline replaces the scale estimate of UniMetric with one derived
from a pretrained metric depth estimator, while the \chg{GFM} predictions
remain fixed. We use the public checkpoints of Depth
Pro~\citep{bochkovskii2025depthpro}, UniDepth~v2
(ViT-L/14)~\citep{piccinelli2025unidepthv2}, Depth
Anything~V2~\citep{yang2024depthanythingv2}, and
ZoeDepth-NK~\citep{bhat2023zoedepth} without fine-tuning on
MetricBench. For every frame of a sequence, the estimator predicts a metric
depth map $D^{\mathrm{md}}_t$ from the RGB image $I_t$ alone, without
ground-truth intrinsics. The per-sequence scale is the median depth ratio
over all frames,
\begin{equation}
s^{\mathrm{md}}=\operatorname*{median}_{t,\;\mathbf{u}\in\Omega_t}
\frac{D^{\mathrm{md}}_t(\mathbf{u})}{D_t(\mathbf{u})},
\label{eq:metric_depth_calibration}
\end{equation}
where $D_t$ is the \chg{GFM} depth of frame $t$ and $\Omega_t$ is the set of
pixels at which both depths are finite and positive. The median makes the
estimate robust to local failures of either depth map. The resulting factor
rescales pointmaps and camera translations exactly as $\hat{s}$ in
\Cref{eq:apply}, and all methods are evaluated under the same protocol
(\Cref{app:bench_metrics}). Oracle instead selects, for each reported
metric, the positive scale per sequence that minimizes that metric against
ground truth; because it uses ground truth, it serves only as an upper
bound for global rescaling.}}

\section{MetricBench}
\label{app:bench}

The platform, sensor, and subset-size summary in \Cref{subsec:metricbench}
is complemented here by the sequence-selection details, evaluation
definitions, and extended physical-scale statistics needed to interpret
the reported comparisons.

\subsection{Dataset Coverage and Subset Construction}
\label{app:bench_datasets}
\label{app:bench_subsets}
\label{app:bench_design}

\rev{MetricBench uses fixed subsets rather than the complete official splits
of its source datasets. \chg{Each frame provides an RGB image, a calibrated
metric point cloud, and the camera calibration and metric pose.} LiDAR supplies the reference point clouds for
ten subsets; DROID and Humanoid Everyday use depth-camera-derived point
clouds. Dataset metric poses supply the trajectory reference where ego pose
is applicable.

\emph{Driving and infrastructure.} KITTI~\citep{geiger2012we},
nuScenes~\citep{caesar2020nuscenes}, Waymo~\citep{sun2020scalability}, and
Argoverse~\citep{chang2019argoverse} contribute passenger-vehicle sequences.
V2X-Radar-I~\citep{yangv2x} and DAIR-V2X-I~\citep{yu2022dair} contribute
static roadside views, which retain long-range scene geometry without
camera motion. \chg{The Waymo subset combines 53 training and 96 validation
sequences, and the KITTI subset combines 84 raw-data
sequences~\citep{geiger2013vision} with 12
testing sequences.}

\emph{Embodied AI.} \chg{CEAR~\citep{zhu2024cear} contributes 20 indoor
Mini-Cheetah sequences, and JRDB~\citep{martin2021jrdb} 27 JackRabbot
sequences.} FusionPortable~\citep{jiao2022fusionportable} contains
two handheld sequences (1{,}489 frames), one legged-robot sequence (675
frames), and two UGV sequences (1{,}683 frames). DROID~\citep{khazatsky2024droid}
contains 45 Franka-manipulation sessions sampled from an external ZED view at
5\,Hz, with depth-derived point clouds. \chg{For Humanoid
Everyday~\citep{zhao2025humanoid}, we use 30 of the 45 G1 episodes
(3{,}949 frames), sampled at 5\,Hz from the 30\,Hz source. For
3EED~\citep{li20253eed}, we use all 12 drone sequences (2{,}368 frames) and
8 of the 13 quadruped sequences (1{,}476 frames) at 10\,Hz. The release
manifest lists every evaluated sequence.}

\chg{On CEAR and FusionPortable, the prompt additionally includes a
category-size table, and on 3EED, VGGT-$\Omega$ receives the dataset
intrinsics. \Cref{tab:bench_detail} lists the platforms and data sizes of
all subsets in the main text.}}

\subsection{Evaluation Protocol and Metrics}
\label{app:bench_metrics}
\label{app:bench_windows}

\rev{\chg{\textbf{Alignment.}
Where prediction and reference are expressed in different coordinate
frames, we register them with a rigid transform, which removes differences
of origin and orientation but leaves scale untouched. Neither similarity
fitting~\citep{umeyama1991least} nor per-image median
scaling~\citep{eigen2014depth} is applied, so an error in the predicted
scale enters every metric.

\textbf{Sequences.}
The unit of evaluation is the sequence, and UniMetric estimates one scale
per sequence (\Cref{app:impl}). \chg{Each sequence is evaluated in full, without
truncation or splitting.} The \emph{None} and \emph{UniMetric} conditions share the same model
outputs and frames and differ only in the scale applied to the points and
camera translations.

\textbf{Depth.}
Reference depth comes from the camera's depth sensor or from LiDAR returns
projected into the image with the dataset calibration; predicted depth is
the $z$-component of the pointmap in the same camera frame. Reference points
are kept when they are finite, lie in front of the camera and inside the
image, and are at most 100\,m away; when several project to one pixel at the
pointmap resolution, the nearest is used. A pixel is valid when both depths
are finite and positive. For predicted and reference depths $\hat d$ and
$d$, Abs Rel is the mean of $|\hat d-d|/d$ over valid pixels, and $\delta_1$
is the percentage of valid pixels with $\max(\hat d/d,\,d/\hat d)<1.25$
(\Cref{tab:exps_depth_main,tab:exps_depth_autonomous_driving_infrastructure,tab:exps_depth_embodied_ai}).
Depth MAE and RMSE in metres are reported in
\Cref{tab:supp_depth_mae_rmse_autonomous_driving_infrastructure,tab:supp_depth_mae_rmse_embodied_ai}.

\textbf{3D reconstruction.}
The predicted point cloud of a sequence is formed by placing its pointmaps
in a common frame with the predicted poses, and the reference cloud from the
sensor returns with the dataset calibration and reference poses. Both use
the same samples and 100\,m range as the depth evaluation and are expressed
in the first camera frame of the sequence. After rigid $SE(3)$ registration,
obtained in closed form from paired predicted and reference points and
estimating rotation and translation only, CD is the mean of the two directed
nearest-neighbour distances, in metres (\Cref{tab:exps_3d_reconstruction});
no further subsampling is applied. Static-camera subsets are fused with
their fixed reference poses and evaluated in the same way.

\textbf{Ego pose.}
ATE Mean is the mean Euclidean distance, in metres, between predicted and
reference camera positions after rigid alignment of the trajectories with
scale fixed to one (\Cref{tab:exps_ego_pose}). The static roadside subsets
are evaluated for depth and reconstruction only.

\textbf{Aggregation.}
Depth metrics are computed over the valid pixels of each frame and averaged
with equal weight over the valid frames of a subset. CD is computed per
sequence and averaged with equal weight over sequences. For ATE Mean,
each sequence is rigidly aligned independently, and the position errors
are pooled over all valid trajectory frames in the subset, weighting
each sequence by its number of valid frames.
Frames or sequences without valid data are excluded from the means. Suite
summaries weight subsets equally: six subsets per suite for depth and CD,
and four driving and six embodied subsets for ATE
(\Cref{tab:exps_summary_suite}). Reported error reductions compare means
before and after correction over the stated model--subset pairs,
$100\,(1-\bar e_{\mathrm{UniMetric}}/\bar e_{\mathrm{None}})\%$, and counts
of improved pairs are taken at the precision shown in the tables.}}

\subsection{Physical Scale Statistics}
\label{app:bench_scale}

\rev{\chg{\Cref{tab:bench_scale} quantifies the physical range of the evaluated
subsets. Each reference point cloud is projected into its camera image with
the dataset calibration, keeping depths in $(0,100]$\,m and the nearest
return per pixel; median and P90 depth are then computed over
\chg{all projected samples of each subset}.
Trajectory length is the sum of the translation increments along a
sequence, averaged with equal weight over sequences.}

Median reference depth ranges from 0.695\,m for Humanoid Everyday to
36.865\,m for V2X-Radar-I, while P90 reaches 81.245\,m on DAIR-V2X-I. \chg{The
subsets thus span} more than a $50\times$ change in characteristic
depth, from tabletop and humanoid manipulation to roadside perception.
Mean moving-camera trajectories range from 1.305\,m to 251.145\,m; \chg{the
DROID external camera is stationary, and the two roadside cameras are
static.}}

\begin{table}[htbp]
\centering
\caption{\rev{\textbf{Physical scale of each subset.} Median and
90th-percentile (P90) projected reference depth and mean metric trajectory
length \chg{per sequence}. Dashes mark static roadside cameras excluded
from ego-pose evaluation.}}
\label{tab:bench_scale}
\unimetrictablefont
\setlength{\tabcolsep}{6pt}
\begin{tabular*}{\linewidth}{@{\extracolsep{\fill}}lrrr@{}}
\toprule
\textbf{Dataset} & \shortstack{\textbf{Median depth}\\\textbf{(m)}} & \shortstack{\textbf{P90 depth}\\\textbf{(m)}} & \shortstack{\chg{\textbf{Trajectory / seq.}}\\\textbf{(m)}} \\
\midrule
KITTI & 11.755 & 29.345 & 167.160 \\
nuScenes & 11.585 & 40.495 & 94.325 \\
Waymo & 19.095 & 47.595 & 135.491 \\
Argoverse & 21.215 & 53.165 & 109.193 \\
V2X-Radar-I & 36.865 & 68.115 & -- \\
DAIR-V2X-I & 33.655 & 81.245 & -- \\
\midrule
CEAR & 2.895 & 8.195 & 55.585 \\
JRDB & 4.655 & 14.805 & 14.697 \\
FusionPortable & 6.505 & 20.395 & 251.145 \\
DROID & 0.885 & 2.185 & 0.000 \\
Humanoid Everyday & 0.695 & 1.425 & 1.305 \\
3EED & 10.035 & 23.965 & 163.352 \\
\bottomrule
\end{tabular*}
\end{table}

\FloatBarrier
\begingroup
\makeatletter
\setlength{\@fptop}{0pt}
\setlength{\@fpsep}{12pt}
\setlength{\@fpbot}{0pt plus 1fil}
\makeatother
\setlength{\textfloatsep}{10pt plus 2pt minus 2pt}
\setlength{\floatsep}{8pt plus 2pt minus 2pt}
\setlength{\intextsep}{8pt plus 2pt minus 2pt}
\interlinepenalty=0
\section{Supplementary Quantitative Results}
\label{sec:supplementary_material}

Complementing \Cref{subsec:main_results}, this section reports complete
per-subset depth threshold accuracy, depth errors in metres, and geometric
errors at fixed reference correspondences. These metrics characterize
aspects of the corrected predictions beyond the main comparisons.

\subsection{Depth Accuracy and Errors in Metres}
\label{app:results_depth}

\Crefrange{tab:exps_depth_autonomous_driving_infrastructure}{tab:exps_depth_embodied_ai}
report both Abs Rel and $\delta_1$ for each model--subset pair. Both
metrics improve in all 108 pairs, showing that the reduction in average
relative error is accompanied by more predictions within the accuracy
threshold.

\chg{Threshold accuracy measures the fraction of depths within a factor of
1.25 of the reference, complementing the mean error measured by Abs Rel.
On the driving
and roadside subsets, eight of the nine \chg{GFMs} place at most 0.5\% of
pixels within a factor of 1.25 of the reference depth, with Abs Rel between 0.881 and
0.985
(\Cref{tab:exps_depth_autonomous_driving_infrastructure,tab:exps_depth_embodied_ai}).
A single scale fitted to the reference removes most of this error: for
VGGT-$\Omega$ on nuScenes, it lowers Abs Rel from 0.979 to 0.087
(\Cref{tab:ablation}). UniMetric reaches 0.152 on the same sequences without
access to the reference, and it improves all 108 model--subset pairs.}

\begin{table}[H]
\centering
\caption{\textbf{Quantitative depth estimation on six autonomous driving benchmarks.}
Abs Rel (absolute relative error; lower is better) measures relative depth error;
$\delta_1$ is the fraction of depths within a factor of 1.25 of reference depth, reported in percent (higher is better).}

\unimetrictablefont
\setlength{\tabcolsep}{1.6pt}
\renewcommand{\arraystretch}{0.95}

\begin{tabular*}{\linewidth}{@{\extracolsep{\fill}}llcccccccccccc@{}}
\toprule
\multirow{2}{*}{\textbf{Model}}
& \multirow{2}{*}{\textbf{Scale}}
& \multicolumn{2}{c}{\textbf{KITTI}}
& \multicolumn{2}{c}{\textbf{nuScenes}}
& \multicolumn{2}{c}{\textbf{Waymo}}
& \multicolumn{2}{c}{\textbf{Argoverse}}
& \multicolumn{2}{c}{\textbf{V2X-Radar-I}}
& \multicolumn{2}{c}{\textbf{DAIR-V2X-I}} \\
\cmidrule(lr){3-4}
\cmidrule(lr){5-6}
\cmidrule(lr){7-8}
\cmidrule(lr){9-10}
\cmidrule(lr){11-12}
\cmidrule(lr){13-14}
& \textbf{Recovery}
& Abs Rel$\mathord{\downarrow}$ & $\delta_1(\%)\mathord{\uparrow}$
& Abs Rel$\mathord{\downarrow}$ & $\delta_1(\%)\mathord{\uparrow}$
& Abs Rel$\mathord{\downarrow}$ & $\delta_1(\%)\mathord{\uparrow}$
& Abs Rel$\mathord{\downarrow}$ & $\delta_1(\%)\mathord{\uparrow}$
& Abs Rel$\mathord{\downarrow}$ & $\delta_1(\%)\mathord{\uparrow}$
& Abs Rel$\mathord{\downarrow}$ & $\delta_1(\%)\mathord{\uparrow}$ \\
\midrule

\multirow{2}{*}{VGGT}
& None & 0.974 & 0.0 & 0.983 & 0.0 & 0.983 & 0.0 & 0.974 & 0.0 & 0.943 & 0.0 & 0.965 & 0.0 \\
& \textbf{UniMetric} & \textbf{0.722} & \textbf{46.7} & \textbf{0.316} & \textbf{57.7} & \textbf{0.311} & \textbf{42.9} & \textbf{0.311} & \textbf{51.1} & \textbf{0.392} & \textbf{45.6} & \textbf{0.866} & \textbf{49.3} \\
\midrule

\multirow{2}{*}{VGGT-$\Omega$}
& None & 0.971 & 0.0 & 0.979 & 0.0 & 0.983 & 0.0 & 0.972 & 0.0 & 0.949 & 0.0 & 0.966 & 0.0 \\
& \textbf{UniMetric} & \textbf{0.347} & \textbf{58.2} & \textbf{0.152} & \textbf{82.0} & \textbf{0.154} & \textbf{77.9} & \textbf{0.244} & \textbf{67.4} & \textbf{0.123} & \textbf{86.5} & \textbf{0.542} & \textbf{71.0} \\
\midrule

\multirow{2}{*}{$\pi^3$}
& None & 0.190 & 64.7 & 0.297 & 20.3 & 0.165 & 68.7 & 0.282 & 25.3 & 0.341 & 18.2 & 0.704 & 0.8 \\
& \textbf{UniMetric} & \textbf{0.188} & \textbf{73.3} & \textbf{0.098} & \textbf{93.0} & \textbf{0.095} & \textbf{91.1} & \textbf{0.140} & \textbf{84.6} & \textbf{0.147} & \textbf{76.3} & \textbf{0.471} & \textbf{65.3} \\
\midrule

\multirow{2}{*}{DA-3}
& None & 0.967 & 0.0 & 0.948 & 0.0 & 0.962 & 0.0 & 0.937 & 0.0 & 0.922 & 0.0 & 0.956 & 0.0 \\
& \textbf{UniMetric} & \textbf{0.372} & \textbf{53.7} & \textbf{0.331} & \textbf{43.2} & \textbf{0.162} & \textbf{76.2} & \textbf{0.425} & \textbf{32.7} & \textbf{0.168} & \textbf{77.5} & \textbf{0.531} & \textbf{51.6} \\
\midrule

\multirow{2}{*}{Scal3R}
& None & 0.966 & 0.0 & 0.971 & 0.0 & 0.974 & 0.0 & 0.959 & 0.0 & 0.938 & 0.0 & 0.967 & 0.0 \\
& \textbf{UniMetric} & \textbf{0.231} & \textbf{62.5} & \textbf{0.234} & \textbf{56.1} & \textbf{0.241} & \textbf{43.7} & \textbf{0.287} & \textbf{42.4} & \textbf{0.411} & \textbf{43.2} & \textbf{0.856} & \textbf{45.7} \\
\midrule

\multirow{2}{*}{LingBot-Map}
& None & 0.958 & 0.0 & 0.971 & 0.0 & 0.975 & 0.0 & 0.964 & 0.0 & 0.952 & 0.0 & 0.973 & 0.0 \\
& \textbf{UniMetric} & \textbf{0.244} & \textbf{57.5} & \textbf{0.272} & \textbf{47.9} & \textbf{0.325} & \textbf{31.0} & \textbf{0.343} & \textbf{38.1} & \textbf{0.351} & \textbf{32.3} & \textbf{0.658} & \textbf{16.2} \\
\midrule

\multirow{2}{*}{4RC}
& None & 0.973 & 0.0 & 0.979 & 0.0 & 0.985 & 0.0 & 0.976 & 0.0 & 0.938 & 0.0 & 0.960 & 0.0 \\
& \textbf{UniMetric} & \textbf{0.326} & \textbf{50.4} & \textbf{0.310} & \textbf{39.5} & \textbf{0.192} & \textbf{74.2} & \textbf{0.480} & \textbf{24.0} & \textbf{0.151} & \textbf{75.8} & \textbf{0.524} & \textbf{54.0} \\
\midrule

\multirow{2}{*}{DVLT}
& None & 0.957 & 0.0 & 0.968 & 0.0 & 0.974 & 0.0 & 0.962 & 0.0 & 0.881 & 0.0 & 0.926 & 0.5 \\
& \textbf{UniMetric} & \textbf{0.435} & \textbf{10.4} & \textbf{0.220} & \textbf{51.4} & \textbf{0.156} & \textbf{78.2} & \textbf{0.245} & \textbf{54.3} & \textbf{0.171} & \textbf{76.8} & \textbf{0.576} & \textbf{16.4} \\
\midrule

\multirow{2}{*}{STream3R}
& None & 0.960 & 0.0 & 0.971 & 0.0 & 0.970 & 0.0 & 0.961 & 0.0 & 0.956 & 0.0 & 0.970 & 0.0 \\
& \textbf{UniMetric} & \textbf{0.515} & \textbf{41.6} & \textbf{0.205} & \textbf{60.8} & \textbf{0.242} & \textbf{52.4} & \textbf{0.250} & \textbf{57.2} & \textbf{0.497} & \textbf{45.5} & \textbf{0.685} & \textbf{58.6} \\

\bottomrule
\end{tabular*}
\label{tab:exps_depth_autonomous_driving_infrastructure}

\end{table}

\begin{table}[H]
\centering
\caption{\textbf{Quantitative depth estimation on six embodied AI benchmarks.}
Abs Rel (absolute relative error; lower is better) measures relative depth error;
$\delta_1$ is the fraction of depths within a factor of 1.25 of reference depth,
reported in percent (higher is better).}

\unimetrictablefont
\setlength{\tabcolsep}{1.3pt}
\renewcommand{\arraystretch}{0.95}

\begin{tabular*}{\linewidth}{@{\extracolsep{\fill}}llcccccccccccc@{}}
\toprule
\multirow{2}{*}{\textbf{Model}}
& \multirow{2}{*}{\textbf{Scale}}
& \multicolumn{2}{c}{\textbf{CEAR}}
& \multicolumn{2}{c}{\textbf{JRDB}}
& \multicolumn{2}{c}{\textbf{FusionPortable}}
& \multicolumn{2}{c}{\textbf{DROID}}
& \multicolumn{2}{c}{\textbf{Humanoid Everyday}}
& \multicolumn{2}{c}{\textbf{3EED}} \\
\cmidrule(lr){3-4}
\cmidrule(lr){5-6}
\cmidrule(lr){7-8}
\cmidrule(lr){9-10}
\cmidrule(lr){11-12}
\cmidrule(lr){13-14}
& \textbf{Recovery}
& Abs Rel$\mathord{\downarrow}$ & $\delta_1(\%)\mathord{\uparrow}$
& Abs Rel$\mathord{\downarrow}$ & $\delta_1(\%)\mathord{\uparrow}$
& Abs Rel$\mathord{\downarrow}$ & $\delta_1(\%)\mathord{\uparrow}$
& Abs Rel$\mathord{\downarrow}$ & $\delta_1(\%)\mathord{\uparrow}$
& Abs Rel$\mathord{\downarrow}$ & $\delta_1(\%)\mathord{\uparrow}$
& Abs Rel$\mathord{\downarrow}$ & $\delta_1(\%)\mathord{\uparrow}$ \\
\midrule

\multirow{2}{*}{VGGT}
& None & 0.850 & 0.1 & 0.805 & 0.2 & 0.949 & 0.2 & 0.439 & 33.1 & 0.359 & 35.5 & 0.936 & 0.1 \\
& \textbf{UniMetric} & \textbf{0.551} & \textbf{43.7} & \textbf{0.520} & \textbf{35.1} & \textbf{0.676} & \textbf{18.2} & \textbf{0.332} & \textbf{40.9} & \textbf{0.335} & \textbf{40.4} & \textbf{0.534} & \textbf{34.9} \\
\midrule

\multirow{2}{*}{VGGT-$\Omega$}
& None & 0.786 & 0.4 & 0.790 & 0.5 & 0.885 & 0.9 & 0.294 & 42.6 & 0.311 & 38.2 & 0.937 & 0.1 \\
& \textbf{UniMetric} & \textbf{0.538} & \textbf{65.3} & \textbf{0.312} & \textbf{54.9} & \textbf{0.331} & \textbf{52.2} & \textbf{0.284} & \textbf{51.5} & \textbf{0.221} & \textbf{60.1} & \textbf{0.343} & \textbf{50.2} \\
\midrule

\multirow{2}{*}{$\pi^3$}
& None & 0.328 & 47.4 & 0.310 & 39.0 & 0.330 & 31.8 & 0.315 & 49.5 & 1.009 & 4.3 & 0.354 & 32.9 \\
& \textbf{UniMetric} & \textbf{0.322} & \textbf{62.0} & \textbf{0.253} & \textbf{68.8} & \textbf{0.219} & \textbf{63.2} & \textbf{0.221} & \textbf{59.2} & \textbf{0.480} & \textbf{34.1} & \textbf{0.295} & \textbf{58.8} \\
\midrule

\multirow{2}{*}{DA-3}
& None & 0.754 & 0.5 & 0.788 & 0.6 & 0.894 & 0.3 & 0.335 & 44.8 & 0.482 & 44.2 & 0.915 & 0.1 \\
& \textbf{UniMetric} & \textbf{0.406} & \textbf{55.5} & \textbf{0.345} & \textbf{49.7} & \textbf{0.358} & \textbf{37.2} & \textbf{0.327} & \textbf{45.4} & \textbf{0.364} & \textbf{52.5} & \textbf{0.436} & \textbf{36.8} \\
\midrule

\multirow{2}{*}{Scal3R}
& None & 0.763 & 0.4 & 0.770 & 2.1 & 0.904 & 0.1 & 0.338 & 33.2 & 0.320 & 38.9 & 0.898 & 0.1 \\
& \textbf{UniMetric} & \textbf{0.361} & \textbf{30.5} & \textbf{0.528} & \textbf{29.6} & \textbf{0.478} & \textbf{29.1} & \textbf{0.336} & \textbf{48.9} & \textbf{0.263} & \textbf{49.2} & \textbf{0.489} & \textbf{32.1} \\
\midrule

\multirow{2}{*}{LingBot-Map}
& None & 0.749 & 0.5 & 0.799 & 0.5 & 0.887 & 0.2 & 0.356 & 28.1 & 0.381 & 19.1 & 0.929 & 0.1 \\
& \textbf{UniMetric} & \textbf{0.283} & \textbf{34.5} & \textbf{0.338} & \textbf{40.0} & \textbf{0.403} & \textbf{37.5} & \textbf{0.297} & \textbf{47.4} & \textbf{0.282} & \textbf{36.8} & \textbf{0.347} & \textbf{47.6} \\
\midrule

\multirow{2}{*}{4RC}
& None & 0.777 & 0.3 & 0.789 & 0.5 & 0.906 & 0.2 & 0.324 & 35.2 & 0.436 & 32.2 & 0.932 & 0.0 \\
& \textbf{UniMetric} & \textbf{0.579} & \textbf{63.7} & \textbf{0.287} & \textbf{54.5} & \textbf{0.370} & \textbf{39.8} & \textbf{0.302} & \textbf{46.3} & \textbf{0.270} & \textbf{46.0} & \textbf{0.340} & \textbf{43.8} \\
\midrule

\multirow{2}{*}{DVLT}
& None & 0.672 & 0.8 & 0.569 & 13.9 & 0.867 & 0.5 & 1.001 & 11.8 & 0.722 & 39.4 & 0.898 & 0.1 \\
& \textbf{UniMetric} & \textbf{0.448} & \textbf{42.2} & \textbf{0.392} & \textbf{43.0} & \textbf{0.447} & \textbf{33.3} & \textbf{0.264} & \textbf{53.2} & \textbf{0.348} & \textbf{55.8} & \textbf{0.538} & \textbf{26.1} \\
\midrule

\multirow{2}{*}{STream3R}
& None & 0.719 & 1.1 & 0.796 & 0.5 & 0.858 & 0.9 & 0.421 & 9.9 & 0.379 & 23.6 & 0.932 & 0.0 \\
& \textbf{UniMetric} & \textbf{0.380} & \textbf{47.0} & \textbf{0.330} & \textbf{52.7} & \textbf{0.734} & \textbf{15.5} & \textbf{0.241} & \textbf{59.7} & \textbf{0.317} & \textbf{34.6} & \textbf{0.449} & \textbf{28.7} \\

\bottomrule
\end{tabular*}
\label{tab:exps_depth_embodied_ai}

\end{table}

\chg{Depth MAE measures average absolute error, while RMSE gives greater
weight to large errors. These absolute errors also depend on the physical
extent of the scene.
Median reference depth varies from
0.7\,m on Humanoid Everyday to 37\,m on V2X-Radar-I (\Cref{tab:bench_scale}).
For VGGT-$\Omega$, the corrected MAE
is 3.061\,m on nuScenes and 0.291\,m on DROID
(\Cref{tab:supp_depth_mae_rmse_autonomous_driving_infrastructure,tab:supp_depth_mae_rmse_embodied_ai}).
The correction is small where a \chg{GFM} is already close to metric.
$\pi^3$ improves on all twelve subsets, on KITTI only from 0.190 to 0.188
Abs Rel.}

\begin{table}[H]
\centering
\caption{\textbf{Supplementary depth estimation on six autonomous driving benchmarks.}
MAE (mean absolute error) summarizes average metric-scale depth error, while RMSE
(root mean square error) emphasizes larger errors. Both are in \chg{metres}; lower is
better.}

\unimetrictablefont
\setlength{\tabcolsep}{0.8pt}
\renewcommand{\arraystretch}{0.85}

\begin{tabular*}{\linewidth}{@{\extracolsep{\fill}}llcccccccccccc@{}}
\toprule
\multirow{2}{*}{\textbf{Model}}
& \textbf{Scale}
& \multicolumn{2}{c}{\textbf{KITTI}}
& \multicolumn{2}{c}{\textbf{nuScenes}}
& \multicolumn{2}{c}{\textbf{Waymo}}
& \multicolumn{2}{c}{\textbf{Argoverse}}
& \multicolumn{2}{c}{\textbf{V2X-Radar-I}}
& \multicolumn{2}{c}{\textbf{DAIR-V2X-I}} \\
\cmidrule(lr){3-4}
\cmidrule(lr){5-6}
\cmidrule(lr){7-8}
\cmidrule(lr){9-10}
\cmidrule(lr){11-12}
\cmidrule(lr){13-14}
& \textbf{Recovery}
& MAE$\downarrow$ & RMSE$\downarrow$
& MAE$\downarrow$ & RMSE$\downarrow$
& MAE$\downarrow$ & RMSE$\downarrow$
& MAE$\downarrow$ & RMSE$\downarrow$
& MAE$\downarrow$ & RMSE$\downarrow$
& MAE$\downarrow$ & RMSE$\downarrow$ \\
\midrule

\multirow{2}{*}{VGGT}
& None & 14.798 & 17.982 & 17.519 & 22.928 & 23.318 & 27.824 & 26.115 & 31.689 & 38.772 & 42.720 & 39.580 & 46.124 \\
& \textbf{UniMetric} & \textbf{9.622} & \textbf{11.842} & \textbf{5.613} & \textbf{9.099} & \textbf{7.653} & \textbf{10.055} & \textbf{8.387} & \textbf{12.292} & \textbf{15.102} & \textbf{18.630} & \textbf{19.193} & \textbf{24.132} \\
\midrule

\multirow{2}{*}{VGGT-$\Omega$}
& None & 14.799 & 17.988 & 17.455 & 22.824 & 23.142 & 27.529 & 26.197 & 31.729 & 38.872 & 42.672 & 39.962 & 46.505 \\
& \textbf{UniMetric} & \textbf{5.140} & \textbf{9.442} & \textbf{3.061} & \textbf{6.885} & \textbf{3.726} & \textbf{6.568} & \textbf{6.675} & \textbf{11.954} & \textbf{5.029} & \textbf{8.210} & \textbf{9.063} & \textbf{13.800} \\
\midrule

\multirow{2}{*}{$\pi^3$}
& None & 3.010 & 4.899 & 5.659 & 8.776 & 4.085 & 6.110 & 7.854 & 11.170 & 14.543 & 17.074 & 24.687 & 28.306 \\
& \textbf{UniMetric} & \textbf{2.835} & \textbf{4.422} & \textbf{2.231} & \textbf{5.948} & \textbf{2.443} & \textbf{4.721} & \textbf{4.037} & \textbf{8.385} & \textbf{6.292} & \textbf{9.002} & \textbf{8.230} & \textbf{12.071} \\
\midrule

\multirow{2}{*}{DA-3}
& None & 14.810 & 18.014 & 16.914 & 22.198 & 23.140 & 27.592 & 25.565 & 31.048 & 37.794 & 41.514 & 39.857 & 46.268 \\
& \textbf{UniMetric} & \textbf{5.891} & \textbf{8.167} & \textbf{5.630} & \textbf{8.828} & \textbf{4.063} & \textbf{6.785} & \textbf{10.609} & \textbf{14.493} & \textbf{6.673} & \textbf{9.880} & \textbf{9.922} & \textbf{13.639} \\
\midrule

\multirow{2}{*}{Scal3R}
& None & 14.684 & 17.842 & 17.317 & 22.697 & 23.113 & 27.588 & 25.786 & 31.337 & 38.554 & 42.557 & 39.633 & 46.184 \\
& \textbf{UniMetric} & \textbf{3.435} & \textbf{5.291} & \textbf{4.470} & \textbf{7.728} & \textbf{5.810} & \textbf{7.939} & \textbf{8.249} & \textbf{12.418} & \textbf{16.885} & \textbf{21.900} & \textbf{19.775} & \textbf{25.199} \\
\midrule

\multirow{2}{*}{LingBot-Map}
& None & 14.643 & 17.787 & 17.284 & 22.622 & 23.155 & 27.619 & 25.964 & 31.450 & 38.976 & 42.779 & 40.114 & 46.617 \\
& \textbf{UniMetric} & \textbf{3.958} & \textbf{6.053} & \textbf{5.046} & \textbf{9.038} & \textbf{7.640} & \textbf{10.132} & \textbf{9.365} & \textbf{17.400} & \textbf{13.490} & \textbf{16.300} & \textbf{15.392} & \textbf{18.590} \\
\midrule

\multirow{2}{*}{4RC}
& None & 14.910 & 18.135 & 17.466 & 22.872 & 23.184 & 27.583 & 25.549 & 31.100 & 38.409 & 42.308 & 39.310 & 45.795 \\
& \textbf{UniMetric} & \textbf{5.371} & \textbf{7.876} & \textbf{5.013} & \textbf{7.362} & \textbf{4.700} & \textbf{6.878} & \textbf{11.283} & \textbf{14.139} & \textbf{7.101} & \textbf{11.347} & \textbf{9.067} & \textbf{11.163} \\
\midrule

\multirow{2}{*}{DVLT}
& None & 14.671 & 17.854 & 17.270 & 22.604 & 22.926 & 27.277 & 30.640 & 39.624 & 36.135 & 39.793 & 38.609 & 44.900 \\
& \textbf{UniMetric} & \textbf{6.705} & \textbf{8.775} & \textbf{4.405} & \textbf{7.591} & \textbf{3.760} & \textbf{5.625} & \textbf{6.380} & \textbf{10.207} & \textbf{6.895} & \textbf{10.417} & \textbf{13.439} & \textbf{15.143} \\
\midrule

\multirow{2}{*}{STream3R}
& None & 14.593 & 17.737 & 17.352 & 22.712 & 22.827 & 27.163 & 25.207 & 30.665 & 39.175 & 43.013 & 39.695 & 46.241 \\
& \textbf{UniMetric} & \textbf{7.237} & \textbf{9.612} & \textbf{4.215} & \textbf{7.460} & \textbf{5.799} & \textbf{8.014} & \textbf{6.840} & \textbf{10.449} & \textbf{18.424} & \textbf{21.583} & \textbf{13.137} & \textbf{18.876} \\

\bottomrule
\end{tabular*}
\label{tab:supp_depth_mae_rmse_autonomous_driving_infrastructure}
\end{table}

\begin{table}[H]
\centering
\caption{\textbf{Supplementary depth estimation on six embodied AI benchmarks.}
MAE (mean absolute error) summarizes average metric-scale depth error, while RMSE
(root mean square error) emphasizes larger errors. Both are in \chg{metres}; lower is
better. Bold marks the better result within each None/UniMetric pair.
DA-3 denotes \chg{Depth Anything 3}. Evaluation subsets are described in \Cref{app:bench_subsets}.}

\unimetrictablefont
\setlength{\tabcolsep}{0.8pt}
\renewcommand{\arraystretch}{0.85}

\begin{tabular*}{\linewidth}{@{\extracolsep{\fill}}llcccccccccccc@{}}
\toprule
\multirow{2}{*}{\textbf{Model}}
& \textbf{Scale}
& \multicolumn{2}{c}{\textbf{CEAR}}
& \multicolumn{2}{c}{\textbf{JRDB}}
& \multicolumn{2}{c}{\textbf{FusionPortable}}
& \multicolumn{2}{c}{\textbf{DROID}}
& \multicolumn{2}{c}{\textbf{Humanoid Everyday}}
& \multicolumn{2}{c}{\textbf{3EED}} \\
\cmidrule(lr){3-4}
\cmidrule(lr){5-6}
\cmidrule(lr){7-8}
\cmidrule(lr){9-10}
\cmidrule(lr){11-12}
\cmidrule(lr){13-14}
& \textbf{Recovery}
& MAE$\downarrow$ & RMSE$\downarrow$
& MAE$\downarrow$ & RMSE$\downarrow$
& MAE$\downarrow$ & RMSE$\downarrow$
& MAE$\downarrow$ & RMSE$\downarrow$
& MAE$\downarrow$ & RMSE$\downarrow$
& MAE$\downarrow$ & RMSE$\downarrow$ \\
\midrule

\multirow{2}{*}{VGGT}
& None & 3.633 & 4.641 & 6.386 & 8.146 & 9.811 & 12.805 & 0.508 & 0.695 & 0.347 & 0.472 & 13.116 & 16.615 \\
& \textbf{UniMetric} & \textbf{2.238} & \textbf{3.136} & \textbf{3.645} & \textbf{5.456} & \textbf{6.345} & \textbf{9.117} & \textbf{0.391} & \textbf{0.542} & \textbf{0.328} & \textbf{0.452} & \textbf{6.138} & \textbf{9.111} \\
\midrule

\multirow{2}{*}{VGGT-$\Omega$}
& None & 3.339 & 4.281 & 6.253 & 7.884 & 9.247 & 12.108 & 0.335 & 0.442 & 0.314 & 0.425 & 13.148 & 16.624 \\
& \textbf{UniMetric} & \textbf{1.891} & \textbf{2.836} & \textbf{2.329} & \textbf{3.951} & \textbf{2.665} & \textbf{4.607} & \textbf{0.291} & \textbf{0.393} & \textbf{0.256} & \textbf{0.374} & \textbf{4.475} & \textbf{7.481} \\
\midrule

\multirow{2}{*}{$\pi^3$}
& None & 1.446 & 2.331 & 2.664 & 4.218 & 3.388 & 5.125 & 0.300 & 0.382 & 0.825 & 0.932 & 5.191 & 8.218 \\
& \textbf{UniMetric} & \textbf{1.382} & \textbf{2.267} & \textbf{2.033} & \textbf{3.759} & \textbf{2.244} & \textbf{4.081} & \textbf{0.247} & \textbf{0.340} & \textbf{0.461} & \textbf{0.579} & \textbf{4.029} & \textbf{7.349} \\
\midrule

\multirow{2}{*}{DA-3}
& None & 3.257 & 4.213 & 6.190 & 7.838 & 9.315 & 12.210 & 0.331 & 0.434 & 0.377 & 0.492 & 12.869 & 16.363 \\
& \textbf{UniMetric} & \textbf{1.688} & \textbf{2.537} & \textbf{2.457} & \textbf{4.056} & \textbf{3.646} & \textbf{5.629} & \textbf{0.306} & \textbf{0.398} & \textbf{0.313} & \textbf{0.429} & \textbf{5.524} & \textbf{8.530} \\
\midrule

\multirow{2}{*}{Scal3R}
& None & 3.323 & 4.321 & 6.300 & 8.072 & 9.448 & 12.407 & 0.423 & 0.549 & 0.329 & 0.466 & 12.868 & 16.453 \\
& \textbf{UniMetric} & \textbf{1.726} & \textbf{2.707} & \textbf{3.448} & \textbf{5.213} & \textbf{5.091} & \textbf{7.437} & \textbf{0.334} & \textbf{0.434} & \textbf{0.300} & \textbf{0.444} & \textbf{5.797} & \textbf{8.931} \\
\midrule

\multirow{2}{*}{LingBot-Map}
& None & 3.224 & 4.156 & 6.262 & 7.896 & 9.264 & 12.138 & 0.457 & 0.585 & 0.377 & 0.505 & 12.986 & 16.446 \\
& \textbf{UniMetric} & \textbf{1.324} & \textbf{2.133} & \textbf{2.660} & \textbf{4.210} & \textbf{3.897} & \textbf{6.390} & \textbf{0.331} & \textbf{0.449} & \textbf{0.315} & \textbf{0.449} & \textbf{4.879} & \textbf{8.227} \\
\midrule

\multirow{2}{*}{4RC}
& None & 3.317 & 4.282 & 6.271 & 7.956 & 9.363 & 12.253 & 0.420 & 0.538 & 0.388 & 0.514 & 13.064 & 16.577 \\
& \textbf{UniMetric} & \textbf{2.124} & \textbf{3.102} & \textbf{2.433} & \textbf{4.167} & \textbf{3.793} & \textbf{5.849} & \textbf{0.326} & \textbf{0.431} & \textbf{0.302} & \textbf{0.436} & \textbf{4.697} & \textbf{7.720} \\
\midrule

\multirow{2}{*}{DVLT}
& None & 3.006 & 4.001 & 5.231 & 7.023 & 9.190 & 12.113 & 0.806 & 0.897 & 0.528 & 0.637 & 12.744 & 16.233 \\
& \textbf{UniMetric} & \textbf{1.687} & \textbf{2.504} & \textbf{2.887} & \textbf{4.631} & \textbf{4.281} & \textbf{6.584} & \textbf{0.289} & \textbf{0.431} & \textbf{0.317} & \textbf{0.432} & \textbf{6.681} & \textbf{9.500} \\
\midrule

\multirow{2}{*}{STream3R}
& None & 3.151 & 4.123 & 6.224 & 7.891 & 8.962 & 11.788 & 0.508 & 0.622 & 0.377 & 0.489 & 13.015 & 16.475 \\
& \textbf{UniMetric} & \textbf{1.455} & \textbf{2.263} & \textbf{2.752} & \textbf{4.458} & \textbf{7.916} & \textbf{10.834} & \textbf{0.264} & \textbf{0.370} & \textbf{0.325} & \textbf{0.436} & \textbf{6.075} & \textbf{9.185} \\

\bottomrule
\end{tabular*}
\label{tab:supp_depth_mae_rmse_embodied_ai}
\end{table}

Depth MAE and RMSE both decrease in all 108 model--subset pairs in
\Cref{tab:supp_depth_mae_rmse_autonomous_driving_infrastructure,tab:supp_depth_mae_rmse_embodied_ai}.
Thus, the gains in relative and threshold accuracy also extend to
absolute depth errors, including the larger discrepancies emphasized by
RMSE.
\FloatBarrier

\subsection{Corresponding-Point Geometry Errors}
\label{app:results_reconstruction}

\chg{To complement the CD results in \Cref{tab:exps_3d_reconstruction},
we report errors between corresponding points, obtained by sampling the predicted
pointmaps at the projected reference locations
(\Cref{tab:supp_pairwise_mae_rmse_autonomous_driving_infrastructure,tab:supp_pairwise_mae_rmse_embodied_ai}).
Unlike CD, this measure does not allow a point to match an unrelated surface.
Corresponding-point MAE decreases in all 108 pairs and RMSE in 104.}

\begin{table}[H]
\centering
\caption{\textbf{Supplementary 3D reconstruction on six autonomous driving benchmarks.}
MAE (mean absolute error) and RMSE (root mean square error) summarize Euclidean
distances between corresponding predicted and reference 3D points after rigid
alignment, without estimating any additional scale. RMSE emphasizes larger discrepancies.
Both are in metres; lower is better. Bold marks the better result within each None/UniMetric pair.
DA-3 denotes \chg{Depth Anything 3}. Evaluation subsets are described in \Cref{app:bench_subsets}.}

\unimetrictablefont
\setlength{\tabcolsep}{0.8pt}
\renewcommand{\arraystretch}{0.85}

\begin{tabular*}{\linewidth}{@{\extracolsep{\fill}}llcccccccccccc@{}}
\toprule
\multirow{2}{*}{\textbf{Model}}
& \textbf{Scale}
& \multicolumn{2}{c}{\textbf{KITTI}}
& \multicolumn{2}{c}{\textbf{nuScenes}}
& \multicolumn{2}{c}{\textbf{Waymo}}
& \multicolumn{2}{c}{\textbf{Argoverse}}
& \multicolumn{2}{c}{\textbf{V2X-Radar-I}}
& \multicolumn{2}{c}{\textbf{DAIR-V2X-I}} \\
\cmidrule(lr){3-4}
\cmidrule(lr){5-6}
\cmidrule(lr){7-8}
\cmidrule(lr){9-10}
\cmidrule(lr){11-12}
\cmidrule(lr){13-14}
& \textbf{Recovery}
& MAE$\downarrow$ & RMSE$\downarrow$
& MAE$\downarrow$ & RMSE$\downarrow$
& MAE$\downarrow$ & RMSE$\downarrow$
& MAE$\downarrow$ & RMSE$\downarrow$
& MAE$\downarrow$ & RMSE$\downarrow$
& MAE$\downarrow$ & RMSE$\downarrow$ \\
\midrule

\multirow{2}{*}{VGGT}
& None & 45.367 & 51.918 & 29.785 & 34.824 & 39.999 & 46.320 & 33.105 & 38.572 & 40.887 & 45.105 & 41.042 & 47.989 \\
& \textbf{UniMetric} & \textbf{22.942} & \textbf{26.723} & \textbf{8.332} & \textbf{11.292} & \textbf{14.638} & \textbf{17.650} & \textbf{9.581} & \textbf{12.927} & \textbf{15.938} & \textbf{19.585} & \textbf{20.931} & \textbf{26.192} \\
\midrule

\multirow{2}{*}{VGGT-$\Omega$}
& None & 45.317 & 51.897 & 29.458 & 34.420 & 38.892 & 45.109 & 33.079 & 38.518 & 40.955 & 45.016 & 41.446 & 48.391 \\
& \textbf{UniMetric} & \textbf{12.269} & \textbf{16.408} & \textbf{4.489} & \textbf{8.125} & \textbf{6.426} & \textbf{9.430} & \textbf{7.581} & \textbf{12.889} & \textbf{5.498} & \textbf{8.717} & \textbf{9.335} & \textbf{14.245} \\
\midrule

\multirow{2}{*}{$\pi^3$}
& None & 8.557 & 10.743 & 9.214 & 11.803 & 7.331 & 9.445 & 10.445 & 13.500 & 15.238 & 17.910 & 25.513 & 29.335 \\
& \textbf{UniMetric} & \textbf{6.877} & \textbf{9.062} & \textbf{3.260} & \textbf{6.846} & \textbf{4.297} & \textbf{6.535} & \textbf{4.942} & \textbf{9.199} & \textbf{6.683} & \textbf{9.450} & \textbf{8.574} & \textbf{12.562} \\
\midrule

\multirow{2}{*}{DA-3}
& None & 45.199 & 51.760 & 28.794 & 33.678 & 39.262 & 45.448 & 32.100 & 37.399 & 39.823 & 43.800 & 41.333 & 48.139 \\
& \textbf{UniMetric} & \textbf{18.880} & \textbf{22.642} & \textbf{9.668} & \textbf{12.612} & \textbf{7.448} & \textbf{10.288} & \textbf{13.337} & \textbf{17.179} & \textbf{7.090} & \textbf{10.461} & \textbf{10.371} & \textbf{14.223} \\
\midrule

\multirow{2}{*}{Scal3R}
& None & 45.001 & 51.501 & 29.503 & 34.503 & 39.679 & 45.953 & 32.685 & 38.108 & 40.606 & 44.883 & 41.095 & 48.048 \\
& \textbf{UniMetric} & \textbf{10.821} & \textbf{13.321} & \textbf{7.487} & \textbf{10.198} & \textbf{11.341} & \textbf{13.909} & \textbf{11.604} & \textbf{15.194} & \textbf{17.776} & \textbf{22.882} & \textbf{21.867} & \textbf{27.617} \\
\midrule

\multirow{2}{*}{LingBot-Map}
& None & 44.405 & 50.858 & 29.399 & 34.370 & 39.602 & 45.860 & 32.747 & 38.136 & 41.068 & 45.130 & 41.606 & 48.509 \\
& \textbf{UniMetric} & \textbf{12.330} & \textbf{14.879} & \textbf{8.046} & \textbf{11.470} & \textbf{15.133} & \textbf{18.518} & \textbf{12.086} & \textbf{20.910} & \textbf{14.317} & \textbf{17.291} & \textbf{16.040} & \textbf{19.331} \\
\midrule

\multirow{2}{*}{4RC}
& None & 45.390 & 51.986 & 29.509 & 34.490 & 38.957 & 45.183 & 33.989 & 39.565 & 40.483 & 44.653 & 40.765 & 47.650 \\
& \textbf{UniMetric} & \textbf{16.321} & \textbf{20.008} & \textbf{8.683} & \textbf{11.372} & \textbf{9.119} & \textbf{11.653} & \textbf{15.217} & \textbf{18.869} & \textbf{7.588} & \textbf{12.033} & \textbf{9.422} & \textbf{11.621} \\
\midrule

\multirow{2}{*}{DVLT}
& None & 44.868 & 51.402 & 29.215 & 34.144 & 38.645 & 44.824 & 33.540 & 39.041 & 38.035 & 41.945 & 39.975 & 46.627 \\
& \textbf{UniMetric} & \textbf{21.247} & \textbf{26.508} & \textbf{7.617} & \textbf{10.915} & \textbf{7.631} & \textbf{9.647} & \textbf{8.356} & \textbf{12.200} & \textbf{7.475} & \textbf{11.047} & \textbf{14.256} & \textbf{16.027} \\
\midrule

\multirow{2}{*}{STream3R}
& None & 45.101 & 51.606 & 29.219 & 34.159 & 38.388 & 44.543 & 33.354 & 38.821 & 41.293 & 45.394 & 41.165 & 48.115 \\
& \textbf{UniMetric} & \textbf{27.271} & \textbf{32.364} & \textbf{8.422} & \textbf{11.132} & \textbf{13.794} & \textbf{17.090} & \textbf{8.721} & \textbf{12.173} & \textbf{19.433} & \textbf{22.732} & \textbf{14.352} & \textbf{20.241} \\

\bottomrule
\end{tabular*}
\label{tab:supp_pairwise_mae_rmse_autonomous_driving_infrastructure}
\end{table}

\begin{table}[H]
\centering
\caption{\textbf{Supplementary 3D reconstruction on six embodied AI benchmarks.}
MAE (mean absolute error) and RMSE (root mean square error) summarize Euclidean
distances between corresponding predicted and reference 3D points after rigid
alignment, without estimating any additional scale. RMSE emphasizes larger discrepancies.
Both are in metres; lower is better. Bold marks the better result within each None/UniMetric pair.
DA-3 denotes \chg{Depth Anything 3}.}

\unimetrictablefont
\setlength{\tabcolsep}{0.8pt}
\renewcommand{\arraystretch}{0.85}

\begin{tabular*}{\linewidth}{@{\extracolsep{\fill}}llcccccccccccc@{}}
\toprule
\multirow{2}{*}{\textbf{Model}}
& \textbf{Scale}
& \multicolumn{2}{c}{\textbf{CEAR}}
& \multicolumn{2}{c}{\textbf{JRDB}}
& \multicolumn{2}{c}{\textbf{FusionPortable}}
& \multicolumn{2}{c}{\textbf{DROID}}
& \multicolumn{2}{c}{\textbf{Humanoid Everyday}}
& \multicolumn{2}{c}{\textbf{3EED}} \\
\cmidrule(lr){3-4}
\cmidrule(lr){5-6}
\cmidrule(lr){7-8}
\cmidrule(lr){9-10}
\cmidrule(lr){11-12}
\cmidrule(lr){13-14}
& \textbf{Recovery}
& MAE$\downarrow$ & RMSE$\downarrow$
& MAE$\downarrow$ & RMSE$\downarrow$
& MAE$\downarrow$ & RMSE$\downarrow$
& MAE$\downarrow$ & RMSE$\downarrow$
& MAE$\downarrow$ & RMSE$\downarrow$
& MAE$\downarrow$ & RMSE$\downarrow$ \\
\midrule

\multirow{2}{*}{VGGT}
& None & 6.951 & \textbf{7.776} & 7.059 & 8.507 & 37.490 & 42.829 & 0.521 & 0.672 & 0.438 & 0.592 & 14.534 & 17.992 \\
& \textbf{UniMetric} & \textbf{6.740} & 7.969 & \textbf{5.254} & \textbf{6.779} & \textbf{19.452} & \textbf{23.547} & \textbf{0.444} & \textbf{0.575} & \textbf{0.432} & \textbf{0.582} & \textbf{8.253} & \textbf{11.274} \\
\midrule

\multirow{2}{*}{VGGT-$\Omega$}
& None & 6.416 & 7.166 & 6.817 & 8.212 & 35.921 & 41.015 & 0.379 & 0.480 & 0.479 & 0.614 & 14.522 & 17.965 \\
& \textbf{UniMetric} & \textbf{4.934} & \textbf{6.054} & \textbf{3.538} & \textbf{5.095} & \textbf{5.374} & \textbf{6.896} & \textbf{0.307} & \textbf{0.423} & \textbf{0.445} & \textbf{0.583} & \textbf{6.786} & \textbf{9.606} \\
\midrule

\multirow{2}{*}{$\pi^3$}
& None & 4.273 & \textbf{4.951} & 3.828 & 5.323 & 10.405 & 12.636 & 0.257 & \textbf{0.359} & 0.724 & 0.866 & 8.537 & 11.349 \\
& \textbf{UniMetric} & \textbf{4.209} & 4.996 & \textbf{3.206} & \textbf{4.941} & \textbf{5.100} & \textbf{6.838} & \textbf{0.255} & 0.362 & \textbf{0.534} & \textbf{0.690} & \textbf{7.469} & \textbf{10.554} \\
\midrule

\multirow{2}{*}{DA-3}
& None & 6.392 & 7.155 & 6.814 & 8.207 & 35.999 & 41.132 & 0.378 & 0.470 & 0.450 & 0.597 & 14.420 & 17.861 \\
& \textbf{UniMetric} & \textbf{4.454} & \textbf{5.267} & \textbf{3.803} & \textbf{5.313} & \textbf{12.341} & \textbf{14.512} & \textbf{0.322} & \textbf{0.417} & \textbf{0.427} & \textbf{0.570} & \textbf{7.991} & \textbf{10.961} \\
\midrule

\multirow{2}{*}{Scal3R}
& None & 6.566 & 7.361 & 6.946 & 8.393 & 36.349 & 41.600 & 0.476 & 0.579 & 0.465 & 0.611 & 14.491 & 17.992 \\
& \textbf{UniMetric} & \textbf{4.988} & \textbf{5.755} & \textbf{5.077} & \textbf{6.689} & \textbf{14.451} & \textbf{18.029} & \textbf{0.318} & \textbf{0.428} & \textbf{0.449} & \textbf{0.597} & \textbf{10.121} & \textbf{13.116} \\
\midrule

\multirow{2}{*}{LingBot-Map}
& None & 6.282 & 7.010 & 6.797 & 8.173 & 36.236 & 41.375 & 0.495 & 0.599 & 0.491 & 0.648 & 14.384 & 17.805 \\
& \textbf{UniMetric} & \textbf{4.000} & \textbf{4.594} & \textbf{3.947} & \textbf{5.464} & \textbf{11.419} & \textbf{13.643} & \textbf{0.379} & \textbf{0.492} & \textbf{0.448} & \textbf{0.610} & \textbf{7.629} & \textbf{10.524} \\
\midrule

\multirow{2}{*}{4RC}
& None & 6.374 & 7.126 & 6.885 & 8.308 & 35.258 & 40.300 & 0.444 & 0.541 & 0.525 & 0.681 & 14.532 & 17.995 \\
& \textbf{UniMetric} & \textbf{5.026} & \textbf{6.138} & \textbf{3.618} & \textbf{5.252} & \textbf{7.921} & \textbf{9.660} & \textbf{0.323} & \textbf{0.428} & \textbf{0.461} & \textbf{0.619} & \textbf{8.840} & \textbf{11.649} \\
\midrule

\multirow{2}{*}{DVLT}
& None & 6.154 & 6.926 & 6.439 & 7.897 & 36.374 & 41.612 & 0.746 & 0.860 & 0.582 & 0.723 & 14.246 & 17.670 \\
& \textbf{UniMetric} & \textbf{4.645} & \textbf{5.497} & \textbf{4.129} & \textbf{5.699} & \textbf{11.175} & \textbf{13.738} & \textbf{0.333} & \textbf{0.474} & \textbf{0.476} & \textbf{0.617} & \textbf{7.496} & \textbf{10.259} \\
\midrule

\multirow{2}{*}{STream3R}
& None & 6.726 & 7.487 & 6.792 & 8.191 & 33.815 & \textbf{37.867} & 0.475 & 0.574 & 0.448 & 0.582 & 14.396 & 17.835 \\
& \textbf{UniMetric} & \textbf{6.115} & \textbf{6.955} & \textbf{3.905} & \textbf{5.474} & \textbf{30.240} & 43.410 & \textbf{0.280} & \textbf{0.397} & \textbf{0.418} & \textbf{0.548} & \textbf{8.489} & \textbf{11.521} \\

\bottomrule
\end{tabular*}
\label{tab:supp_pairwise_mae_rmse_embodied_ai}
\end{table}

\chg{MAE and RMSE need not improve together. RMSE gives greater weight to large
residuals, and a global scale does not independently adjust individual
points. For STream3R on FusionPortable, which retains the highest
Abs Rel on that subset after correction (0.734),
MAE decreases from 33.815 to 30.240\,m while RMSE increases from 37.867 to
43.410\,m. This example shows why the average reduction in metric error
should be considered alongside sensitivity to large residuals. Together
with the depth and trajectory results in \Cref{subsec:main_results}, these
measurements characterize both the gains from scale recovery and the
geometric errors that remain after applying one shared factor.}
\FloatBarrier
\endgroup

\clearpage
\raggedbottom
\section{Supplementary Qualitative Results}
\label{sec:supp_qualitative}
\label{app:qualitative}

\begingroup
\setlength{\intextsep}{4pt}
\setlength{\abovecaptionskip}{5pt}
\setlength{\belowcaptionskip}{0pt}
\subsection{Depth Visualization across Datasets}
\label{app:depth_visualizations}

\chg{\Crefrange{fig:supp-depth-nuscenes}{fig:supp-depth-3eed} extend the
comparisons in \Cref{fig:qualitative_main,fig:camera_pose_eight_models}
to every MetricBench subset, covering close-range manipulation, mobile
robots, autonomous driving, roadside capture, and aerial views.
They expose spatial differences across distances and model backbones.}

\chg{Rows denote GFMs; columns show RGB, sensor-derived reference depth,
original predictions (None), and UniMetric. A shared reference color range
makes depth changes visible. Abs Rel and $\delta_1$ are frame-level metrics;
caption averages are across the displayed GFMs. Scales are estimated over
the sequence.}

\chg{One shared scale preserves relative scene structure. It calibrates
physical distances but cannot correct errors at object boundaries or
inconsistencies between near and far surfaces inherited from the backbone.}

\begin{figure}[H]
  \centering
  \includegraphics[width=\linewidth]{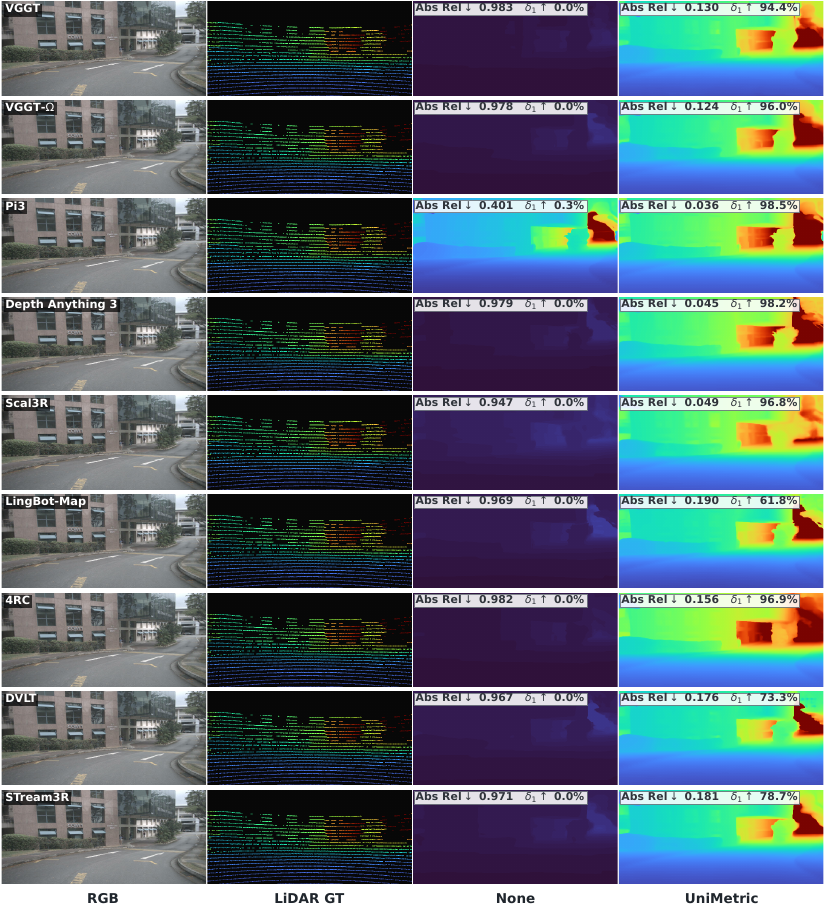}
  \caption{\chg{\textbf{Metric depth recovery on nuScenes (urban driving; 0--46\,m).}
    The scene contains a facade, hedges, and pedestrians at different distances.
    Before correction, no GFM exceeds 0.3\% $\delta_1$. A shared sequence-level
    scale raises the mean to 88.3\%, with $\pi^3$ reaching 98.5\%, while
    retaining the spatial structure of each prediction.}}
  \label{fig:supp-depth-nuscenes}
\end{figure}
\clearpage

\begin{figure}[H]
  \centering
  \includegraphics[width=\linewidth]{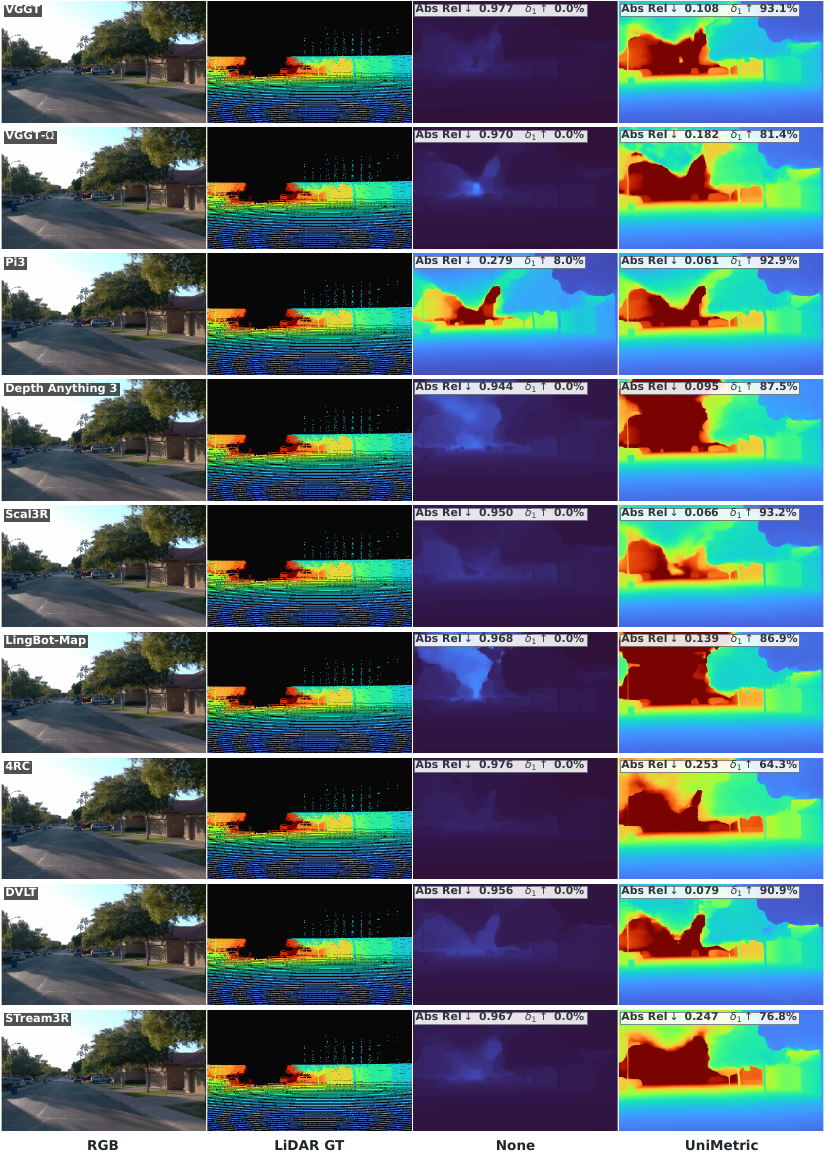}
  \caption{\chg{\textbf{Metric depth recovery on Waymo (residential street; 0--63\,m).}
    Parked vehicles, houses, and the receding roadway span near and far depths.
    A shared scale increases the mean frame-level $\delta_1$ from 0.9\% to
    85.2\%; VGGT improves from 0.0\% to 93.1\%.}}
  \label{fig:supp-depth-waymo}
\end{figure}
\clearpage

\begin{figure}[H]
  \centering
  \includegraphics[width=\linewidth]{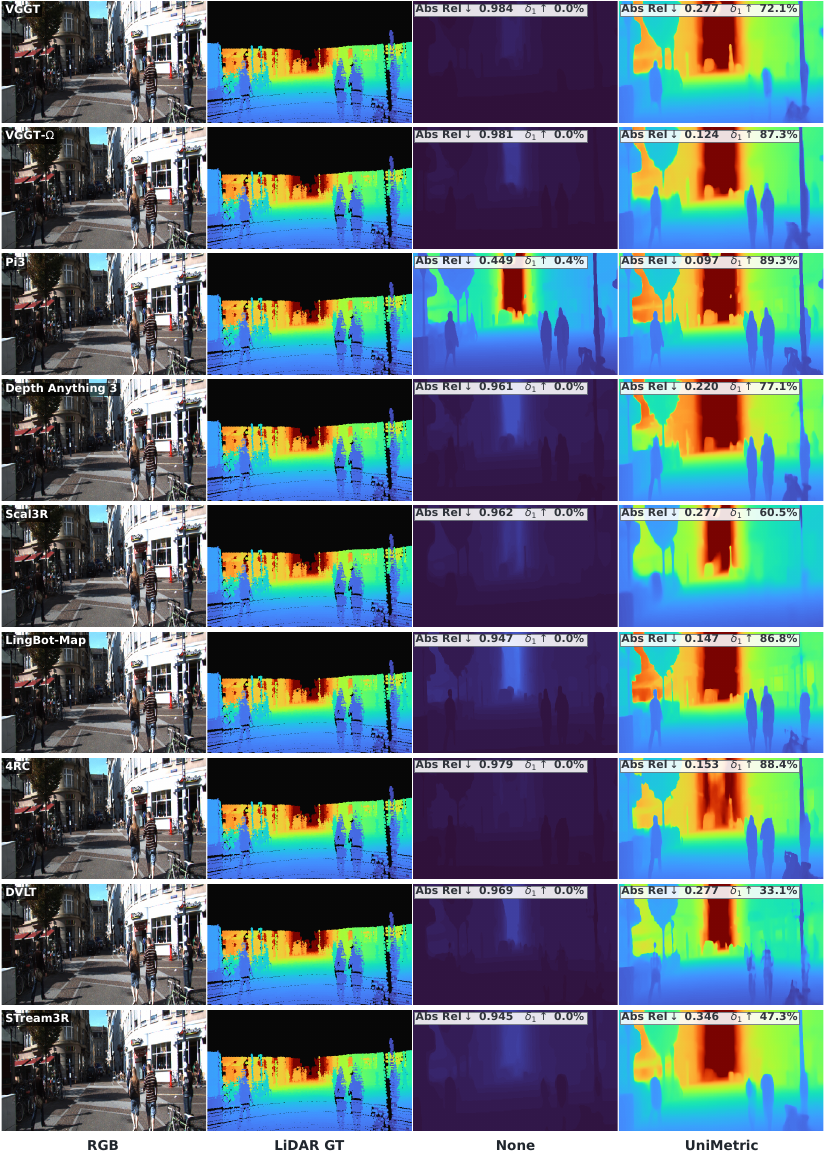}
  \caption{\chg{\textbf{Metric depth recovery on KITTI (shopping street; 0--44\,m).}
    Nearby pedestrians and bicycles introduce occlusions and depth discontinuities.
    The mean frame-level $\delta_1$ increases from 0.0\% to 71.3\%, and
    $\pi^3$ from 0.4\% to 89.3\%. Rescaling retains the predicted object boundaries.}}
  \label{fig:supp-depth-kitti}
\end{figure}
\clearpage

\begin{figure}[H]
  \centering
  \includegraphics[width=\linewidth]{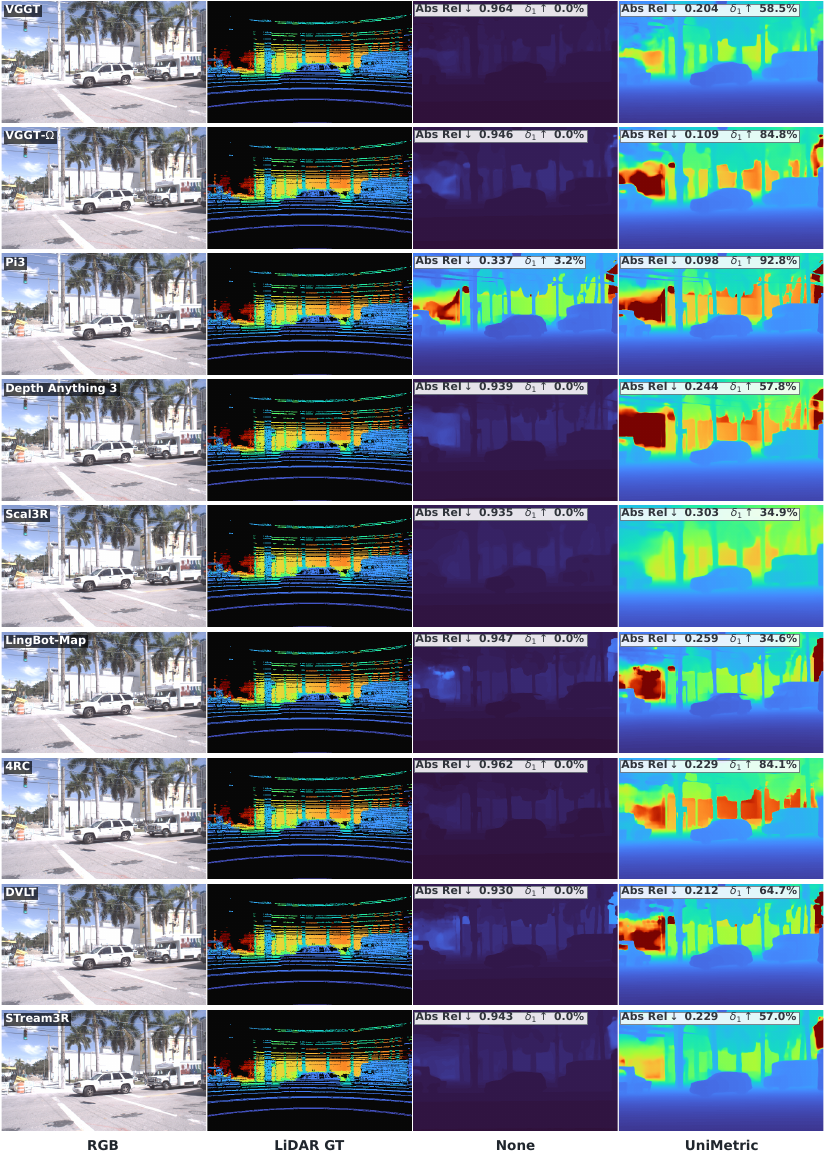}
  \caption{\chg{\textbf{Metric depth recovery on Argoverse (urban intersection; 0--70\,m).}
    Foreground vehicles and a background facade provide a comparison across distances.
    The mean frame-level $\delta_1$ increases from 0.4\% to 63.2\%;
    $\pi^3$ and VGGT-$\Omega$ reach 92.8\% and 84.8\%, respectively.}}
  \label{fig:supp-depth-argoverse}
\end{figure}
\clearpage

\begin{figure}[H]
  \centering
  \includegraphics[width=\linewidth]{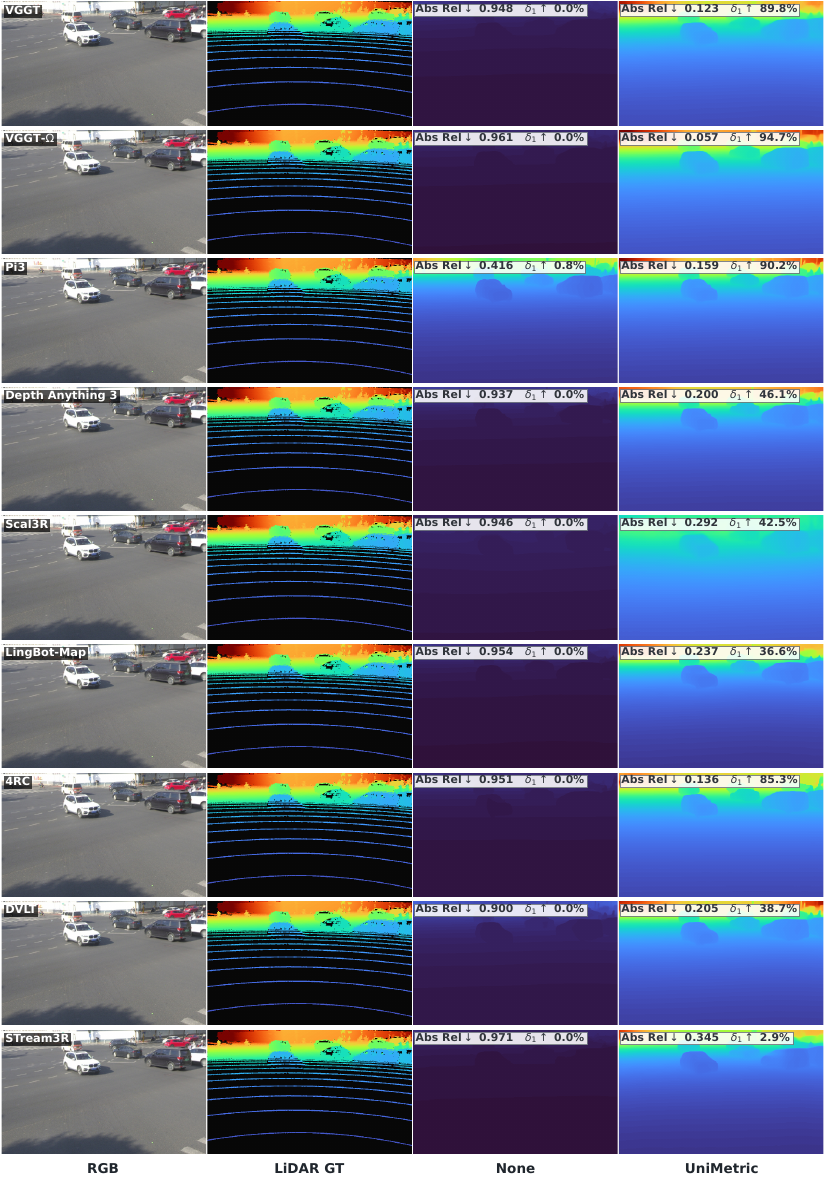}
  \caption{\chg{\textbf{Metric depth recovery on V2X-Radar-I (elevated roadside view; 0--86\,m).}
    Vehicles occupy a small part of a largely weakly textured road scene.
    The mean frame-level $\delta_1$ increases from 0.1\% to 58.5\%, and
    VGGT-$\Omega$ from 0.0\% to 94.7\%, illustrating scale recovery from
    an elevated viewpoint.}}
  \label{fig:supp-depth-v2x-radar-i}
\end{figure}
\clearpage

\begin{figure}[H]
  \centering
  \includegraphics[width=\linewidth]{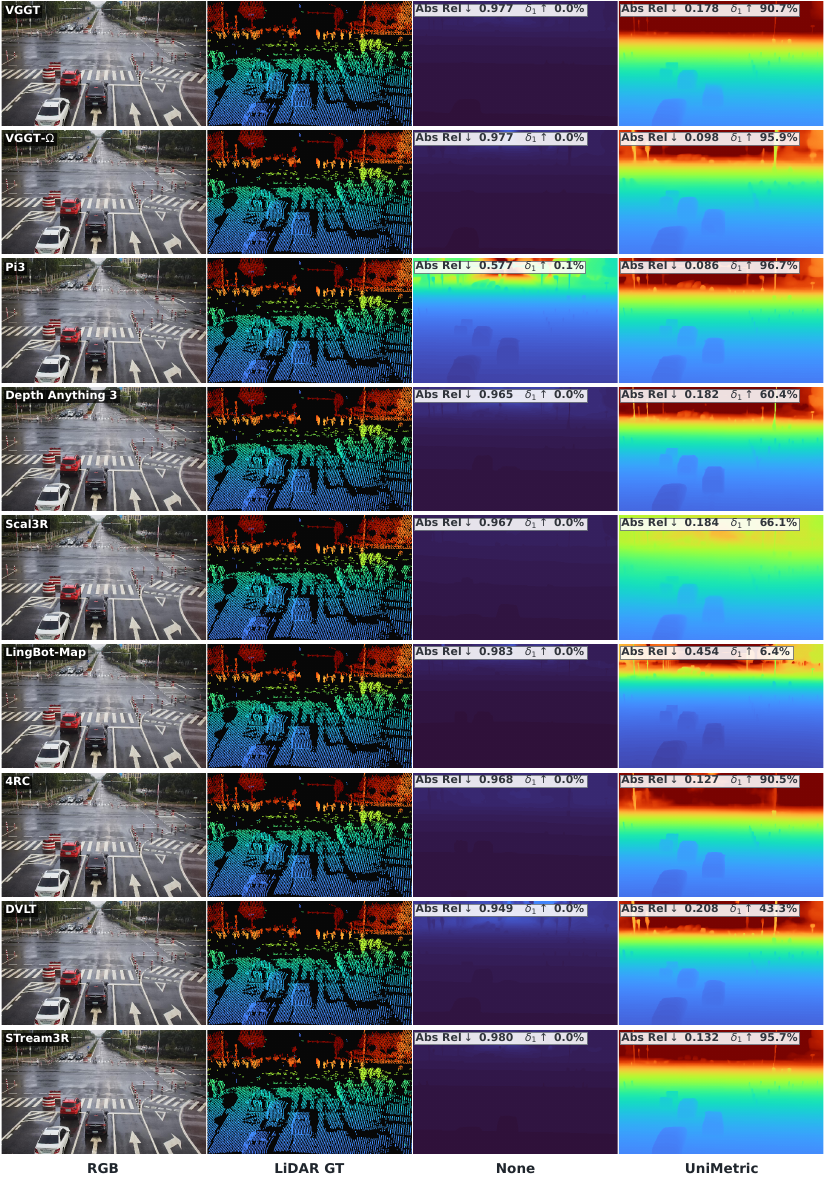}
  \caption{\chg{\textbf{Metric depth recovery on DAIR-V2X-I (roadside view; 0--99\,m).}
    Foreground cars and a distant tree-lined avenue span the widest depth range
    among these examples. The mean frame-level $\delta_1$ increases from
    0.0\% to 71.7\%; $\pi^3$ and STream3R reach 96.7\% and 95.7\%, respectively.}}
  \label{fig:supp-depth-dair-v2x-i}
\end{figure}
\clearpage

\begin{figure}[H]
  \centering
  \includegraphics[width=\linewidth]{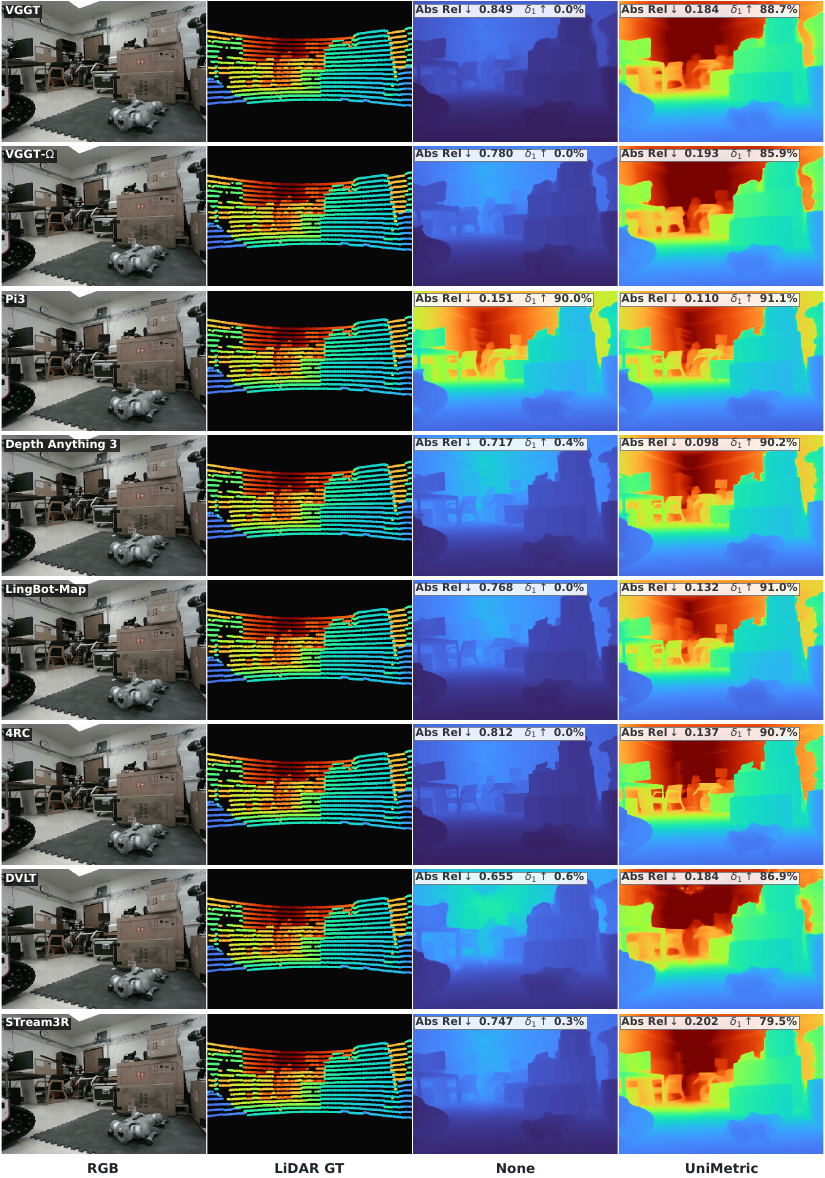}
  \caption{\chg{\textbf{Metric depth recovery on CEAR (indoor quadruped view; 0--5.5\,m).}
    A low viewpoint captures a robot, stacked boxes, and lab furniture.
    The mean frame-level $\delta_1$ increases from 11.4\% to 88.0\%, and
    DA-3 from 0.4\% to 90.2\%. For $\pi^3$, which is already close to metric,
    $\delta_1$ changes from 90.0\% to 91.1\%.}}
  \label{fig:supp-depth-cear}
\end{figure}
\clearpage

\begin{figure}[H]
  \centering
  \includegraphics[width=\linewidth]{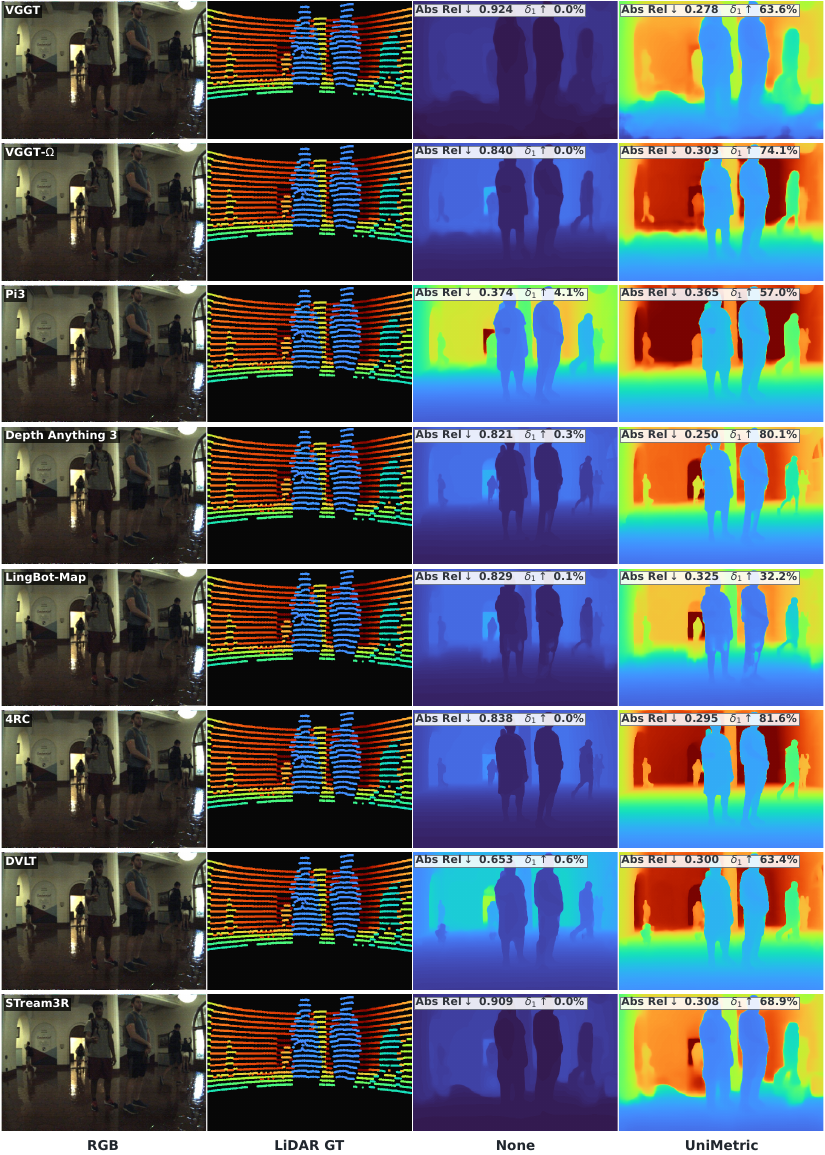}
  \caption{\chg{\textbf{Metric depth recovery on JRDB (wheeled robot in a hall; 0--11\,m).}
    People at different distances appear among columns and a reflective floor.
    The mean frame-level $\delta_1$ increases from 0.6\% to 65.1\%, with
    4RC and DA-3 reaching 81.6\% and 80.1\%, respectively.}}
  \label{fig:supp-depth-jrdb}
\end{figure}
\clearpage

\begin{figure}[H]
  \centering
  \includegraphics[width=\linewidth]{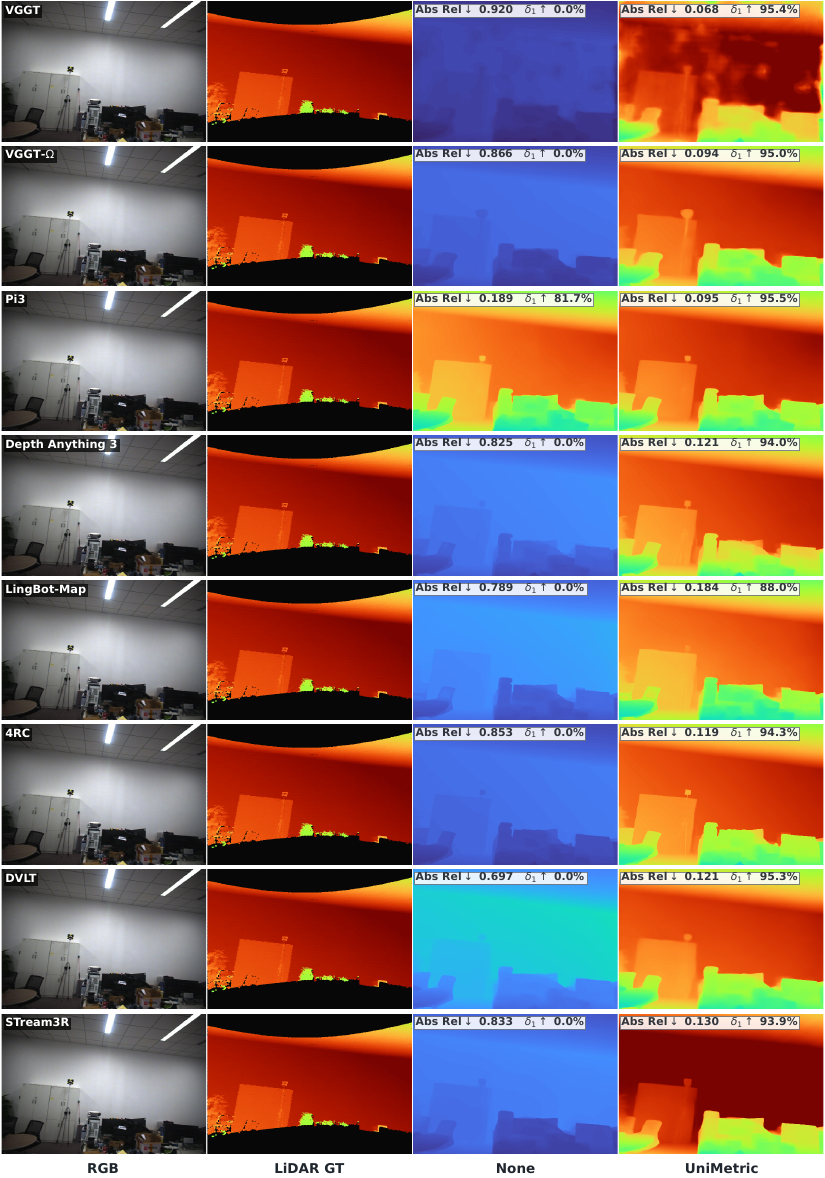}
  \caption{\chg{\textbf{Metric depth recovery on FusionPortable (handheld office capture; 0--4.9\,m).}
    Cabinets, monitors, and equipment appear beneath largely blank walls and ceiling.
    The mean frame-level $\delta_1$ increases from 10.2\% to 93.9\%, with
    every displayed GFM reaching at least 88.0\%.}}
  \label{fig:supp-depth-fusionportable}
\end{figure}
\clearpage

\begin{figure}[H]
  \centering
  \includegraphics[width=\linewidth]{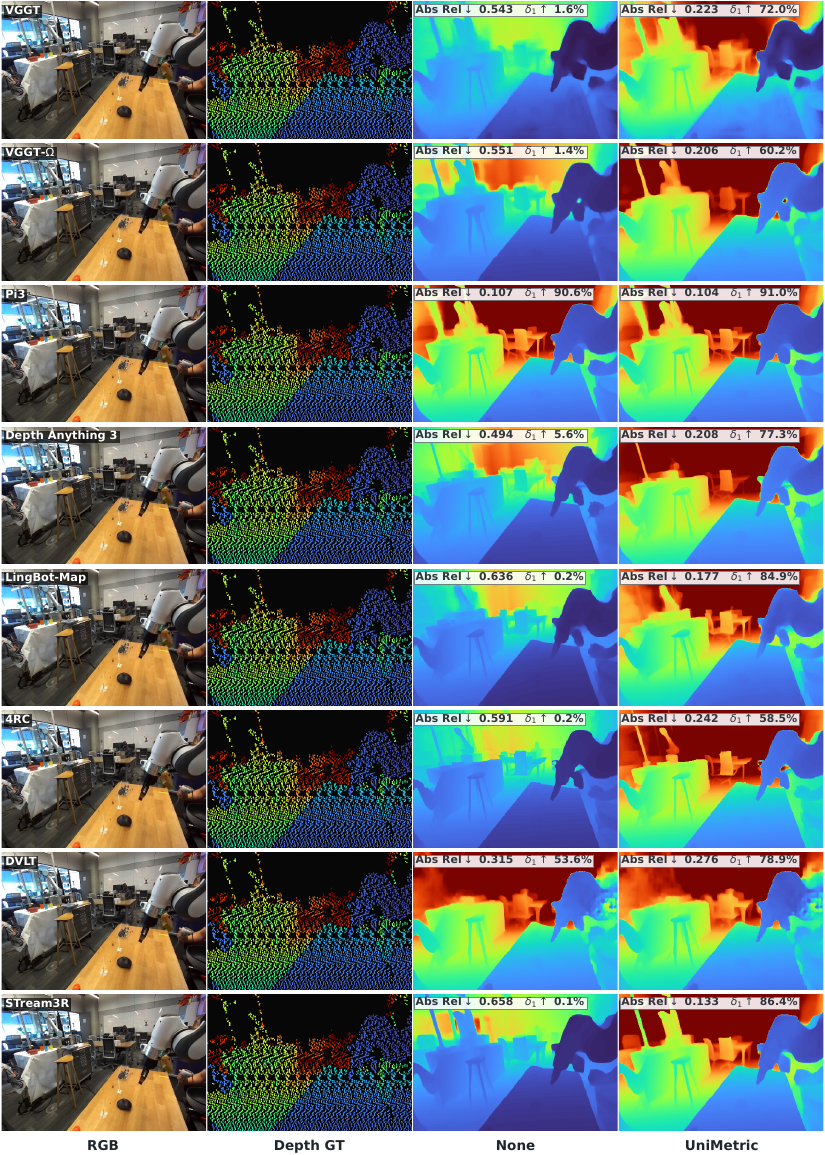}
  \caption{\chg{\textbf{Metric depth recovery on DROID (tabletop manipulation; 0--3.7\,m).}
    A robot gripper reaches over small objects, with lab equipment in the background.
    STream3R's $\delta_1$ increases from 0.1\% to 86.4\%, while $\pi^3$
    changes from 90.6\% to 91.0\%. The correction is small where the original
    prediction is already close to metric.}}
  \label{fig:supp-depth-droid}
\end{figure}
\clearpage

\begin{figure}[H]
  \centering
  \includegraphics[width=\linewidth]{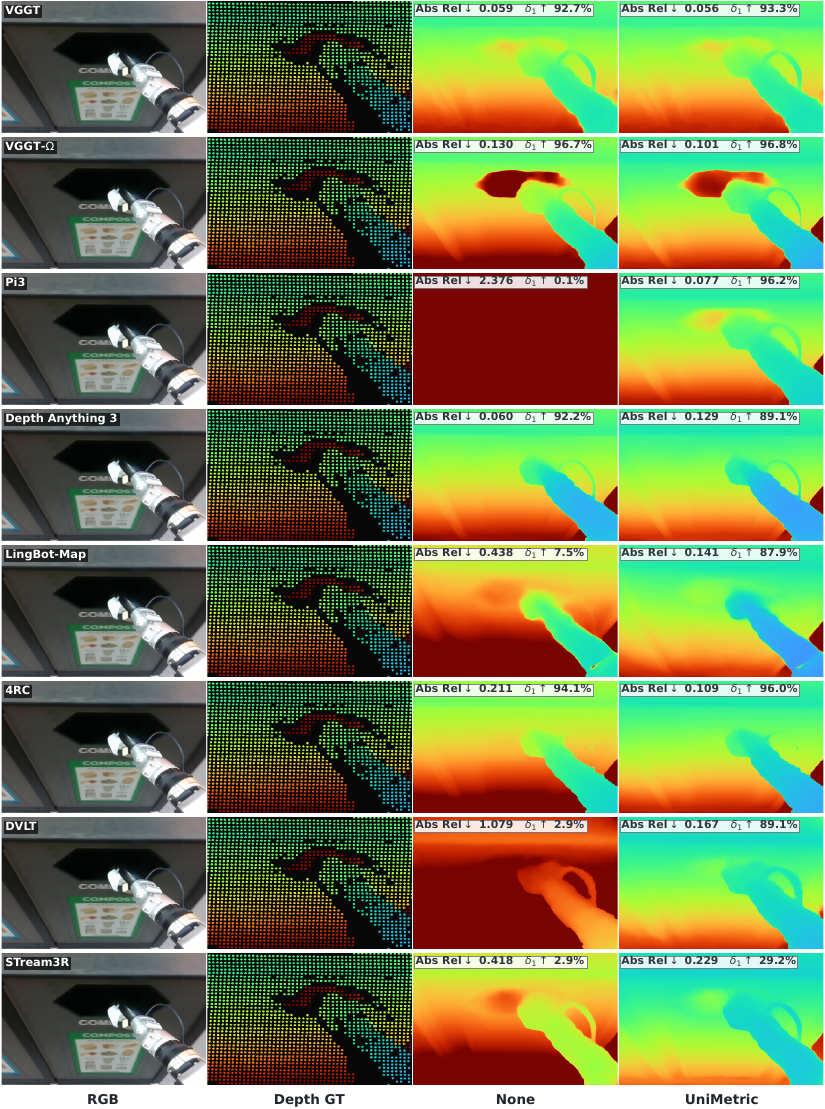}
  \caption{\chg{\textbf{Metric depth recovery on Humanoid Everyday (close-range manipulation; 0--1.2\,m).}
    A gripper holds paper above a waste bin at the closest range among these examples.
    $\pi^3$ improves from 0.1\% to 96.2\% $\delta_1$, and DVLT from 2.9\%
    to 89.1\%, while VGGT and VGGT-$\Omega$ remain near 93\% and 97\%, respectively.}}
  \label{fig:supp-depth-humanoid}
\end{figure}
\clearpage

\begin{figure}[H]
  \centering
  \includegraphics[width=\linewidth]{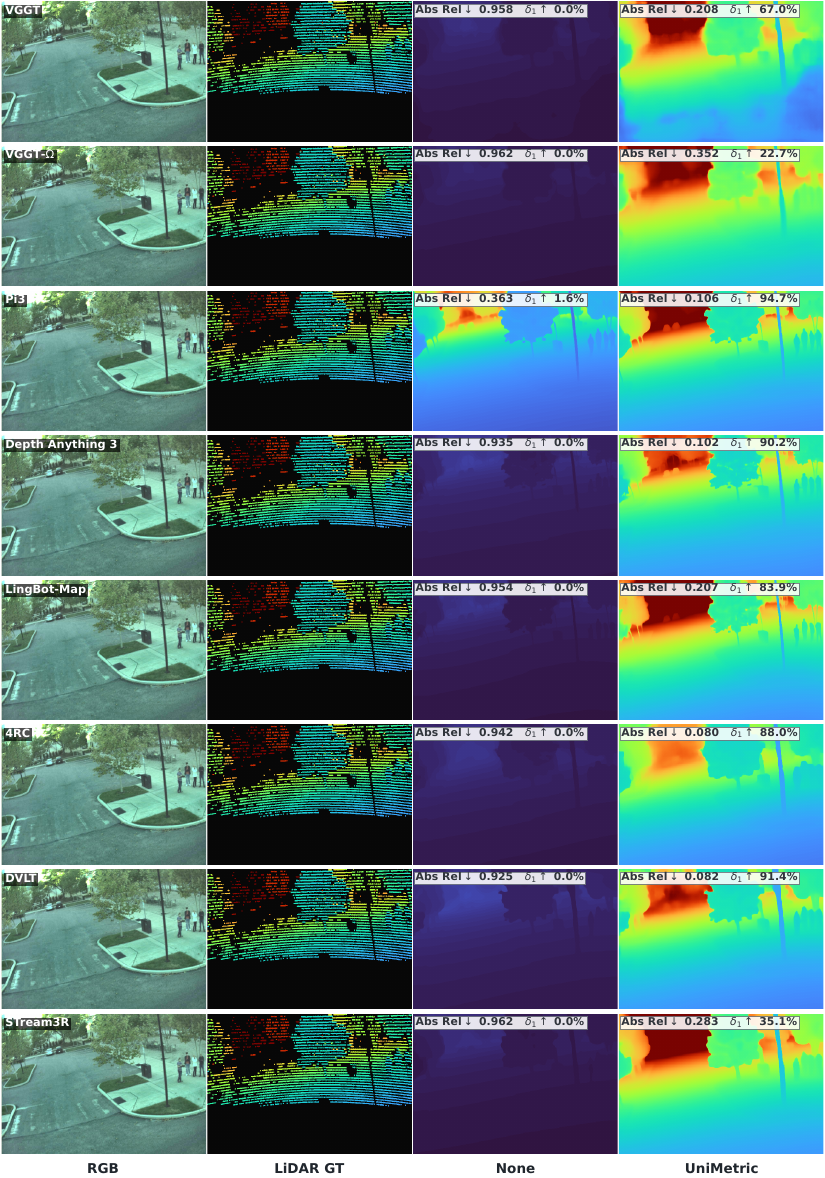}
  \caption{\chg{\textbf{Metric depth recovery on 3EED (aerial view; 0--52\,m).}
    The oblique view covers roads, trees, parked cars, and pedestrians.
    The mean frame-level $\delta_1$ increases from 0.2\% to 71.6\%, with
    $\pi^3$ and DVLT reaching 94.7\% and 91.4\%, respectively, from an aerial viewpoint.}}
  \label{fig:supp-depth-3eed}
\end{figure}
\clearpage

\endgroup

\end{document}